\documentclass[letterpaper]{article} % DO NOT CHANGE THIS
\usepackage{aaai2026}  % DO NOT CHANGE THIS
\usepackage{times}  % DO NOT CHANGE THIS
\usepackage{helvet}  % DO NOT CHANGE THIS
\usepackage{courier}  % DO NOT CHANGE THIS
\usepackage[hyphens]{url}  % DO NOT CHANGE THIS
\usepackage{graphicx} % DO NOT CHANGE THIS
\usepackage{natbib}  % DO NOT CHANGE THIS AND DO NOT ADD ANY OPTIONS TO IT
\usepackage{caption} % DO NOT CHANGE THIS AND DO NOT ADD ANY OPTIONS TO IT
\usepackage{algorithm}
\usepackage{newfloat}
\usepackage{listings}
\DeclareCaptionStyle{ruled}{labelfont=normalfont,labelsep=colon,strut=off} % DO NOT CHANGE THIS
\floatstyle{ruled}
\newfloat{listing}{tb}{lst}{}
\floatname{listing}{Listing}
\usepackage[utf8]{inputenc} % allow utf-8 input
\usepackage{url}            % simple URL typesetting
\usepackage{booktabs}       % professional-quality tables
\usepackage{amsfonts}       % blackboard math symbols
\usepackage{multirow}
\usepackage{amsmath}       % blackboard math symbols
\usepackage{nicefrac}       % compact symbols for 1/2, etc.
\usepackage{microtype}      % microtypography
\usepackage{xcolor}
\usepackage{listings}

\usepackage{tikz}
\usetikzlibrary{positioning, arrows.meta}

\usepackage{subfigure}

\usepackage{amssymb}
\usepackage{mathtools}
\usepackage{amsthm}
\usepackage{algpseudocode}

\usepackage[capitalize,noabbrev]{cleveref}

\usepackage{multirow}
\usepackage{xcolor}
\usepackage{adjustbox}

\title{Masked Mineral Modeling: Continent-Scale Mineral Prospecting\\ via Geospatial Infilling}
\author {
    Sujay Nair\textsuperscript{\rm 1,2*},
    Evan Coleman\textsuperscript{\rm 1*},
    Sherrie Wang\textsuperscript{\rm 1},
    Elsa Olivetti\textsuperscript{\rm 1}
}
\affiliations {
    \textsuperscript{\rm 1}Massachusetts Institute of Technology \quad \textsuperscript{\rm 2}Georgia Institute of Technology \quad
    \textsuperscript{\rm *}Equal contribution\\
    sujaynr@mit.edu, ecol@mit.edu, sherwang@mit.edu, elsao@mit.edu
}

\begin{document}

\maketitle

\begin{abstract}
    Minerals play a critical role in the advanced energy technologies necessary for decarbonization, but characterizing mineral deposits hidden underground remains costly and challenging. Inspired by recent progress in generative modeling, we develop a learning method which infers the locations of minerals by masking and infilling geospatial maps of resource availability. We demonstrate this technique using mineral data for the conterminous United States, and train performant models, with the best achieving Dice coefficients of 
    $0.31 \pm 0.01$ and recalls of $0.22 \pm 0.02$ on test data at 1$\times$1 mi$^2$ spatial resolution. One major advantage of our approach is that it can easily incorporate auxiliary data sources for prediction which may be more abundant than mineral data. 
    We highlight the capabilities of our model by adding input layers derived from geophysical sources, along with a nation-wide ground survey of soils originally intended for agronomic purposes. We find that employing such auxiliary features can improve inference performance, while also enabling model evaluation in regions with no recorded minerals.
\end{abstract}

\section{Introduction}
\label{sec:introduction}

Addressing climate change by transitioning the global economy away from its dependence on fossil energy sources is anticipated to require the mining of unprecedented quantities of raw material~\cite{daehn2024key,WhiteHouse2021SupplyChains,franks2023mineral}. 
Many of these deposits of metals have only recently been characterized, with notable discoveries having even occurred within developed countries where thorough searches (also known as mineral prospecting) have already been performed~\cite{wang2024regional}. The McDermitt Caldera on the border of Nevada and Oregon is a salient case: evidence suggests it may be the largest deposit of lithium-bearing rock in the world, but its potential magnitude was detailed in 2023~\cite{benson2023hydrothermal}. 
Motivated by this picture and the present need to identify and characterize mineral deposits around the globe, we seek to discover scalable approaches for mineral prospecting and subsurface mapping which are robust to the reality that existing records can be incomplete.

Meanwhile, powerful data-driven techniques to infill missing information have recently been developed and applied successfully in other fields. These include the use of masked autoencoders (MAE) for restoring patches of otherwise complete images~\cite{he2021mae}, and applications of bidirectional encoder representations from transformers (BERT) to natural language generation~\cite{devlin2018bert}. Motivated by these developments in generative modeling, we explore whether a similar paradigm can enable more powerful inference for mineral prospecting over large survey areas. 

In this work, we train such models and find that infilling techniques can indeed recover missing or ablated mineral data.
Notably, we observe that incumbent transformer-based architectures for infilling tasks (e.g. ViT) behave poorly out of the box in the subsurface infilling domain due to the highly sparse and spatially clustered nature of mineral deposits. To mitigate this, we propose a ResNet U-Net-based approach, which makes use of the strong spatial bias in mineral co-occurrence and achieves signficiantly higher performance than both ViT and classical geostatistical baselines.

As a bonus, the structure of our proposed model makes it easy to include auxiliary data. In fact, there are many datasets potentially worth applying to mineral inference which are far less sparse than mineralogical data. Over the past two decades, sizable aerial and ground-based efforts have developed substantial open-source datasets of terrestrial features~\cite{RaCADATA,orgiazzi2018lucas,kokaly2017usgs, qian2021hyperspectral,hajaj2024review}.
Due to their different intended applications, these datasets can have drastically different spatial and depth coverages than those applied to traditional mineral prospecting. An example is hyperspectral imaging of field samples: geologists have recently used these advanced cameras 
to perform field-scale prospecting~\cite{wan2021application}. However continent-scale datasets of these same measurements have already been collected over the US and Europe to model soil and ecological health~\cite{RaCADATA,orgiazzi2018lucas}. The growing number of individual such large-scale measurement campaigns presents an opportunity to fuse these diverse data sources and exploit variations in their coverage to accelerate the development, refinement, and testing of hypotheses regarding the subsurface resources accessible on Earth~\cite{qian2021hyperspectral}.

In this work, we present 4 contributions. \textbf{\textit{(1)}} We introduce \textbf{M}asked \textbf{M}ineral \textbf{M}odeling \textbf{(M3)}, a novel method and practical baseline for continent-scale mineral prospecting via geospatial infilling. \textbf{\textit{(2)}} We perform a series of experiments on the proposed model, illustrating improved performance over both learning-based and classical prior work, and favorable scaling properties. \textbf{\textit{(3)}} We explore conditioning the model on auxiliary data, finding significant performance improvements and extension of model capabilities, and lastly \textbf{\textit{(4)}}  We analyze the capabilites of the learned model and the implicitly learned relationships between different minerals.

\begin{figure*}[t]
\begin{adjustbox}{center}
    \includegraphics[width=1.0\linewidth, trim=12pt 15pt 5pt 8pt, clip]{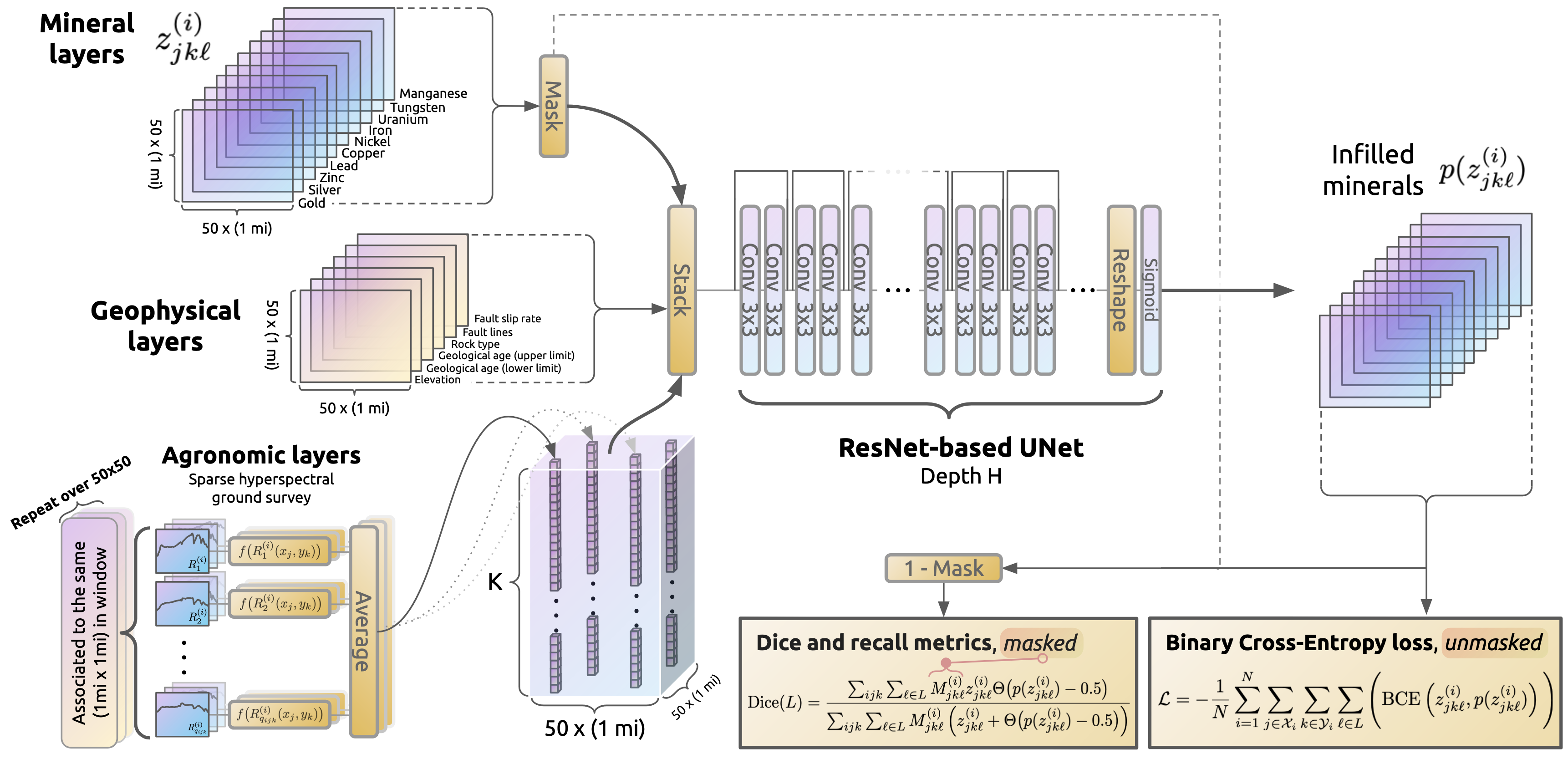}
\end{adjustbox}
    \caption{\textbf{General model architecture and data pipeline.} For a location $(x_j,y_k)$ in the $i^{\text{th}}$ (50~mi)$\times$(50~mi) context window, the availability of resource $\ell$ is marked with a per-pixel binary flag $z^{(i)}_{jk\ell}\in\{0,1\}.$ The mineral data layers are masked, stacked with geophysical and agronomic inputs, and jointly passed through either a ResNet of depth $H$ (shown) or a ViT. Masked minerals are infilled by the network, and the output is a per-pixel probability $p(z^{(i)}_{jk\ell})$ of the presence of resource $\ell$. We train to maximize Dice coefficients computed over the masked region. Due to their combination of spatial sparsity and feature richness, the agronomic data for a given pixel pass through a feature reduction network $f$ which maps them to vectors of length $K$ at that location.}
    \label{fig:model-arch}
\end{figure*}

\begin{figure*}[h]
\begin{adjustbox}{center}
    \includegraphics[width=0.9\linewidth]{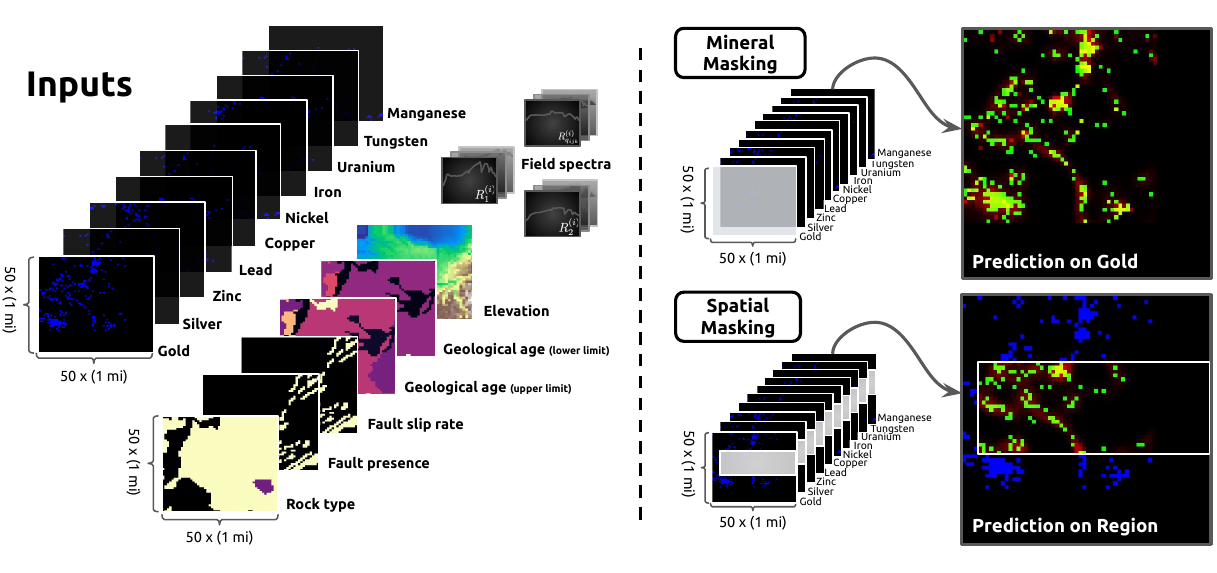}
\end{adjustbox}
    \caption{\textbf{Example input features and model outputs.} The displayed predictions are generated by the model trained in Experiment~\ref{sec:exp-georac} with $N=10$K, $H=152$, and $K=64$. The left panel shows all possible forms of input data for this context window. The first 10 layers consist of the availability of 10 minerals [Gold, Silver, Zinc, Lead, Copper, Nickel, Iron, Uranium, Tungsten, Manganese] on (50 mi)$\times$(50 mi) regions. The next 6 layers consist of geophysical inputs [Rock type, Fault presence, Fault slip rate, Geological age (upper/lower limit), Elevation]. Finally, the agronomic data is embedded and included. The panel on the right shows the two possible masking strategies, mineral and spatial, and example predictions. Mineral masking removes an entire mineral layer, while spatial masking removes a random rectangular region. The blue points are ground truth outside of the masked region, the green points are ground truth within the masked region, and the red is prediction.}
    \label{fig:model-examples}
\end{figure*}

\section{Related Work}

\textbf{Subsurface mapping.} The geospatial infilling procedure studied in this work is a form of prospectivity mapping~\cite{yousefi2021data}, also known as mineral abundance inversion~\cite{chen2024hyperspectral}, or as fossicking when done by hobbyists~\cite{franks2023mineral}. This is a task of interest within the fields of geology, lithology, and mineralogy. It falls within the broader category of subsurface mapping and characterization, the modeling of below-ground resources based on sparse subsurface data combined with surface-accessible features~\cite{misra2019machine,vohra2024automated}. Remote sensing and data fusion have been applied with varying degrees of success to such challenges, for mineral resource analysis but also other climate-relevant applications, including geological CO$_2$ reservoir modeling~\cite{liu2020petrophysical,narayan2024machine,wang2024machine} and drilling viability for the extraction of geothermal energy~\cite{wang2025machine}. Many metrics are used to compare the performance of feature reconstruction, depending on the structure of the outputs. For supervised multi-class classification of lithological phenotypes, accuracy (overall and per-class) and Cohen's $\kappa$ are common, e.g.~\cite{dong2024fusion}. For regression of elemental content, metrics are standard to spectroscopy: mean squared error (MSE), bias, ratio of performance to deviation (RPD), and the $R^2$ of a linear fit between expected and predicted abundances, e.g.~\cite{jiang2024estimation}. Focusing on ML-based mineral resource estimation, most studies prior to 2022 used MLPs or SVMs (60\%), analyzed one mineral (84\%), and relied on field-level data collection (88\%)~\cite{dumakor2021machine}.\\[-1.075em]

\textbf{Data fusion for improving geospatial inference}. Many geospatial features are frequently analyzed together in subsurface mapping, including ground-based magnetometry, ground-penetrating radar, and seismometry data, as well as satellite-based aeromagnetic and imaging spectroscopy data~\cite{vohra2024automated}. Due to its growing availability, applications of hyperspectral remote sensing and ground surveys to subsurface and mineral resource mapping have drawn recent interest within the literature~\cite{hajaj2024review,wan2021application}. Hyperspectral satellites such as AVIRIS-NG provide more granular color information than traditional (multispectral) satellites like Landsat, measuring $\sim$100$\times$ more spectral bands ($\sim$1000 colors instead of $\sim$10) with spectral resolutions $\sim$10$\times$ finer~\cite{qian2021hyperspectral}. While using such data requires processing a large number of input channels, it can support more performant models. Most recently, using feature reduction techniques and fusion of hyperspectral and multispectral remote sensing data, the authors of~\cite{dong2024fusion} showed that continent-scale lithological mapping (roughly, a precursor of the data we use as ground truth for M3) could be achieved remotely across the Tibetan plain with 97\% accuracy and only 1\% of the available training data, using ViT with dynamic graph convolutional layers.

\textbf{Masked geospatial infilling of satellite imagery.} There has been some recent interest within the literature regarding geospatial applications of MAE. Techniques and foundation models such as SatMAE and ScaleMAE have been developed which can perform spatial, spectral, and temporal infilling of satellite imagery~\cite{cong2022satmae}, impute continuous surface-accessible features such as land use/land cover (LULC), and extend geospatial correlations across distance scales~\cite{reed2023scale,tang2023cross}. Due to the continuous nature of remotely sensed imagery, infilling tasks on satellite data fall well-within the domain of vanilla MAE designed for RGB images, and the performance of ViT architectures is strong in these settings. However, in this work we focus on sparse subsurface mineral readings, rather than surface observations and satellite imagery, and find empirically that ViT-based approaches struggle over our ResNet U-Net approach, as it leverages a spatial inductive bias which is well-suited to the spatial clusters characteristic of mineral prospecting data.  

\textbf{This work: contributions and differences.} Multiple aspects of the present study are unaddressed by existing literature. 
The application of subsurface mapping techniques to mineral availability on the scale of the conterminous United States (which makes up roughly 33\% of the North American continent's surface) represents a significant increase in effective survey area, and an initial step away from site-level analysis toward Earth-scale prediction. We employ a hyperspectral ground survey dataset which was developed by the US Department of Agriculture for agronomic purposes, and which has not been used for any subsurface mapping tasks. Our problem specification is also distinct, since we perform multiple binary classifications instead of a single multi-class classification, to flag the locations of mineralogical resources over a masked region. This infilling technique is a generative approach similar to MAE, which has not been studied in this context~\cite{he2021mae}. In addition, relative to previous works which infer lithological labels from geospatial features, the ground truth data we infill is more indicative of mining viability, i.e. it flags the locations of ore of some measurable grade (concentration) which could in principle be separated from waste rock and refined by a metallurgical process.\footnote{For this initial work, we do not differentiate resources by the economics of their respective separation processes, although the grades of resources are logged.} This change in interpretation has consequences for the choice of performance metrics appropriate to the method. For example, because the classes of interest are not mutually exclusive (two resource types can coexist), the $\kappa$ score is not appropriate for model comparison. Overall accuracy is also problematic, as the mineralogical data we employ as ground-truth are an incomplete collection of true positives~\cite{schweitzer2019record}. Instead, we consider metrics appropriate to image segmentation and object detection which de-emphasize true negative detection. We use recall to track reproduction of true positives, and Dice coefficients to indicate rates of new resource prediction. Both metrics fall between 0 and 1. Larger values indicate greater agreement between predictions and ground-truth.

\section{Data}
\label{sec:data}

We aggregate multiple geospatial data layers spanning the conterminous United States, grouped into three categories: mineral, geophysical, and agronomic.

\textbf{Mineral layers.} Mineral presence layers were collected from the United States Geological Survey (USGS) Mineral Resource Data System (MRDS)~\cite{mcfaul2000us}. We extracted resource geolocations for the 10 most data-rich metallic ores; these metals are listed in Figure~\ref{fig:model-examples}. We excluded all records labeled as processing plants (as opposed to occurrences, prospects, or producers), due to inconsistent documentation on the geographic origin of processed ore.

\textbf{Geophysical layers.} For this initial study, we included a minimal suite of 6 geophysical layers from a mix of USGS-based sources. Elevation data was pulled from the USGS National Map Elevation Point Query Service~\cite{arundel2018assimilation}. We used fault presence and slip rate maps from the USGS Quaternary Fault and Fold Database~\cite{USGS2019FaultDatabase}. Rock types and geological age ranges were derived from the USGS Geologic Map of North America (GMNA)~\cite{garrity2009database}. The rock type classifications considered in the GMNA are derived from the mode of formation: sedimentary, metamorphic, igneous volcanic, igneous plutonic, and ice. We purposefully excluded the lithological classification layers of the GMNA to prevent indirectly passing mineral labels to the model and to maintain the simplicity of the geophysical inputs in our analysis, as these data are highly stratified but cannot be arranged along a single axis by a unique physical observable (e.g. geological age). 

\textbf{Agronomic layers.} We use soil survey data from the USDA Rapid Carbon Assessment Project (RaCA)~\cite{RaCADATA}, a campaign which covered the conterminous United States and collected single-pixel hyperspectral images (scans) of over 120K soil samples using an ASD LabSpec spectrophotometer at 1~nm resolution within the Visible-Near Infrared (Vis-NIR) spectral window of 350-2500~nm. Prior to scanning, each sample was air-dried, ground, and sieved to particle sizes below 2~mm. A representative fraction of the samples was sent to a laboratory for elemental analysis (via dry combustion). We identified 23,591 spectra with complete records. We removed one outlier with 48\% water content by mass, as well as 40 spectra having unlogged reflectances for one or more wavelengths. We removed wavelengths below 365 nm from consideration, due to their higher incidence of pixel failures. After applying all selection cuts, we obtain 23,550 soil spectra within a spectral window of 365-2500~nm, alongside corresponding geolocation data.

\textbf{Context window datasets.} To produce training and validation data for our analysis, we selected 100K square context windows of side length 50~mi at random from the conterminous US, and all data appropriate to that region were aggregated. Test data were prepared similarly under both in-distribution (ID) and out-of-distribution (OOD) configurations. In the in-distribution case, we pulled a single fixed set of 5K windows at random from the US. In the OOD case, we pulled 447 windows from a 300~mi square region centered on the McDermitt Caldera, and used as a validation set 417 windows from within the 50~mi square annulus surrounding it. To allow for the development of a sensible model baseline which used only mineral layers as inputs, it was required that at least one resource be present in each sampled context window. The scale of the spatial context in pixels is chosen as 2$\times$ the optimal value determined in~\cite{dong2024fusion}, so that the reduction of features due to masking still leaves sufficient information for inference. The spatial resolution was 1~mi, set by the minimum viable resolution of our choice of digital elevation model~\cite{arundel2018assimilation}. We created a separate dataset for visualization, which included context windows spanning the conterminous US scanning in 50~mi increments along lines of constant latitude and longitude.

\section{Method}
\label{sec:method}

The models trained in this work all use the full set of mineral layers $\mathcal{M}$ as a subset of their inputs, and perform an infilling task over these layers. For a given longitude and latitude $(x_j,y_k)$ in the $i^{\text{th}}$ context window $(\mathcal{X}_i,\mathcal{Y}_i)$, the availability of resource $\ell$ is marked with a per-pixel, per-mineral binary class $z^{(i)}_{jk\ell}\in\{0,1\}.$ The model output is a (50~mi)$\times$(50~mi) per-pixel probability estimate $p(z^{(i)}_{jk\ell})$ of the presence or absence of a given mineral resource. Before prediction, mineral layers are masked in both mineral space and location. Masked values of $z$ are represented as $-1$ in the inputs. The mask for mineral $\ell$ at location $(x_j,y_k)$ in the $i^{\text{th}}$ context window is denoted by $M^{(i)}_{jk\ell}\in\{0,1\}$, with $1$ indicating removal from the inputs. The mask emulates two behaviors that might realistically cause a mineral record to be missing: incomplete recordkeeping due to lack of interest in specific mineral species, and regional inaccessibility due to geographic obstacles. To account for these possibilities, we apply 2 kinds of masking with equal probability: mineral-only (i.e. mask is 1 for all $\ell$ in a subset of $\mathcal{M}$) and spatial-only (mask is 1 across all $\ell$ for all $(x_j,y_k)$ in a rectangle). An aggressiveness hyperparameter $A$ determines the maximum fraction of the mineral layers which can be masked (see Appendix~\ref{app:aggro} for its impact on performance). We sought a high, fixed $A<1$ of $0.8$ for all main results to induce a nontrivial inference task while avoiding training scenarios which represent hallucination in mineral-only inference, as doing so is expected to exaggerate the impacts of auxiliary features.

In our experiments, the geophysical and agronomic layers represent two optional sets of inputs. If selected, geophysical input layers are simply stacked with the masked mineral layers. The agronomic layers require a more sophisticated approach, as multiple hyperspectral scans can be associated to a single pixel in the context window (up to $\sim$30 vectors of length 2,135), and only up to 15 such pixels in a given 2500-pixel window have associated ground survey data. To avoid memory constraints, we apply a feature reduction $f$: each scan $R^{(i)}$ within a given longitude and latitude $(x_j,y_k)$ is passed through a 1D CNN with average pooling, ReLU activations, and a fully connected layer mapping it into $f(R^{(i)}(x_j,y_k))$, a vector of length $K$ (see Appendix~\ref{app:fexp}). All outputs for a given site are averaged, and injected into a 50$\times$50$\times K$ block at the associated geolocation $(x_j,y_k)$. That block is then stacked with the other input data layers.

The combined input channels are passed through a ResNet architecture with varying depth $H$, passed through a sigmoid, and output as an array with the same shape as the mineral layers. Predictions $\hat{z}^{(i)}_{jk\ell}$ are set to $p(z^{(i)}_{jk\ell})~>~0.5$. We use the per-pixel, per-mineral binary cross-entropy as the loss. As an evaluation metric, we consider the Dice coefficient for all masked pixels having positive flags, i.e. $\{\hat{z}^{(i)}_{jk\ell}M^{(i)}_{jk\ell}~=~1\}$ and $\{{z}^{(i)}_{jk\ell}M^{(i)}_{jk\ell}~=~1\}$, see Figure~\ref{fig:model-arch}. 

\section{Experiments}
\label{sec:experiment}

In our experiments, we seek to understand how well M3 compares to prior work and scalably leverages large data sources. In \textbf{Section~\ref{sec:exp-prio}}, we evaluate the approach against learning-based and classical prior work, trained only on mineral data. Then, in \textbf{Section~\ref{sec:exp-sua}} we study how inference performance scales as a function of the dataset and model sizes. In \textbf{Section~\ref{sec:exp-georac}}, we extract performance improvements by introducing auxiliary input layers. Lastly, in \textbf{Section~\ref{sec:exp-maps}} we probe the capabilities of M3, specifically its performance in scenarios where mineral records are nonexistent, and the dependencies of its predictions on the different mineral classes. We report seed-level standard errors of measurement alongside results.

\begin{figure*}[ht!]
    \begin{adjustbox}{center}
    \includegraphics[width=0.25\linewidth, trim= 5pt 5pt 15pt 5pt, clip]{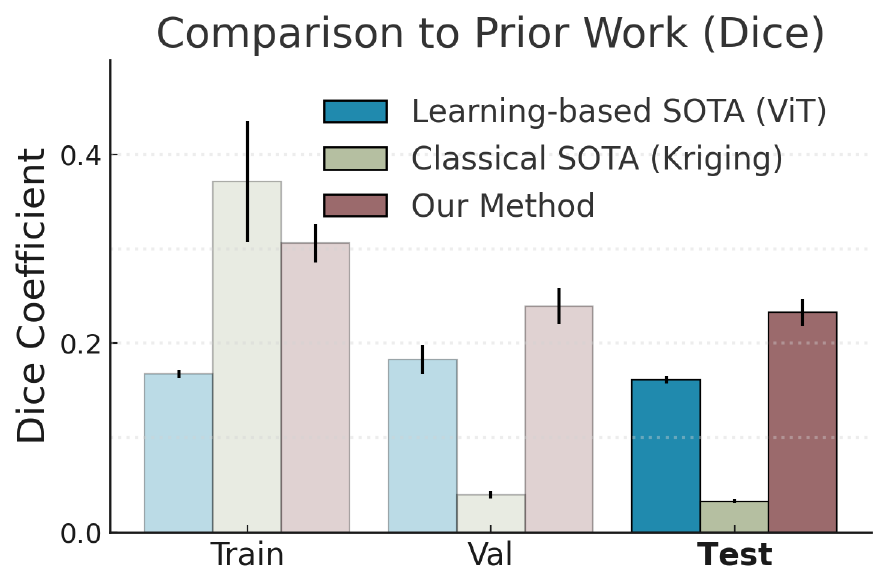}
    \includegraphics[width=0.25\linewidth, trim= 5pt 5pt 15pt 5pt, clip]{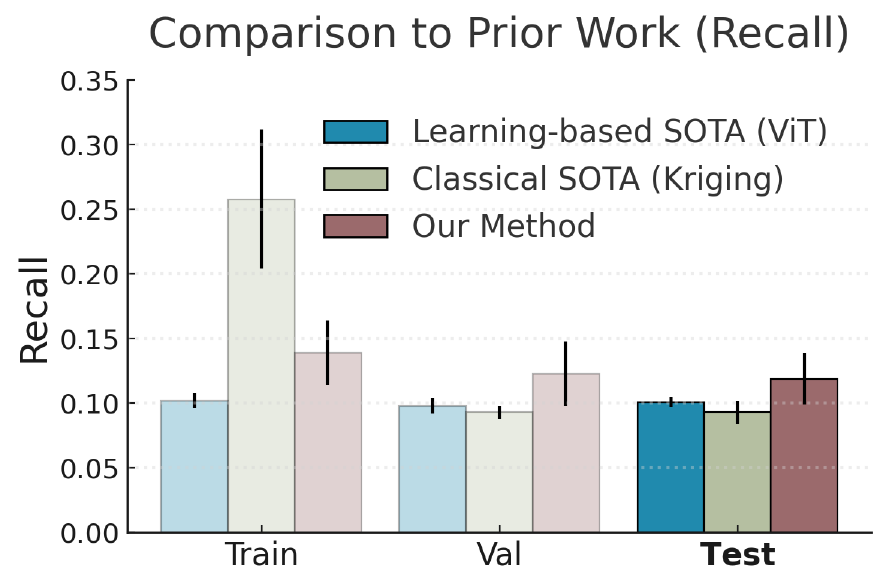}
    \includegraphics[width=0.25\linewidth, trim= 5pt 5pt 15pt 5pt, clip]{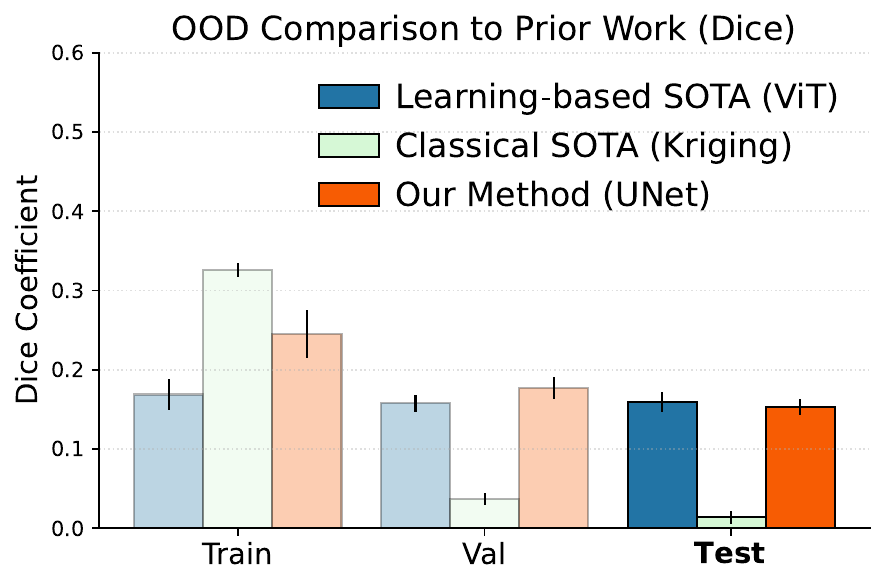}
    \includegraphics[width=0.25\linewidth, trim= 5pt 5pt 15pt 5pt, clip]{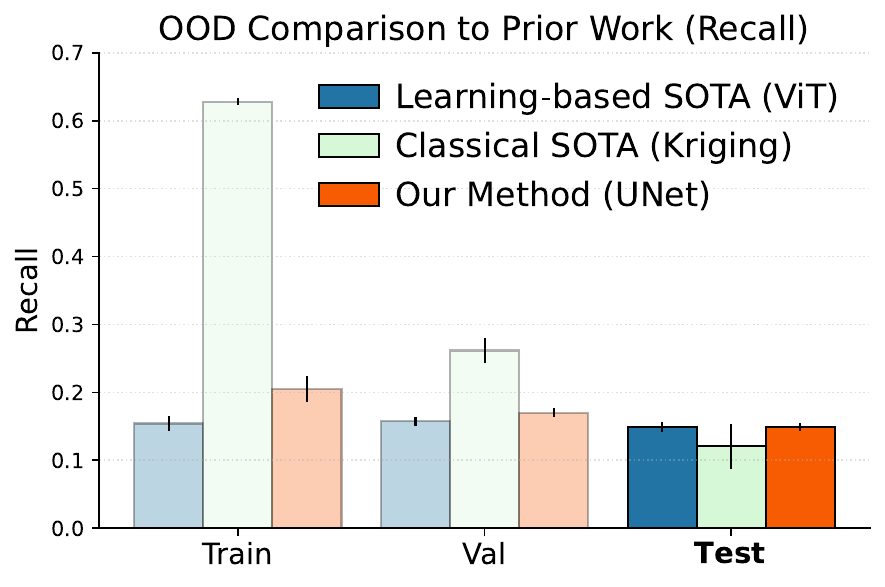}
    \end{adjustbox}
    \caption{\textbf{Comparing M3 to prior work.} Our proposed method, M3, generally outperforms prior work from both learning-based approaches (ViT), and classical geostatistical approaches (Kriging). We observe that ViT struggles to fit the data as effectively due to a lack of spatial inductive bias, and that a ResNet-based approach is better-suited to inference over sparse clustered mineral deposits. Kriging fits well to training data due to its use of a Gaussian Process, but completely fails to generalize to unseen samples. These conclusions hold especially for ID (left) generalization. For OOD generalization (right), ViT required 5$\times$ as many gradient steps to converge to equivalent validation and test performance, with significantly lower training performance. The OOD region is a 90K mi$^2$ square centered at the McDermitt Caldera, and the validation region is the surrounding 50-mi annulus.}
    \label{fig:prior-work}
\end{figure*}

\subsection{Comparisons to prior work}
\label{sec:exp-prio}

In the first experiment, we compare to baselines for mineral-to-mineral inference and understand how model generalization varies based on size and inductive bias. We train a ViT architecture as a learning-based baseline (see Appendix~\ref{app:vit}), a Gaussian Process (GP) as a geostatistical baseline (also known as \textit{kriging}; Appendix~\ref{app:kriging}), and finally our proposed M3 ResNet architecture in Figure~\ref{fig:model-arch}. We perform an initial grid sweep in model sizes and optimizer hyperparameters for a training dataset of size $N=10$K instances (details in Appendix~\ref{app:adamopt}). The models use masked mineral layers as inputs. We train each ML model for $100$K gradient steps, on either a 90:10 train-validation split (for ID) or the OOD split described in Section~\ref{sec:data}. We repeat on 3 fixed random seeds in each configuration, and 7 seeds in the OOD ViT and M3 runs. We set masking aggressiveness to $A=0.8$. We save checkpoints every 10K steps, and evaluate performance when the validation Dice coefficient asymptotes. 

Figure~\ref{fig:prior-work} summarizes the performance statistics, demonstrating equivalent or superior generalization performance of the M3 architecture over both baselines. The GP outperforms M3 in training set performance, but sharply drops during validation and testing. This behavior is expected as kriging is the best linear unbiased predictor (BLUP) to interpolate features in the training distribution without straying from ground-truth, yet we explicitly added nonlinear effects through masking and geographic scale which violate kriging's assumptions of homoskedasticity, stationarity, and isotropy~\cite{kleijnen2017kriging}. However, any superior generalization performance of our approach over ViT is surprising, given that the attention mechanisms in transformers lend themselves to extraction of long-range correlations. Coarse-graining the inputs 10$\times$ for the best hyperparameter combinations and retraining destroys this advantage. These findings suggest that in this setting the mineralogical features are a unique combination of locally clustered and sparsely sampled, so that the inductive bias of convolutional downsampling is more enabling than the versatility of geospatial attention. This argument is similarly supported by the poor performance of the GP, because that baseline hinges upon successful regression of a faithful model of the correlations between sampled points as a function of their relative distances. We propose that this phenomenon might be more widespread within subsurface mapping, where true positives for resources such as critical minerals, drilling locations for geothermal energy harvesting, and safe CO$_2$ injection sites may be rare or sparsely-sampled but locally-clustered due to a larger-scale feature situated deep within the Earth's crust (e.g. magmatic veins, viable caprock).

\subsection{How does performance scale with data quantity and model size?}
\label{sec:exp-sua}

\begin{figure*}[t!]
\begin{adjustbox}{center}
\begin{minipage}{0.25\linewidth}
    \centering
    \includegraphics[width=\linewidth]{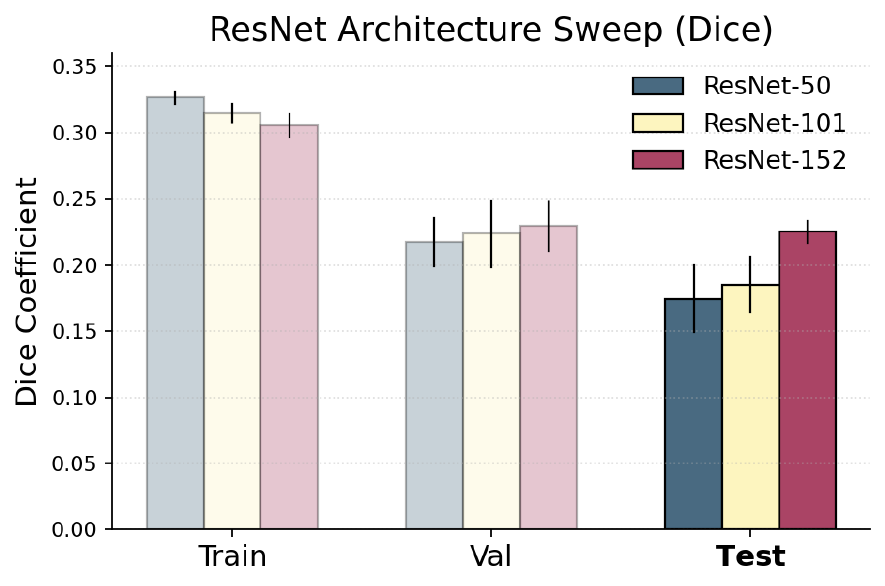}
\end{minipage}
    \begin{minipage}{0.25\linewidth}
        \centering
        \includegraphics[width=\linewidth]{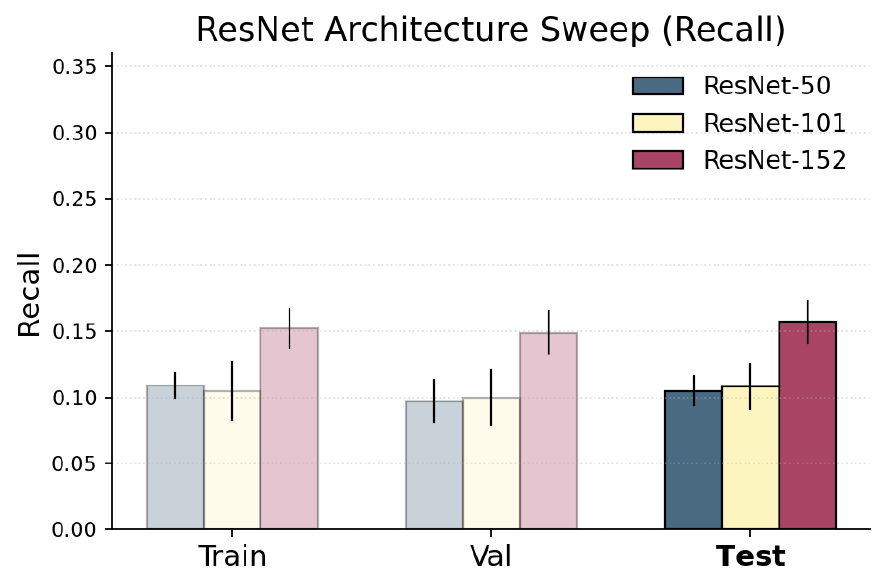}
    \end{minipage}
\begin{minipage}{0.25\linewidth}
    \centering
    \includegraphics[width=\linewidth]{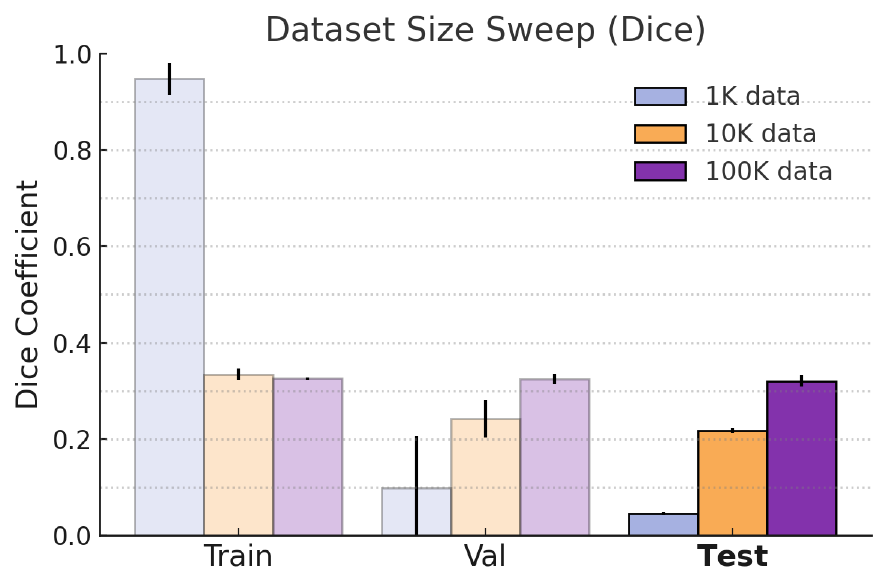}
\end{minipage}
    \begin{minipage}{0.25\linewidth}
        \centering
        \includegraphics[width=\linewidth]{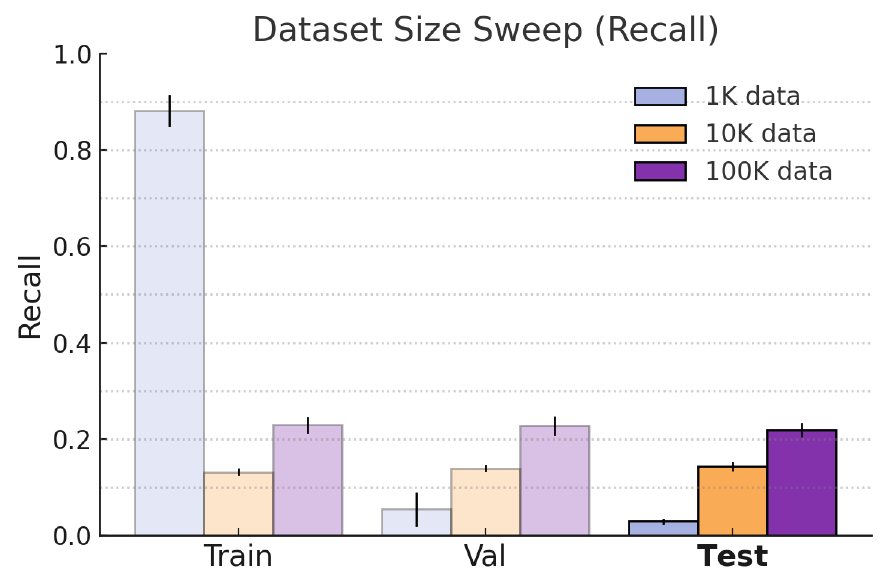}
    \end{minipage}
\end{adjustbox}
    \caption{\textbf{Scaling properties of M3.} results: We study how the M3 model performance scales as a function of  architecture depth $H$ (\textbf{Left}) and dataset size $N$ (\textbf{Right}). Dice coefficients (Top) over the masked regions, averaged over each mineral layer, and the recall on test data (\textbf{Bottom}) over the masked regions in each considered scenario. We observe favorable scaling, with improving performance with bigger models and more data.}
    \label{fig:min2min-perf}
\end{figure*}

Next we seek to understand how the M3 model scales over dataset and model sizes. In Figure~\ref{fig:min2min-perf} we sweep over ID training sets of size $N\in [1\text{K},10\text{K},100\text{K}]$ and model sizes from ResNet50 backbone to ResNet152. We see clearly favorable scaling properties and significant gains in test performance as the training dataset size grows. We also see benefit from increased model size, but interestingly only at the ResNet152 scale, where test performance is considerably higher than both ResNet50 and 101. Overall, these results suggest that as a method M3 has potential to continue improving in performance as more comprehensive datasets are assimilated and resources become available to train larger models.

\begin{figure*}
\begin{adjustbox}{center}
    \includegraphics[width=0.33\linewidth, trim= 0pt 0pt 0pt -100pt]{ 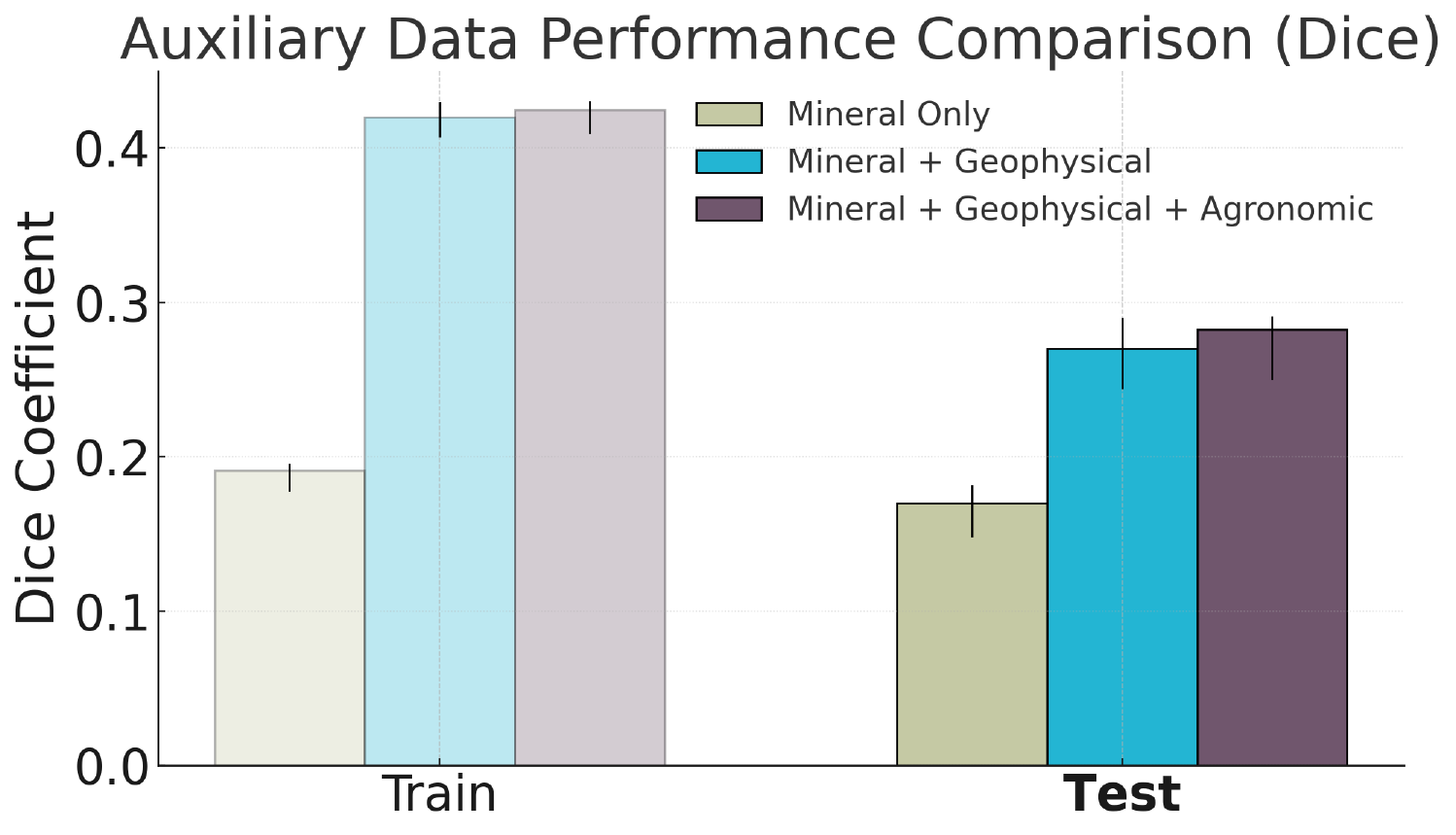}\hspace{0.25cm}
    \includegraphics[width=0.33\linewidth, trim= 250pt 60pt 240pt 60pt, clip]{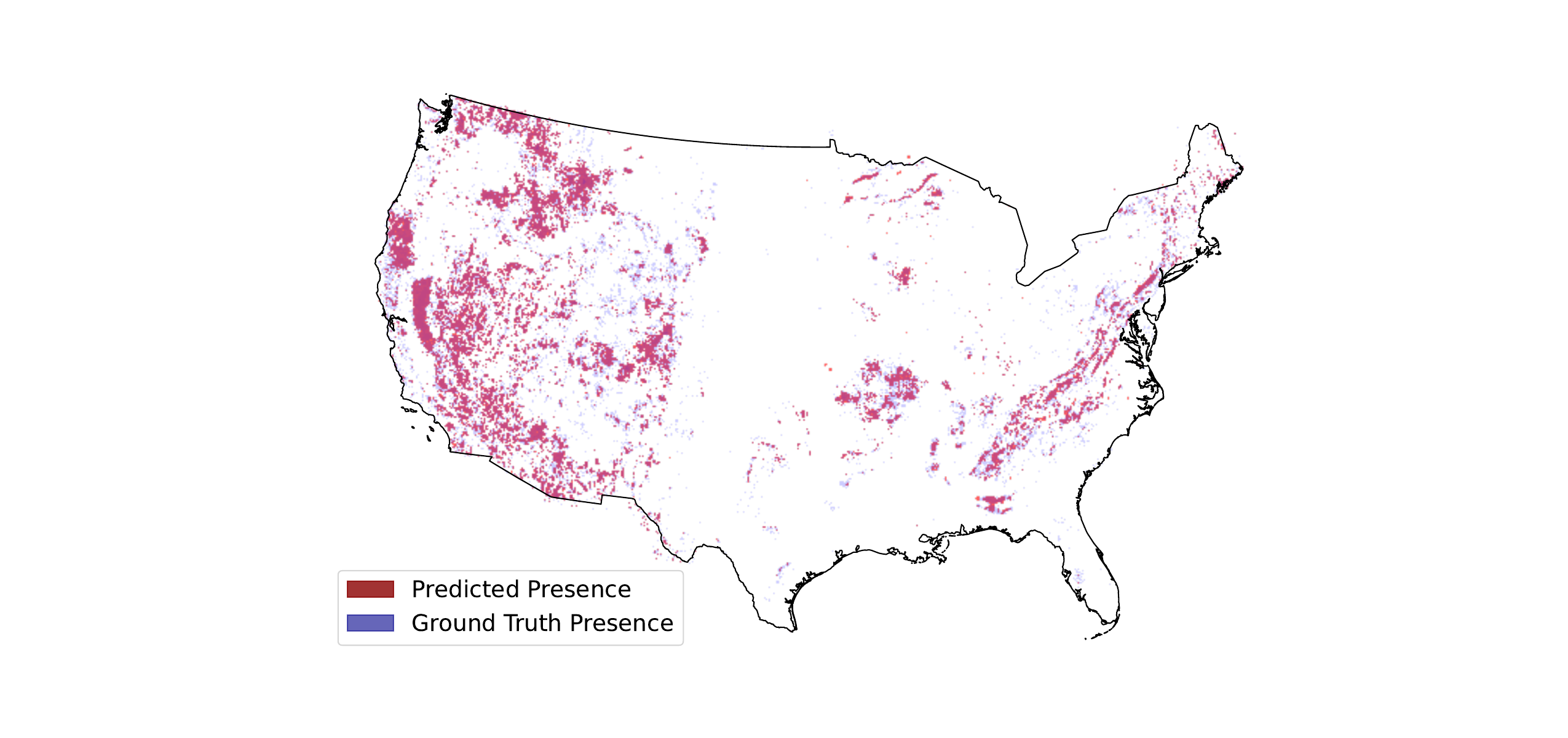}\hspace{0.25cm}
    \includegraphics[width=0.26\linewidth, trim= 90pt 5pt 110pt 20pt, clip]{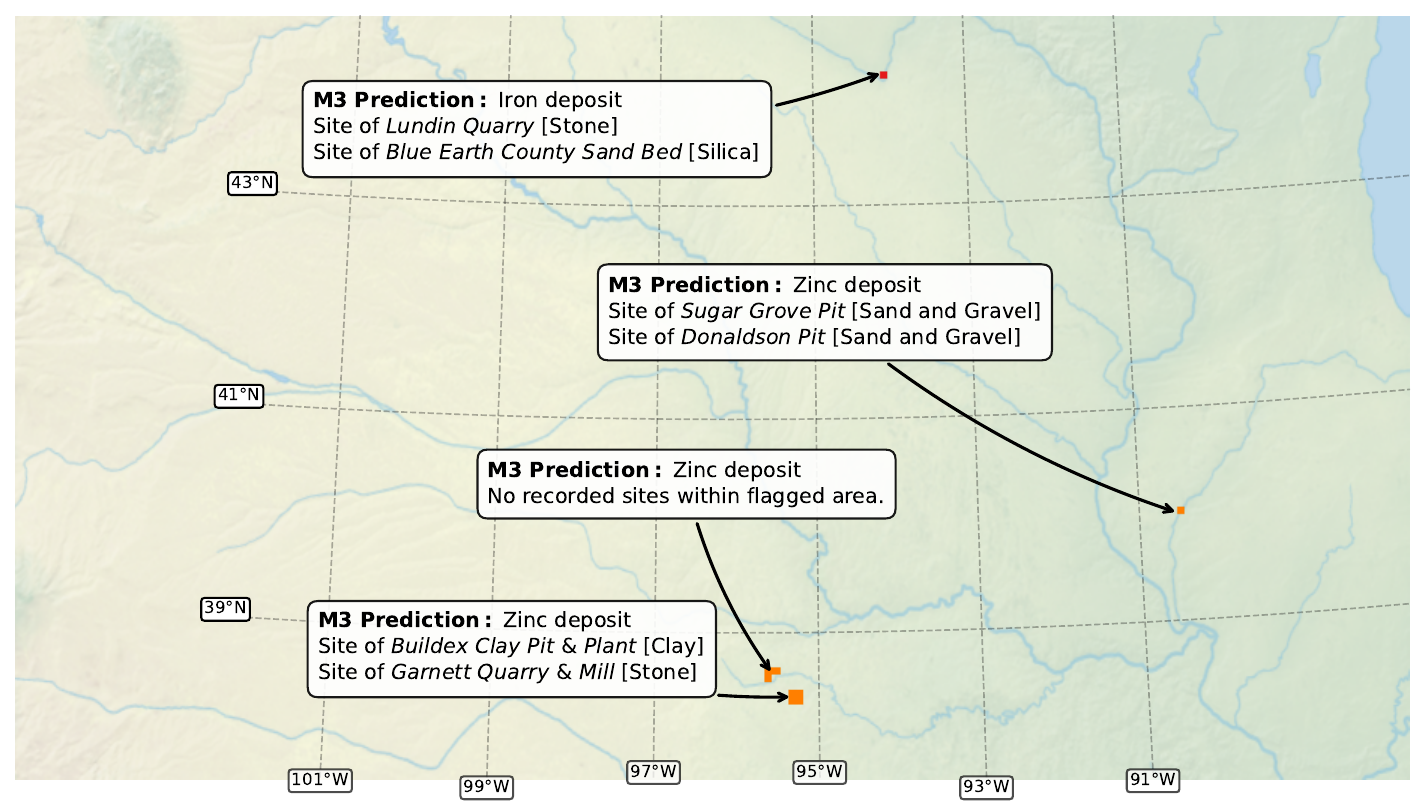}
\end{adjustbox}
    \caption{\textbf{Adding auxiliary data to M3 (Left). } Dice coefficients on training and test data as a function of input layer combinations ($N=10$K, $A=0.8$, $H=152$, $K=64$, 30K gradient steps). The inclusion of auxiliary data leads to faster training and significant generalization improvements, driven by geophysical data. 
    \textbf{Visualizing M3 predictions over the continental US (Center).} An M3-generated map of resources in the conterminous United States (EPSG:5070). The model including agronomic and geophysical input layers in Experiment~\ref{sec:exp-georac} was evaluated 150 times over a visualization dataset whose (50~mi)$\times$(50~mi) tiles covered the United States.
    Model outputs from the masked regions (\textcolor{red}{red}) were summed over mineral layers, with ground truth overlaid in \textcolor{blue}{blue}. \textbf{The model predicts 4 mainland clusters (Right)} of iron and zinc resources where no prospecting records on the top 10 minerals exist; 3 match existing sand, silica, stone, and clay mines.
    (See Appendix~\ref{app:extramaps} for details.)}
    \label{fig:usa-performance}
    \label{fig:var-importances}

\end{figure*}

\subsection{Do auxiliary inputs improve performance?}
\label{sec:exp-georac}

To understand the potential benefits of including the geophysical and agronomic layers, in this experiment we train models ID from scratch for 30K gradient steps, progressively adding the auxiliary feature sets as inputs alongside the mineral data. Based on the results of Experiment~\ref{sec:exp-sua} with $N=10$K, we choose $H=152$ and $K=64$, and repeat on 3 fixed random seeds. Figure~\ref{fig:var-importances} quantifies the performance differences between these 2 approaches and the corresponding mineral-only baseline. We find significant training and generalization improvements from including geophysical data. The agronomic layers contribute a marginal boost in performance, likely due to their spatial sparsity, but did generally lead to better performance throughout the full training process. 

\subsection{Model capabilities and learned features}
\label{sec:exp-maps}

In this section, we probe deeper into model capabilities. First, we study how the model predicts with no mineral input, only geophysical and agronomic data. Then we explore how mineral prediction depends on number and type of input minerals.

\noindent \paragraph{Evaluation on empty mineral records}
\label{exp-empty}

A noteworthy consequence of including auxiliary features is that they provide more complete coverage. For models which only consider mineral layers as inputs, inference over regions with zero prospects is effectively hallucination, i.e. infilling a blank image. However, with auxiliary features, we can evaluate the model in regions not containing any minerals and transfer learned copresence features to new localities using these additional inputs. Figure~\ref{fig:usa-performance} demonstrates this capability over the conterminous US, and plots locations where any resources are predicted when a model from Experiment~\ref{sec:exp-georac} is evaluated 150 times on the visualization dataset (i.e. it infills 150 masks per context window in the dataset). The model is conservative in that it reproduces ground-truth without rampant hallucination in previously unseen regions. In 150 evaluations, it flags only 3,775 mi$^{\rm 2}$ of land in the 819,050 mi$^{\rm 2}$ region having no training data (0.46\%). These sites include 1 iron and 3 zinc clusters on the mainland, 3 matching locations to existing sand, stone, and clay mines. (See Appendix~\ref{app:extramaps} for details.)

\noindent \paragraph{Evaluation on systematically-removed mineral records}
\label{sec:exp-removed}

\begin{figure*}[t!]
\begin{adjustbox}{center}
    \includegraphics[width=0.40\linewidth, trim=180pt 16pt 5pt 11pt, clip]{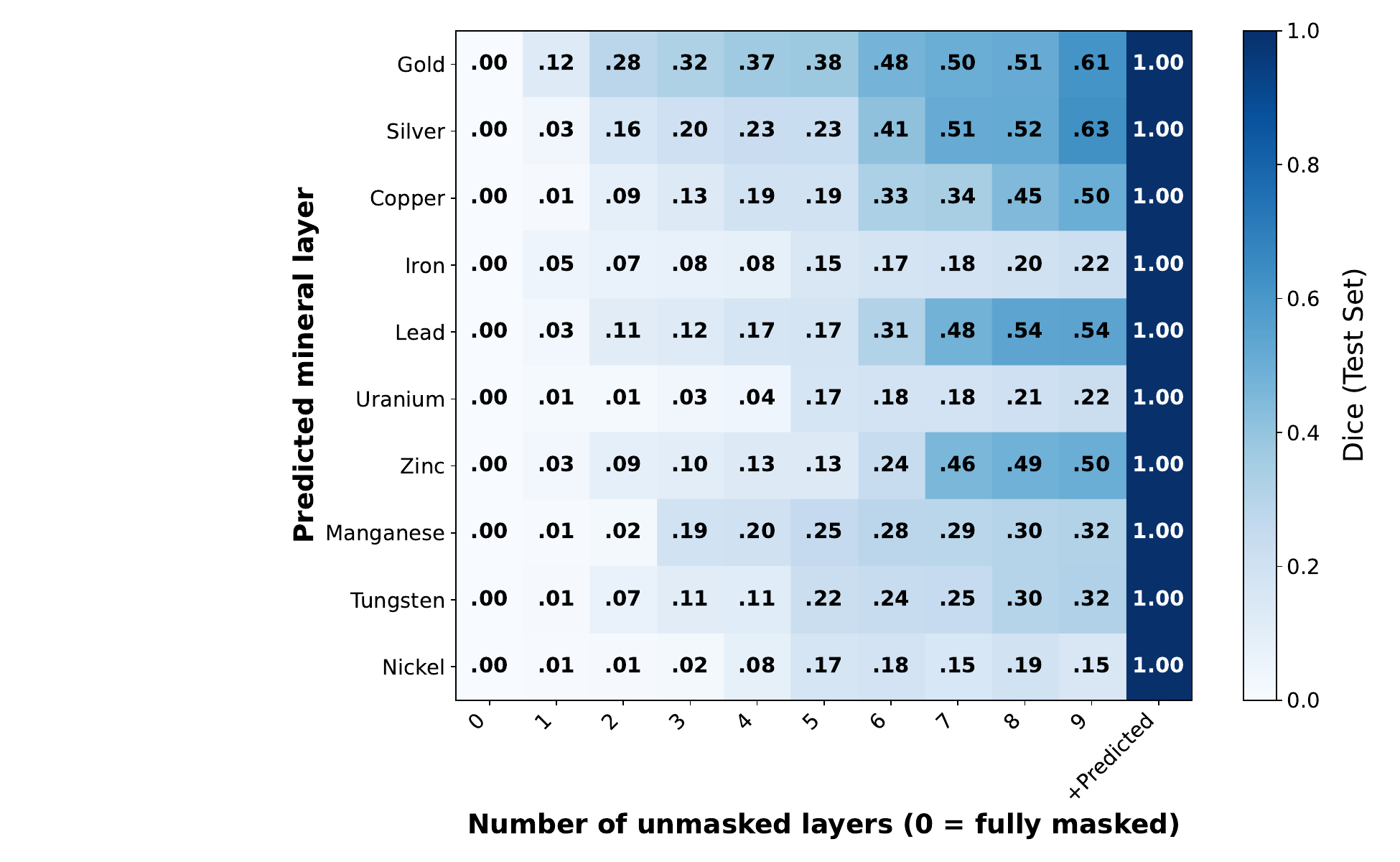}
    \hspace{0.5cm}
    \includegraphics[width=0.40\linewidth, trim=180pt 16pt 5pt 11pt, clip]{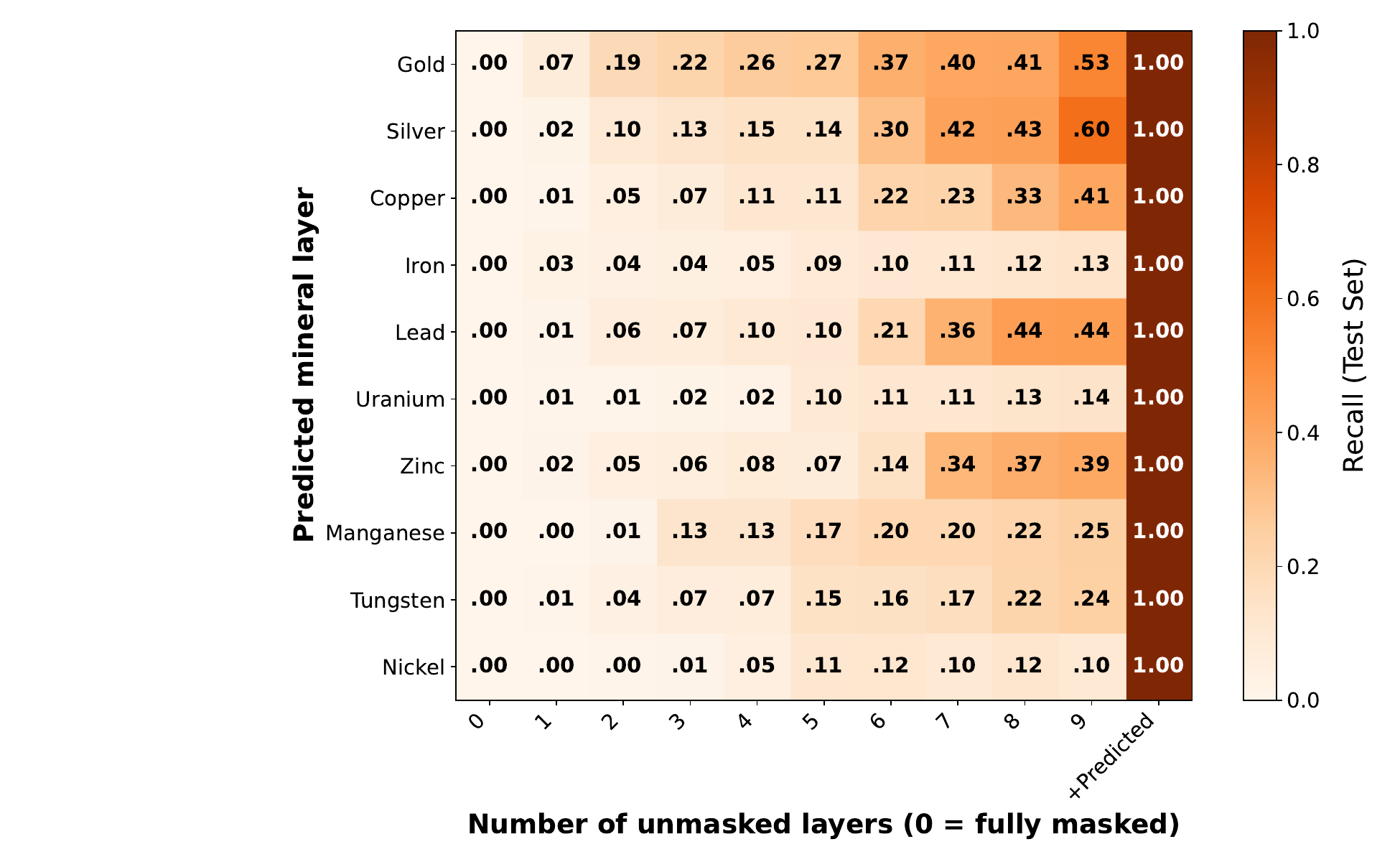}
\end{adjustbox}
    \caption{\textbf{Performance as a function of number of masked input minerals.} Matrix of single-mineral Dice coefficients and recalls. In each of 10 trials, we start from 1:1 mineral-only inference. We predict the mineral on the vertical axis from only one unmasked mineral layer, excluding the output layer's data until the final step. Unmasking occurs in order of increasing statistics, depicted here as the vertical axis labels when read from bottom to top. The model achieves perfect prediction at the final step ($1.00$ in all metrics), indicating that it learns not to infill regions unless they are masked. The model was trained on only mineral layers as inputs, with $H=152$, $N=100$K, and $A=0.8$. We observe a general trend of increasing performance as we progressively unmask, with some minerals like Zinc or Gold requiring fewer unmasked minerals to predict effectively.}
    \label{fig:rel-min-perf}
\end{figure*}

In this experiment, we compute Dice coefficients one mineral population at a time, to understand whether minerals with more prospecting records have outsized impact on performance metrics. Using the best model in Experiment~\ref{sec:exp-prio}, and for each mineral layer, we evaluate on test data after masking all but 1 mineral layer (1:1 prediction). We then compute evaluation metrics for positive predictions of only that mineral layer, and repeat with 1 extra mineral unmasked. We continue this process, unmasking 1 layer at a time until 9 of the 10 mineral layers are implemented. Finally, we unmask the predicted layer, and evaluate one last time. With the exception of the predicted layer, which is always unmasked last, the unmasking sequence is in ascending order of the minerals' commonality in the training dataset, so that an increasing quantity of mineral data is added in with each unmasking step. Figure~\ref{fig:rel-min-perf} shows the performance results on test data for all 10 runs. The vertical axis is ordered by the abundance of mineral data, with the top (gold) being the most common and the bottom (nickel) being the first unmasked layer.

We observe a general trend of increasing performance as we progressively unmask, with some minerals requiring fewer unmaskings to predict effectively. Interestingly, iron and uranium exhibited lower performance despite being the 4$^{\text{th}}$ and 6$^{\text{th}}$ most abundant layers. Though relative inference power suffers for these minerals, their addition as an input feature can seemingly boost inference for other species. An example is iron for silver: iron ties for second-to-last in scale-up, with a Dice coefficient of 0.22 after unblinding all other minerals, yet after unblinding it in the silver test it leads to a Dice coefficient boost of 0.10, taking third only to gold (0.11) and tungsten (0.13). In addition, the substantial inference power obtained after 9 of 10 layers are unmasked is surprising in the context of the literature, because machine learning-based prospectivity mapping has tended to analyze data from only one or two minerals at a time (48/51 or 94\% before 2022, maximum of 4)~\cite{dumakor2021machine,leite2009artificial}. Our result demonstrates that scaling the number of mineral inputs in M3 bolsters inference, and can be pivotal to achieving predictive power for minerals that have fewer prospecting records, as is the case for many minerals critical to addressing climate change (e.g. lithium and the platinum group metals). (See Appendix~\ref{app:influence} for details.)

\section{Future Work}

We introduced Masked Mineral Modeling (M3), a method to infer locations of mineral resources by infilling masked geospatial maps of prospecting records. Using presence data for the 10 most abundant minerals in the conterminous US, we trained ResNet-backed infilling models, benchmarked their performance against both ML and geostatistical baselines, and demonstrated their superiority in this setting. Many avenues for investigation remain. Further scaling of both the number of input features and parameters is likely a straightforward way to improve generalization. It is also worth evaluating whether graph-based architectures extract more predictive power from sparse hyperspectral data in the agronomic layers. Applying satellite imagery and pretrained models such as SatMAE~\cite{cong2022satmae} to this problem are tantalizing possibilies. However, as mining activity is easily identified by satellite~\cite{fonseca2024enhanced}, some effort is required to understand how these data might bias M3. Lastly, M3 can act as the main building block for a more sophisticated technique to reverse-engineer a masking procedure end-to-end, and determine when resources have been systematically ablated from records. (See Appendix~\ref{app:regmask}.)

\newpage
\bibliography{refs}

\newpage

\appendix
\onecolumn

\section{Inter-mineral Dependencies}
\label{sec:exp-mindep}
\label{app:influence}

\begin{figure*}[h!]
    \begin{adjustbox}{center}
        \includegraphics[width=0.525\linewidth]{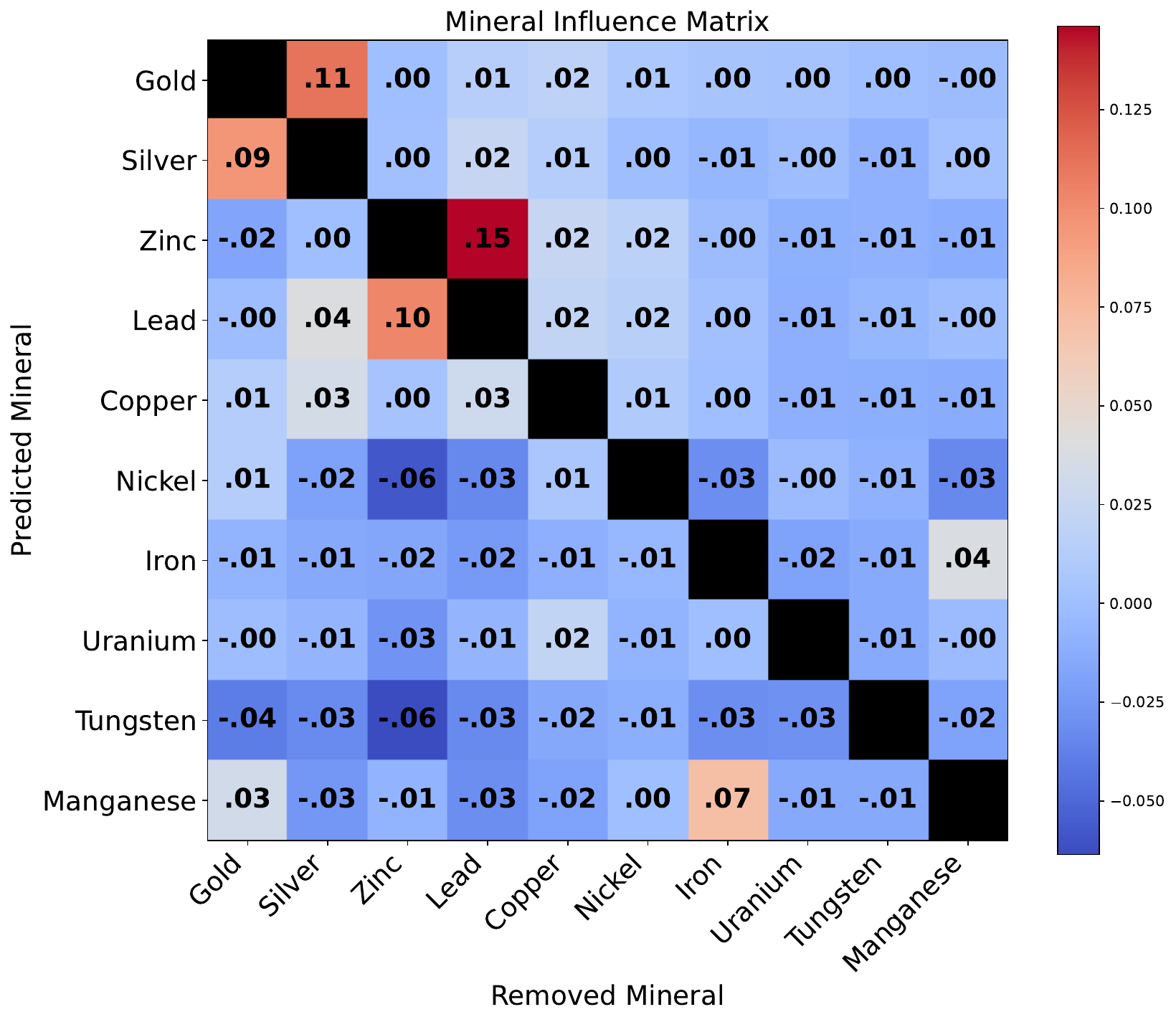}\hspace{10px}
        \includegraphics[width=0.55\linewidth, trim = 0px 0px 0px 0px, clip]{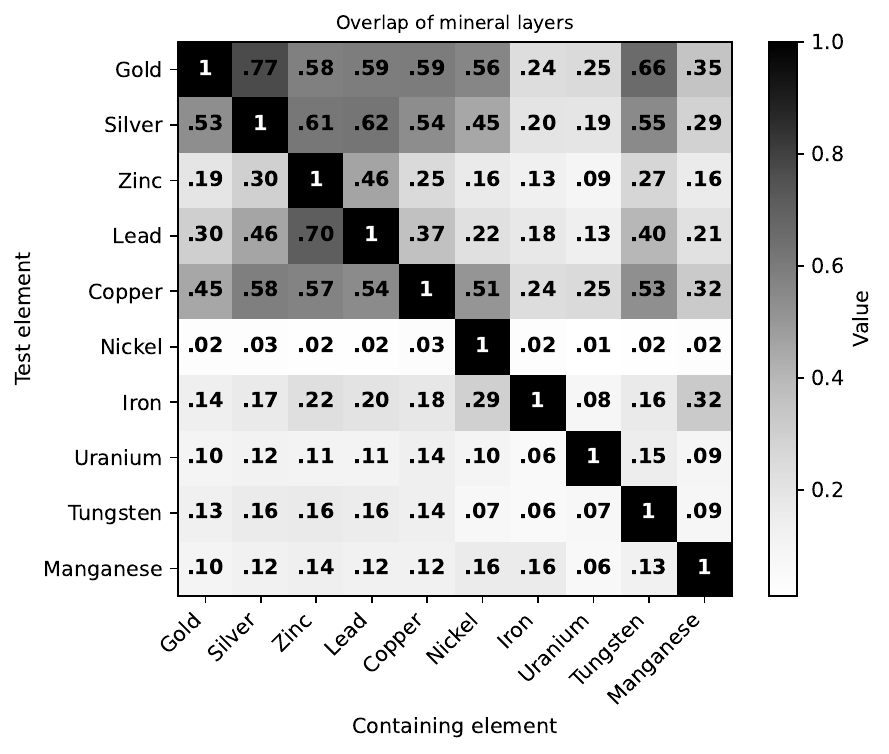}
    \end{adjustbox}
    \caption{\textbf{Learned inter-mineral dependence (Left).} Influence matrix showing the impact of holding out a specific mineral (x-axis) on the ability of the model to predict another mineral (y-axis). Each value is the Dice coefficient of the full model (all minerals included) minus the Dice coefficient of the model with the corresponding mineral held out, evaluated on the test dataset. Gray to red colors signify improvements while blue indicates negligible or negative impact. \textbf{Inter-mineral co-occurence in data (Right).} The right plot measures how often a data point which includes the ``test element'' also contains the ``containing element'', that is, the conditional probability in the data of one given the presence of the other. }
    \label{fig:influence-matrix}
\end{figure*}

    The chart in Figure~\ref{fig:influence-matrix} below provides a coarse metric to flag when data for a given mineral is especially predictive of that for another mineral. We evaluate these metrics on the in-distribution M3 instance from Experiment~\ref{sec:exp-prio}. We compare them side-by-side with summarial statistics of the individual mineral layers over the entire dataset, specifically the fraction of a given (test) element's resources which co-occur with any other (containing) element's resources. The trends across these two phenomena do not generally align, presenting evidence that the impact of a single mineral layer on the predictions of another may not be related to their direct spatial overlap.

\section{Feature reduction architecture}
\label{app:fexp}

In the case where agronomic layers are included as inputs, a feature reduction network $f$ is applied per-pixel to all hyperspectral scans associated to that geolocation in the RaCA dataset. The structure of that network is as follows:

\begin{algorithm}[h!]
\caption{Conv1D Encoder Model Architecture for Feature Reduction $f$}
\begin{algorithmic}[1]
\Require Input dimension $input\_dim = 2150$, Output dimension $K$, Hidden channels $h = 64$
\State \textbf{Step 1: Input Processing}
\State Reshape input $x$ to $(batch\_size, 1, input\_dim)$

\State \textbf{Step 2: Convolutional Layers}
\For{$j = 1$ to $3$}
    \State Apply $j^{\text{th}}$ 1D convolution: $x \mapsto \text{ReLU}(\text{Conv1D}(x, h, \text{kernel}=5, \text{stride}=2, \text{padding}=2))$
\EndFor

\State \textbf{Step 3: Flattening}
\State Reshape $x$ to $(batch\_size, h \times (input\_dim / 8))$

\State \textbf{Step 4: Fully Connected Layer}
\State Compute final embedding of form $\text{Linear}(x, K)$

\State \textbf{Output:} Embedding of size $output\_dim = K$
\end{algorithmic}
\end{algorithm}

This model is then applied to every spectrum $R_{n}^{(i)}$ which is associated to a given pixel $(x_j, y_k)$ in context window $(\mathcal{X}_i, \mathcal{Y}_i)$, where $n=1,\ldots,q_{ijk}$. Each $f(R_{n}^{(i)}(x_j, y_k))$ is averaged across the $n$ index and the resulting vector of length $K$ is embedded into a block of size $(50, 50, K)$ at the pixel corresponding to $(x_j, y_k)$ in the first two dimensions. This block is then passed as $K$ input features into the ResNet, alongside the masked mineral layers $\mathcal{M}$.

\section{Machine learning baselines: ViT architecture and training details}
\label{app:vit}

In this appendix, we explain the vision transformer (ViT) architecture which we use as a machine learning baseline for M3. The original ViT implementation of~\cite{dosovitskiy2020image} can be found in Algorithm~\ref{alg:origvit}; ours in Algorithm~\ref{alg:ourvit}.

\begin{algorithm}
\caption{Vision Transformer (ViT) Architecture from~\cite{dosovitskiy2020image}}
\label{alg:origvit}
\begin{algorithmic}[1]
\State \textbf{Input:} Image $x \in \mathbb{R}^{H \times W \times C}$, patch size $P \times P$, number of patches $N = HW / P^2$
\State Split $x$ into $N$ non-overlapping patches $x_p \in \mathbb{R}^{N \times (P^2 \cdot C)}$
\State Linearly project patches: $x_p E$, where $E \in \mathbb{R}^{(P^2 \cdot C) \times D}$
\State Add learnable classification token: $z^0 = [x_{\text{class}}; x_p E]$
\State Add positional embeddings: $z^0 = z^0 + E_{\text{pos}}$, where $E_{\text{pos}} \in \mathbb{R}^{(N+1) \times D}$
\For{$\ell = 1$ to $L$} \Comment{Transformer layers}
    \State $z'^{\ell} = \text{MSA}(\text{LN}(z^{\ell-1})) + z^{\ell-1}$ \Comment{Multi-head self-attention}
    \State $z^{\ell} = \text{MLP}(\text{LN}(z'^{\ell})) + z'^{\ell}$ \Comment{2-layer MLP with GELU}
\EndFor
\State Final representation: $y = \text{LN}(z^L_0)$ \Comment{Output from class token}
\State \textbf{Output:} Class logits via head MLP or linear layer applied to $y$
\end{algorithmic}
\end{algorithm}

\begin{algorithm}
\caption{ViT Architecture for M3 ML Baseline}
\label{alg:ourvit}
\begin{algorithmic}[1]
\State \textbf{Input:} Tensor $x \in \mathbb{R}^{B \times N \times H \times W}$

\State {Downsample spatial resolution}
\Statex \qquad $x \gets \text{Conv2D}(x, \text{kernel}=3, \text{stride}=2, \text{padding}=1)$ \Comment{$H,W \rightarrow H/2, W/2$}

\State {Flatten and project to Transformer hidden dimension}
\Statex \qquad $x \gets \text{reshape}(x) \in \mathbb{R}^{B \times (H/2 \cdot W/2) \times N}$
\Statex \qquad $x \gets \text{Linear}(x) \in \mathbb{R}^{B \times (H/2 \cdot W/2) \times D}$

\State {Add 2D positional embeddings}
\Statex \qquad $\text{row\_emb} \gets \text{Embedding}(H/2, D/2)$
\Statex \qquad $\text{col\_emb} \gets \text{Embedding}(W/2, D/2)$
\Statex \qquad $\text{pos} \gets \text{concat}(\text{row\_emb}, \text{col\_emb}) \in \mathbb{R}^{(H/2 \cdot W/2) \times D}$
\Statex \qquad $x \gets x + \text{pos}$

\State {Run through Transformer encoder}
\Statex \qquad $x \gets \text{permute}(x) \in \mathbb{R}^{(H/2 \cdot W/2) \times B \times D}$
\Statex \qquad $x \gets \text{TransformerEncoder}(x)$

\State {Project and reshape output}
\Statex \qquad $x \gets \text{permute\_back}(x) \in \mathbb{R}^{B \times (H/2 \cdot W/2) \times D}$
\Statex \qquad $x \gets \text{Linear}(x) \in \mathbb{R}^{B \times (H/2 \cdot W/2) \times M}$
\Statex \qquad $x \gets \text{reshape}(x) \in \mathbb{R}^{B \times M \times H/2 \times W/2}$

\State {Upsample to original resolution}
\Statex \qquad $x \gets \text{ConvTranspose2D}(x, \text{kernel}=3, \text{stride}=2, \text{padding}=1, \text{output\_padding}=1)$
\Statex \qquad $x \gets \text{Sigmoid}(x)$

\State \textbf{Output:} Tensor $x \in \mathbb{R}^{B \times M \times H \times W}$
\end{algorithmic}
\end{algorithm}

The key differences of our approach with respect to the original implementation are as follows:
\begin{itemize}
    \item \textbf{No tiling into patches:} We reduce spatial complexity through convolutional downsampling rather than tiling the input image into patches.
    \item \textbf{Stronger spatial inductive bias:} Whereas the original ViT implementation relies on positional embeddings to learn spatial relationships, we use both positional embeddings and convolutional downsampling.
    \item \textbf{Output type and granularity:} The original implementation outputs a vector of class logits, whereas our approach outputs a per-pixel, multi-channel 50$\times$50 image of class logits.
    \item \textbf{2D positional embeddings:} We use learnable 2D row \& column positional embeddings, vs. ViT's learnable 1D indexed embeddings. This does not represent a significant difference in implementation, but is noted here for completeness.
\end{itemize}

The important similarities between our approach and the original ViT implementation include:
\begin{itemize}
    \item \textbf{Layer stacking:} Both apply stacked transformer encoder layers which use attention and MLP blocks with LayerNorm and residual connections. In addition, both models only apply transformer encoders, no decoders.
    \item \textbf{Global multi-head self-attention:} Both model long-range dependencies across spatial tokens.
    \item \textbf{Linear embeddings:} Input representations are projected linearly into an embedding space, with dimension consistent across transformer layers.
    \item \textbf{Additive positional embeddings:} Instead of concatenating learned positional embeddings, both implementations simply add them to the input representations.
\end{itemize}

In the baseline development using our implementation of ViT, we swept learning rate, batch size, hidden size, number of encoder heads, and number of layers for the same 3 random seeds as in the ResNet implementation. We repeated this sweep using an L2 loss in place of the per-pixel binary cross-entropy, and found no significant difference in performance. In addition, we ran a similar experiment with concatenated positional embeddings and non-learned embeddings, and found that these either degraded or did not impact model performance. The optimal baseline had a batch size of 128, hidden size of 128, 8 encoder heads, and 4 layers. For more details on hyperparameter sweeps, see Appendix~\ref{app:adamopt}.

\section{Geostatistical baselines: Kriging}
\label{app:kriging}

In this work, we compared ResNet-backed M3 against a technique known as kriging, or Gaussian-process regression, as a geostatistical baseline. The major challenge with transferring existing geostatistical approaches to the problem context of M3 is that incumbent methods are typically applied on smaller scales where more granular data can be collected, e.g. for block models in mine planning (a concise review of which can be found in Section 4 of~\cite{poniewierski2019block}). These methods are not obviously suitable for global scale or for situations where data are a collection of true positives represented as binary flags. First, we list the 3 clear candidates and explain their challenges in the context of M3:

\begin{itemize}
    
\item \textbf{Inverse-distance weighting:} Perhaps the simplest geostatistical approach to interpolate missing data. For M3, it generates a trivial prediction because it requires at least partial true negative information. On a “training” dataset of only true positives represented by a boolean flag (i.e. +1 for the $z$'s), it will predict only positives (+1 for all $\hat{z}$'s in the masked region)~\cite{shepard1968two}.

\item \textbf{Multilinear regression:} From the machine learning perspective, a naive implementation has a high-dimensional input space (i.e. 25K independent variables). One could instead consider individual pixels or contiguous pixel groups to reduce the dimensionality, regress, and then impute in masked regions, but this starts to look equivalent to a convolutional architecture with linear activation functions, which would likely be an unfair comparison.

\item \textbf{Kriging:} This method is used in geospatial mapping and other areas of science generally, but was originally developed for prospecting~\cite{cressie1990origins}. In its most basic form it requires spatial stationarity and isotropy of the data~\cite{oliver2015basic}. In short, it involves a weighted average of mineral observations within a region. The weights are data-derived and are selected to minimize prediction variance in the interpolated (here, masked) region. The “training” required is to fit a curve to the variance of the data as a function of distance to a given point (thus the requirements for isotropy and at minimum second-order stationarity). The form of this curve is determined on a per-problem basis. The approach is computationally slow, and both stationarity and isotropy are not necessarily reasonable assumptions at the spatial scales we considered for this analysis.
\end{itemize}
With these considerations in mind, we treat kriging as a geostatistical baseline and develop the closest reasonable analog, structuring the training and evaluation processes so that experiments and metrics align as closely as possible with those in the analyses of M3. Below we present the key definitions appropriate to this approach, as context for the results in Figure~\ref{fig:prior-work}.\\[1em]

\noindent\textbf{Sparse Variational Multitask Gaussian Process Classification} (SVMGPC). We used gpytorch v1.14 to build the model for this baseline, see Listing~\ref{listing:ourcode}. Here, we explain the underlying mechanics. Each $50\text{ mi}\times 50\text{ mi}$ tile studied in the main body of the paper is stored as  
$\mathbf X\in\mathbb R^{C\times 50\times 50}$, with $C=10+N_{\text{cov}}+2$, including the following channels:

\begin{itemize}
\item $10$ sparse \emph{binary} mineral–presence layers indicating the flagged locations of previously-surveyed minerals, 
\item $N_{\text{cov}}$ continuous covariates (e.g. fault, geological age, and elevation),
\item Longitude and latitude (WGS–84).
\end{itemize}

Note the immediate difference with M3, which does not include global geolocation layers and only relies on spatially local features to perform inference. The masking process is identical to M3's, and the covariates and coordinates are never masked. The input to the Multitask Gaussian Process (MGP) is the masked $\mathbf{X}$. We model the target space of the MGP $\mathcal{GP}$ as being the 10 unmasked mineral layers jointly so that it has 10 tasks:
\[
\hat{z}_k(\mathbf x)\sim\mathcal{GP}\!\left(\text{mean }=\mu_k,\;\text{Cov}\left(\hat{z}_k(\mathbf{x},\hat{z}_{k'}(\mathbf{x'}\right)=
      \sum_{p=1}^{10}B_{kk'p}\,\kappa_p(\mathbf x,\mathbf x')\right),\qquad
      k,k'=1\dots 10,
\]
where $B_{kk'p} \in \mathbb{R}^{10\times 10 \times 10}$ is an optional positive–definite ``co-regionalization'' matrix which can be used to model correlations between each pair of mineral layers, and
$\kappa$ is a Radial Basis Function (RBF) kernel:
\[
\kappa_k(\mathbf x,\mathbf x’)
= \exp\!\Bigl(-\tfrac12(\mathbf x-\mathbf x’)^\top\Theta_k^{-2}(\mathbf x-\mathbf x’)\Bigr),
\quad
\Theta_k = \mathrm{diag}(\sigma_{k,1},\dots,\sigma_{k,C}).
\]
In the training, $\mu_k$, $B_{kk'p}$, and $\sigma_{k,i}$ are learned parameters, but can be parameterized. The gpytorch example in Listing~\ref{listing:apicode} defines $B_{kk'p} = a_{kp}a_{k'p}$ and $\sigma_{k,i} = \sigma_k \forall i$. In the results we compare M3 against, we used $B_{kk'p} = a_{kp}^2 \delta_{kk'}$ and $\sigma_{k,i} = \sigma_k \forall i$, as we generally observed that this simpler choice both trained more efficiently and performed better (explained below).

For $N_{tr}$ training points, the joint prior of $\hat{z}_k(\mathbf x)$ has covariance matrix of dimension $10N_{tr}\times 10N_{tr}$, which is prohibitively large to invert on a reasonable timescale as would be necessary to perform direct regression. To circumvent this limitation we apply the sparse variational trick, which is to approximate the true posterior distribution $p(\hat{z}|z)$ using a much smaller number $E \ll N_{tr}$ of ``inducing'' points $\mathbf \{\mathbf{e}_j\}_{j=1}^E$ whose locations are instantiated by k-means and learned during the optimization. To do so, we define:
    \begin{align}
        \hat{\mathbf{z}} &\equiv [\hat{z}_k(\mathbf{x}_n)], & k=1,\ldots,10\,\text{ and }\, n=1,\ldots,N_{tr}.\\        
        \mathbf{u} &\equiv [\hat{z}_k(\mathbf{e}_m)], & k=1,\ldots,10\,\text{ and }\, m=1,\ldots,E.
    \end{align}
We then perform the following estimate of the true posterior distribution $p(\hat{\mathbf{z}}|\mathbf{z})$ by learning it over the inducing points and then transferring it to the full training dataset $\hat{\mathbf{z}}$ by interpolating with respect to that derived over $\mathbf{u}$:
    \begin{equation}
        q(\hat{\mathbf{z}}) = \int p(\mathbf{\hat{z}} | \mathbf{u}) q(\mathbf{u}) d\mathbf{u}
    \end{equation}
This requires the following density functions to be either derived or specified. First, we posit:
    \begin{equation}
        q(\mathbf u)
        =\mathcal N\bigl(\boldsymbol\mu,\mathbf S\bigr)
        \quad\text{over the \(10E\)-vector \(\mathbf u\)}.
    \end{equation}
Then, we derive from the factorization of the joint pdf $p(\hat{\mathbf{z}},\mathbf{u})$:
    \begin{align}
        p(\mathbf u)&=\mathcal N\bigl(\mathbf 0,\;K_{ZZ}\bigr)\\
        p(\hat{\mathbf{z}} | \mathbf u)
&=\mathcal N\Bigl(
K_{XZ}K_{ZZ}^{-1}\,\mathbf u,\;
K_{XX}-K_{XZ}K_{ZZ}^{-1}K_{ZX}
\Bigr)
    \end{align}
Where the Gram matrices $K$ are determined by the structure of the $\mathcal{GP}$ as:
    \begin{align}
        [K_{ZZ}]_{(k,m),(k’,m')}
            &= B_{k k’}\;\kappa(\mathbf e_m,\mathbf e_{m'}),\\
        [K_{XZ}]_{(k,n),(k’,m)}
            &= B_{k k’}\;\kappa(\mathbf x_n,\mathbf e_m),\\
        [K_{XX}]_{(k,n),(k’,n’)}
            &= B_{k k’}\;\kappa(\mathbf x_n,\mathbf x_{n’}).
    \end{align}
Note that the estimate of the true posterior $q(\hat{\mathbf{z}})$ has a closed form with these definitions. We learn $\boldsymbol{\mu}, \mathbf{S}, \mu_k, B_{kk'},$ and the $\Theta_k$. Our effective covariance matrix then has the more manageable dimension $10E\times 10E$. In these terms, we optimize the estimate of the true posterior by maximizing the evidence lower bound (ELBO) on the marginal likelihood of our ground-truth data $z_{n,k}\in\{0,1\}$. To do so, we use a Bernoulli profile for the likelihood:
    \begin{equation}
        p\bigl(z_{n,k}=1 \mid \hat{z}_k(\mathbf x_n)\bigr)
        = \sigma\bigl(\hat{z}_k(\mathbf x_n)\bigr),
        \quad
        \sigma(f)=\frac{1}{1+e^{-f}}.
    \end{equation}
And the ELBO loss which we minimize is
    \begin{equation}
        \mathcal L =
        -\sum_{n=1}^{N_{tr}}
        \sum_{k=1}^{10}
        \mathbb E_{q(\hat{z}_k(\mathbf x_n))}\!\Bigl[
        \log \mathrm{Bern}\bigl(z_{n,k}\mid\sigma(\hat{z}_k(\mathbf x_n))\bigr)
        \Bigr]
        \;-\;
        \text{KL}\!\bigl[q(\mathbf u)\;\Vert\;p(\mathbf u)\bigr]
    \end{equation}
Note that because both $p(\mathbf{u})$ and $q(\mathbf{u})$ are Gaussians in $\mathbb{R}^{10E}$, the KL divergence has a readily-derived closed form.

We use an AdamOptimizer to maximize the ELBO stochastically, in minibatches. Inducing centers $\{\mathbf{e}_m\}$ are initialized via k-means++ on latitude and longitude. We set `learn\_inducing\_locations=True' in the VariationalStrategy object so that these inducing centers are further optimized during training. For evaluation, the outputs of the model are taken as $\sigma\left(\hat{z}_k(\mathbf x_n)\right) > T$ for threshold probability $T$.

\paragraph{Threshold Variations.} To account for the possibility that the MGPC performance heavily depends on the binary classification threshold $T$, due to e.g. class imbalances in the data, for computation of the Dice coefficient we considered every threshold $T\in\{0.01,0.1,0.2,0.3,\ldots,0.9,0.99\}$ and compared the maximum against M3's performance. $T=0.5$ maximized the validation Dice coefficient in nearly all runs, with the exception of 1 seed used in the final results, which had an optimal $T=0.4$.\\

\noindent\textbf{Deduplication of Training Data.} Because tiles overlap geographically in the training set, and the kriging approach is provided access to global coordinates, we deduplicate to avoid oversampling of instances during training. To keep one copy of each $1\text{ mi}\times1\text{ mi}$ cell, we hash every pixel center to a 64-bit key as follows:

\begin{equation}
h(\lambda,\varphi)=
\Bigl\lfloor\frac{\lambda}{\Delta\lambda(\varphi)}\Bigr\rfloor
\;\ll 32\;
+\;\Bigl\lfloor\frac{\varphi}{\Delta\varphi}\Bigr\rfloor,
\qquad
\Delta\varphi=\frac{1}{69^{\circ}},\qquad
\Delta\lambda(\varphi)=\frac{1}{69^{\circ}\cos\varphi}.
\end{equation}
and stream through the training set once prior to training, eliminating duplicates using Algorithm~\ref{alg:dedup}. Only mineral channels are zeroed; covariates and geolocations remain intact. Importantly, during evaluation, we do not perform deduplication so that we maintain comparability with performance metrics for M3.\\ 

\begin{algorithm}[t]
\caption{Streaming 1-mi duplicate removal}\label{alg:dedup}
\begin{algorithmic}
\State \textbf{input:} training tiles $\{\mathbf X^{(t)}\}_{t=1}^T$
\State $S\gets\varnothing$ \Comment{Hash–set of seen cells}
\For{$t=1$ \textbf{to} $T$}
    \State $\mathbf K\gets h(\mathbf X^{(t)}_{\text{lon}},
                             \mathbf X^{(t)}_{\text{lat}})$
    \State $\mathsf{dup}\gets\text{isin}(\mathbf K,S)$
    \State $ \mathbf X^{(t)}_{1:10}[\,\mathsf{dup}\,]\gets 0$
    \State $ S\gets S\cup\{\mathbf K[\lnot\mathsf{dup}]\}$
\EndFor
\State \textbf{return} cleaned tiles
\end{algorithmic}
\end{algorithm}

\noindent\textbf{Training Details and Hyperparameter Optimization.} We trained for 80 epochs each on 3 random seeds for multiple configurations, disabling coregionalization (setting $B_{kk'p} = a_{kp}^2 \delta_{kk'}$) to reduce the initial computational requirements and perform a basic initial sweep. Convergence was defined by saturation of the maximum-achieved Dice coefficient on the validation data, and was typically reached in $40\text{–}60$\,epochs for the largest $E=512$. We ran a grid sweep over learning rate ($5~\times~10^{-2}, 10^{-2}, 10^{-3}$) and batch size ($10, 60$), using the largest number of inducing points which would fit in VRAM, in this case $E=512$. The optimal configuration was a learning rate of $10^{-3}$ and a batch size of $10$ for the in-distribution tests, and a learning rate of $10^{-3}$ and batch size of $60$ for the OOD tests.

We then tested the incorporation of coregionalization (i.e. $B_{kk'p} = a_{kp} a_{k'p}$) for the best performing model from this sweep, repeating the training on the same 3 random seeds. We found a significant performance reduction despite doubling the computation time required, and did not consider this approach further.\\

\begin{lstlisting}[language=Python, caption={GPyTorch SVGP Multitask Example from v1.14 API Documentation}, label=listing:apicode]
num_latents = 3
num_tasks   = 4

class MultitaskGPModel(gpytorch.models.ApproximateGP):
    def __init__(self, num_latents, num_tasks):
        # One set of inducing points per latent function.
        # Randomly initialized, not built with k-means++.
        inducing_points = torch.rand(num_latents, 16, 1)

        # Variational distribution: one per latent
        variational_distribution = 
        gpytorch.variational.CholeskyVariationalDistribution(
            inducing_points.size(-2),
            batch_shape=torch.Size([num_latents])
        )

        # Base variational strategy (per latent)
        base_vs = gpytorch.variational.VariationalStrategy(
            self, inducing_points, variational_distribution,
            learn_inducing_locations=True
        )

        # Wrap in LMC to mix latents into tasks
        variational_strategy = gpytorch.variational.LMCVariationalStrategy(
            base_vs,
            num_tasks=num_tasks,
            num_latents=num_latents,
            latent_dim=-1
        )

        super().__init__(variational_strategy)

        # One mean & kernel per latent
        self.mean_module = gpytorch.means.ConstantMean(
            batch_shape=torch.Size([num_latents])
        )
        self.covar_module = gpytorch.kernels.ScaleKernel(
            gpytorch.kernels.RBFKernel(batch_shape=torch.Size([num_latents])),
            batch_shape=torch.Size([num_latents]]
        )

    def forward(self, x):
        mean_x  = self.mean_module(x)   # (num_latents, N)
        covar_x = self.covar_module(x)  # (num_latents, N, N)
        return gpytorch.distributions.MultivariateNormal(mean_x, covar_x)

model      = MultitaskGPModel(num_latents, num_tasks)
likelihood = gpytorch.likelihoods.MultitaskGaussianLikelihood(num_tasks=num_tasks)
loss = -1 * gpytorch.mlls.VariationalELBO(likelihood, 
    model, 
    num_date=len(training_data))
\end{lstlisting}
\vspace{1cm}
% \label{listing:ourcode}
\begin{lstlisting}[language=Python, caption={MGPC Implementation Used for Baseline}, label=listing:ourcode] 
class MGPC(gpytorch.models.ApproximateGP):
    def __init__(self, inducing, num_tasks):
        # Variational distribution: one per task
        q = gpytorch.variational.CholeskyVariationalDistribution(
            inducing.size(0),
            batch_shape=torch.Size([num_tasks])
        )

        # Base variational strategy (inducing shared across tasks)
        vs = gpytorch.variational.VariationalStrategy(
            self, inducing, q, learn_inducing_locations=True
        )

        # IndependentMultitask: one GP per task, no mixing latents
        mts = gpytorch.variational.MultitaskVariationalStrategy(
            vs,
            num_tasks=num_tasks
        )

        super().__init__(mts)

        # One mean & kernel per task: set \mu_k
        self.mean_module = gpytorch.means.ConstantMean(
            batch_shape=torch.Size([num_tasks])
        )
        # Set B_{kk'} = 1:
        self.covar_module = gpytorch.kernels.ScaleKernel(
            gpytorch.kernels.RBFKernel(
                batch_shape=torch.Size([num_tasks]]
            ),
            batch_shape=torch.Size([num_tasks]]
        )

    def forward(self, x):
        return gpytorch.distributions.MultivariateNormal(
            self.mean_module(x),
            self.covar_module(x)
        )

model = MGPC(inducing_points, num_tasks=10)
likelihood = gpytorch.likelihoods.BernoulliLikelihood()
loss = -1 * gpytorch.mlls.VariationalELBO(likelihood, 
        model, 
        num_data=len(training_data))
\end{lstlisting}

\section{Machine learning experiments: Optimizer details}
\label{app:adamopt}

In all machine learning experiments in this paper, we used an AdamOptimizer to perform stochastic gradient descent on minibatches whose size we treated as a hyperparameter and swept between $\{32, 64, 128,256\}$. As we sought a balance between model and optimizer hyperparameters in the training sweeps, we focused our efforts on learning rate and batch size, and did not optimize the $\beta_1$ or $\beta_2$ hyperparameters. We selected $\beta_1 = .9$ and $\beta_2 = .999$, and swept the learning rate between $\{10^{-4},5* 10^{-3},10^{-3},10^{-2}\}$, finding an optimal learning rate of $10^{-3}$ and batch size 128 for the ResNet-backed models. For the transformer models we again found an optimal learning rate of $10^{-3}$ and batch size of $128$.

To understand the impact of the choice of loss function on our results, we performed similar scans with a per-pixel L2 loss and a composite per-pixel L2 with integral L2 loss. The BCE loss generally outperformed these alternatives. Along with loss function considerations, we explored the na\"ive approach of simply excluding predicted negatives over the masked region from the loss calculation, but doing so led to unstable training, and we were unable to regress any performant model with this configuration (which is unsurprising, as this causes the cross-entropy to be non-convex, but we felt it important to check regardless). 

Architecturally, for M3 we varied the RaCA embedding size $K$ within $\{16,32,64,126\}$ and tested different UNet backbones from the Segmentation Models pytorch package (ResNet50, ResNet101, ResNet152)~\cite{Yakubovskiy:2019}. For the ViT experiments, we swept the number of encoder heads within $\{4,8,12,16\}$, the number of transformer encoder layers within $\{1,2,4\}$, and the hidden layer size $H$ within $\{64,128,256,512\}$. The optimal baseline had $H=512$ for $16$ heads and $1$ transformer encoder layer with a per-pixel L2 loss.

For the geostatistical baseline outlined in Appendix~\ref{app:kriging}, we use a stochastic method with an AdamOptimizer similarly having $\beta_1=.9$, $\beta_2=.999$. We swept over learning rates of $\{0.05,0.01,0.001\}$ and batch sizes of $\{10,60\}$. In-distribution experiments had optimal learning rate of $0.001$ and batch size of $10$. OOD experiments had optimal learning rate of $0.001$ and batch size $60$. Additional architectural variations were considered, and in general we designed this baseline to have every possible advantage relative to the ML approaches, including passing in global coordinates as an input feature layer. The variational approach requires specifying one hyperparameter $E$, the number of ``inducing points.'' In general, it is advisable to make this parameter as large as will fit in VRAM (and we did so) but we also observed through a preliminary check in the ID case that setting $E\in\{64,128,256\}$ leads to reduced performance.

All experiments described in this section were conducted across at least $3$ fixed random seeds.

\section{Model performance versus masking aggressiveness}
\label{app:aggro}

\begin{figure}[h!]
    \begin{adjustbox}{center}
        \includegraphics[width=0.7\linewidth]{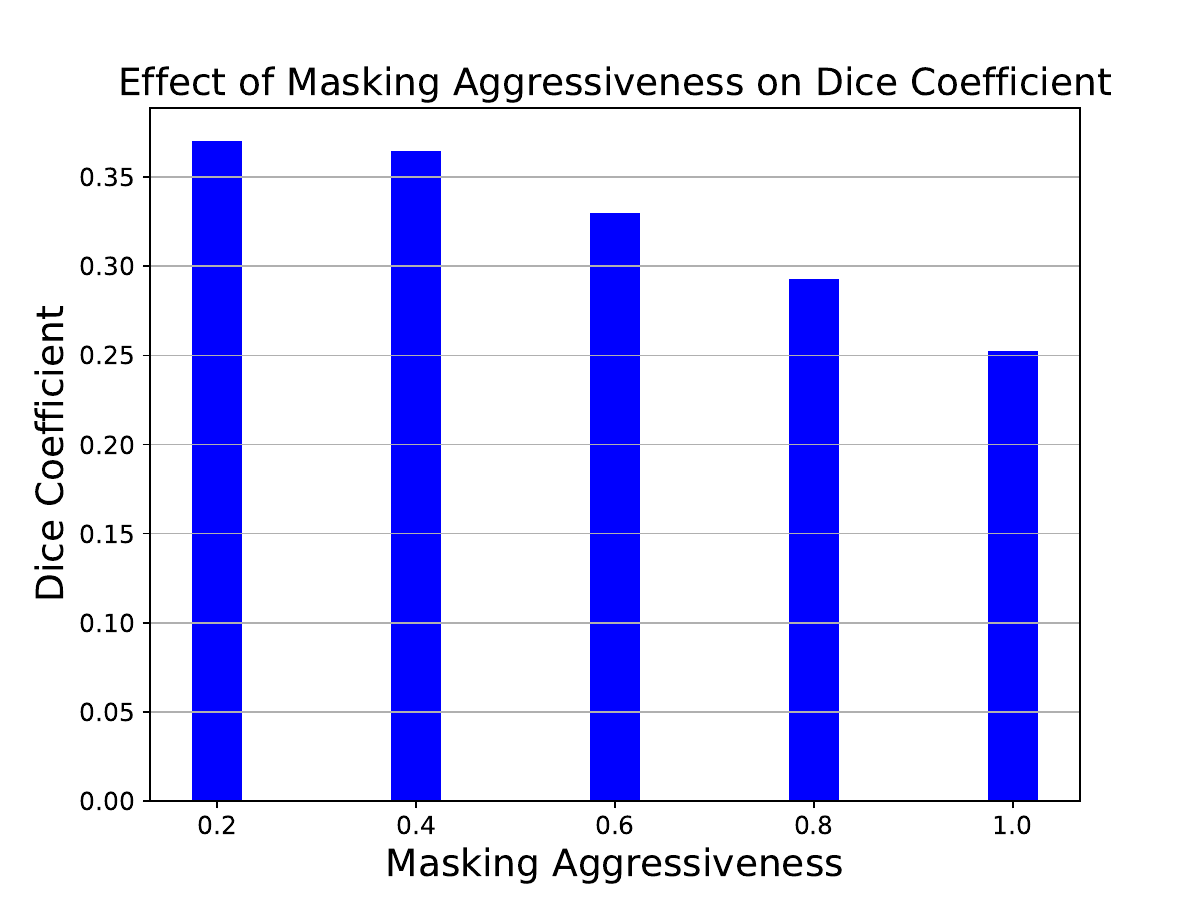}
    \end{adjustbox}
    \caption{Bar chart showing the effect of varied test-time masking aggressiveness $A$ on the dice coefficient. The higher the aggressiveness, the more input content can be hidden from the model during prediction. As $A$ increases from 0.2 to 1, the Dice coefficient decreases, reflecting the increased difficulty of the prediction task.}
    \label{fig:masking-chart}
\end{figure}

As a sanity check on our masking approach, we performed a sweep on the test-time masking aggression using a model trained on only mineral layers as inputs, with $N=10$K, $H=152$, and train-time $A=0.8$. As shown in Figure~\ref{fig:masking-chart}, we find a drop in performance as more of the data is ablated, as expected.

\section{Future direction: Simultaneous regression of masking and mapping}
\label{app:regmask}

In the main results of this work, we fixed $A=0.8$ for the experiments in the main body and evaluated the impact of $A$ on performance in Appendix~\ref{app:aggro} with a basic 1D sweep. More generally, there are many other parameters one might consider tuning within the masking strategy, such as the relative probabilities of mineral vs. spatial masking. In addition, one could consider completely random masking, or masking based on labels determined by e.g. unsupervised segmentation of the data. Relatedly, it is worth considering whether \textit{any} masking is necessary at all, as one could simply learn to predict the presence or absence of a mineral from e.g. geophysical and agronomic features alone (although our ability to predict mineral presence from copresence features makes clear that existing mineralogical records are a valuable feature for such a task).

We developed M3 as a baseline with a more general alternative in mind, which addresses all of these concerns while relaxing some of the non-generic assumptions within our approach. The general procedure involves learning a masking profile simultaneously with the infilling task, i.e. Simultaneous Regression of Masking and Mapping (SRMM). The input space $\mathcal{X}$ is identical to M3's, and the proposed algorithm contains 3 inter-operating modules:
\begin{itemize}
    \item An \textit{infilling} or \textit{recovery} M3 module M3$_R$, as developed in this paper, which estimates the true underlying distribution of minerals from some current best-known sampling.
    \item A masking module $\Phi$ which mimics any lossiness or biases in the record curation process.
    \item A second \textit{discovery} M3 module M3$_D$ which is trained only on completely masked outputs, i.e. without mineral layers. (Note that this implies the auxiliary features are a prerequisite to developing this approach).
\end{itemize}

\begin{figure}[ht!]
\begin{adjustbox}{center}
\begin{tikzpicture}[>=Latex, font=\sffamily]

% Node style for rectangles with rounded corners and fill color
\tikzset{
    state/.style={
        rectangle, draw, rounded corners, minimum width=2.8cm, minimum height=1.2cm,
        align=center, font=\Large, fill=orange!40
    },
    labeltext/.style={
        font=\footnotesize, align=center
    },
    arrowstyle/.style={
        ->, line width=1pt
    }
}

% Positions of the four boxes
\node[state, fill=yellow!4] (A) at (0, 4) {$A$};
\node[state] (B) at (5, 4) {$B$};
\node[state, fill=orange!15] (C) at (0, 1) {$C$};
\node[state] (D) at (5, 1) {$D$};

% Descriptive labels above/below each box
\node[labeltext, above=2pt of A] {Undiscovered, fully masked};
\node[labeltext, above=2pt of B] {Complete mineralogical record};
\node[labeltext, below=2pt of C] {Imperfectly-maintained record};
\node[labeltext, below=2pt of D] {Recovered record};

% Arrows between boxes
\draw[arrowstyle] (A) -- (B) node[midway, above] {\textrm{M3}$_D(A)$};
\draw[arrowstyle] (B) -- (C) node[midway, right, xshift=7pt, yshift=-3pt] {$\Phi(B)$};
\draw[arrowstyle] (C) -- (D) node[midway, below] {\textrm{M3}$_R(C)$};

% Additional equations on the right-hand side
\node[align=left, anchor=west] at (7.5, 3.5) {$\mathcal{L}_{\textrm{M3}} =   \textrm{BCE}\left(\textrm{M3}_R\Big(\Phi\big(\textrm{M3}_D(A)\big)\right),\, \{\textrm{M3}_D(A) > T\}\Big)$};
\node[align=left, anchor=west] at (7.5, 2.5) {$\mathcal{L}_\Phi = \textrm{BCE}\left(\Phi\left(\textrm{M3}_D(A)\right),\{z^{(i)}_{jk\ell}\}\right)$}; 
\node[align=left, anchor=west] at (7.5, 1.5) {$\mathcal{L} = \mathcal{L}_{\textrm{M3}} + \beta \mathcal{L}_{\Phi}$};

\end{tikzpicture}
\end{adjustbox}
\caption{An algorithm for simultaneous regression of masking and mapping (SRMM), which we propose as a future research direction. It builds on M3 to regress a data-driven masking procedure $\Phi$ using two M3 instances M3$_D$ and M3$_R$.}
\label{fig:SRMM}
\end{figure}
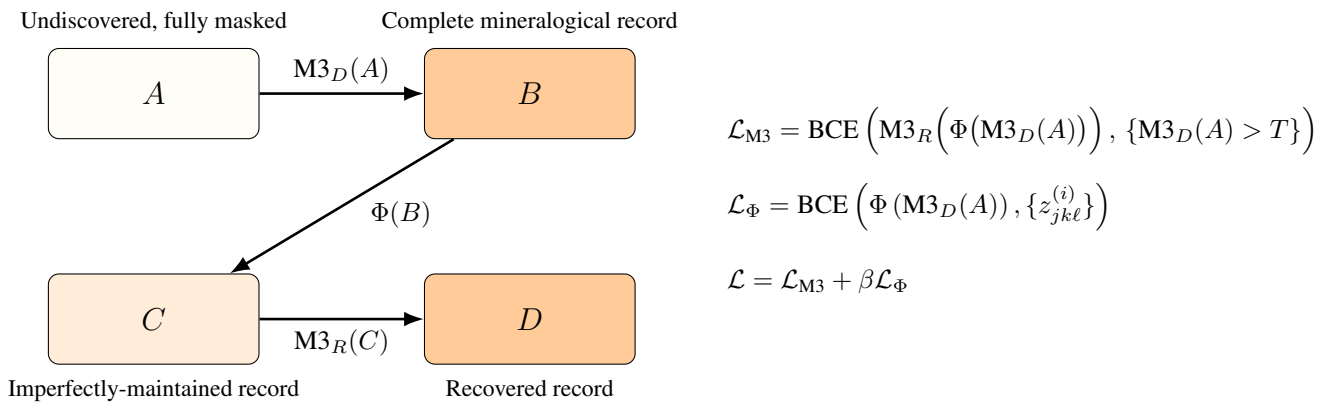

These models interact with each other to mimic the prospecting process end-to-end. First, all records are uncovered without prior knowledge of other resources. Then, some are lost to predictable sampling bias. Lastly, we use M3 to approximate an inversion to the projection. Figure~\ref{fig:SRMM} depicts this process and compares it against ``vanilla'' M3.

The suggested architecture presents many interesting opportunities for experimentation, but is largely designed with the intention of regressing an interpretable $\Phi$ module whose behavior models the human biases which might impact mineralogical surveys. The model specification is designed to have the trivial mask $\Phi(B) = B$ as a global optimum when \textbf{(I)} a complete mineralogical record can be inferred from auxiliary data alone, and \textbf{(II)} the available set of prospecting records is complete. These conditions can both be met in the extreme scenario where (I) every location on the planet is directly sampled and (II) the complete set of those records is kept (i.e. the $z^{(i)}_{jk\ell}$ are assumed to be complete). One can confirm this by plugging the following functional form for $M3_R$ into the loss function in Figure~\ref{fig:SRMM} and showing that $\mathcal{L}\mapsto0$:
\begin{align}
    M3_R\left(\Phi\left(M3_D(A^{(i)}_{jk\ell})\right)\right) = \left[1-2\Theta\left(T - M3_D(A^{(i)}_{jk\ell})\right)\right]\Phi^{-\beta\cdot z^{(i)}_{jk\ell}}(1-\Phi)^{-\beta\cdot (1-z^{(i)}_{jk\ell})} + \Theta\left(T - M3_D(A^{(i)}_{jk\ell})\right)
\end{align}
Where $\Theta(x) = \frac{d}{dx}\rm{ReLU}(x)$. When $\Phi(B) = B$, i.e. the probability $\Phi$ of masking is $0$ at every location specified by $i,j,k,$ and $\ell$, this expression simply matches $M3_R$ to the decision boundary set by $M3_D$, which by assumption (I) is faithful to the complete set of records specified in assumption (II). We include in the Supplementary Materials a Mathematica notebook where we perform these calculations explicitly~\cite{Mathematica}. Due to the chaining of objectives involved and the subsequent potential for a volatile loss landscape, training SRMM will likely require careful optimization and may benefit from recent developments on the optimization of noisy loss landscapes which are bounded from below, e.g.~\cite{DeLuca:2022brp}.

\section{Additional context for Figure~\ref{fig:usa-performance}}
\label{app:extramaps}

\begin{figure*}[p]
    \begin{center}
        \textbf{Side-by-side comparison of ground truth vs. predictions in aggregate}
    \end{center}
    \begin{adjustbox}{center}
        \includegraphics[width=\linewidth, trim = 175pt 0pt 175pt 50pt, clip]{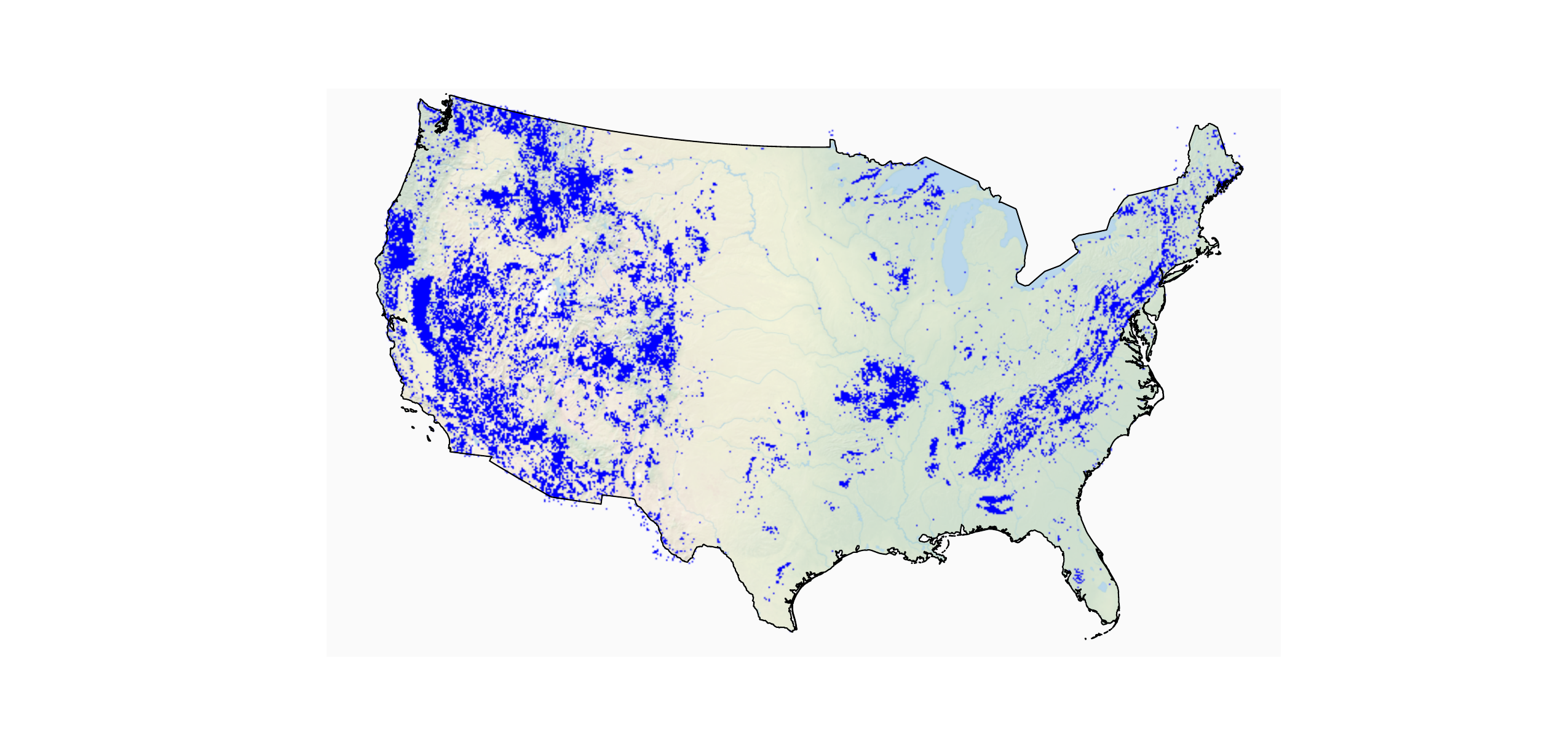}
    \end{adjustbox}
    \caption{Map of the ground-truth data used in the visualization dataset (EPSG:5070). A blue point is a binary flag indicating the presence of at least one resource within a 5mi$\times$5mi bin.}
    \label{fig:map-gt-countcov}
    
    \begin{adjustbox}{center}
        \includegraphics[width=\linewidth, trim = 175pt 0pt 175pt 0pt, clip]{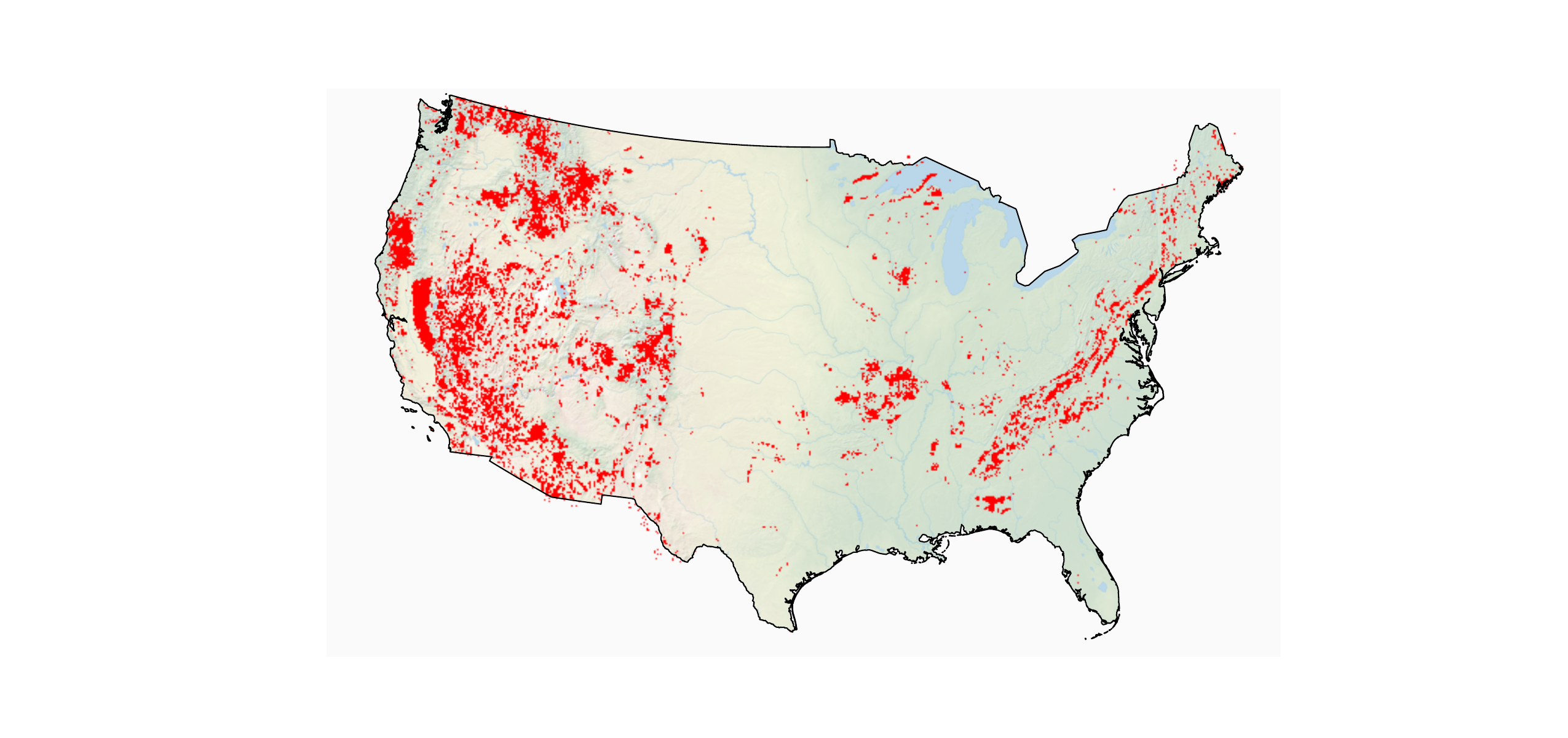}
    \end{adjustbox}
    \caption{The red component of Figure~\ref{fig:usa-performance}, representing the model outputs. A red point is a binary flag indicating the prediction of at least one resource within a 5mi$\times$5mi bin.}
    \label{fig:map-pred-countcov}
\end{figure*}

\begin{figure*}[p]

    \begin{center}
        \textbf{Dataset coverage maps}
    \end{center}
    \begin{adjustbox}{center}
        \includegraphics[width=\linewidth, trim = 175pt 60pt 175pt 0pt, clip]{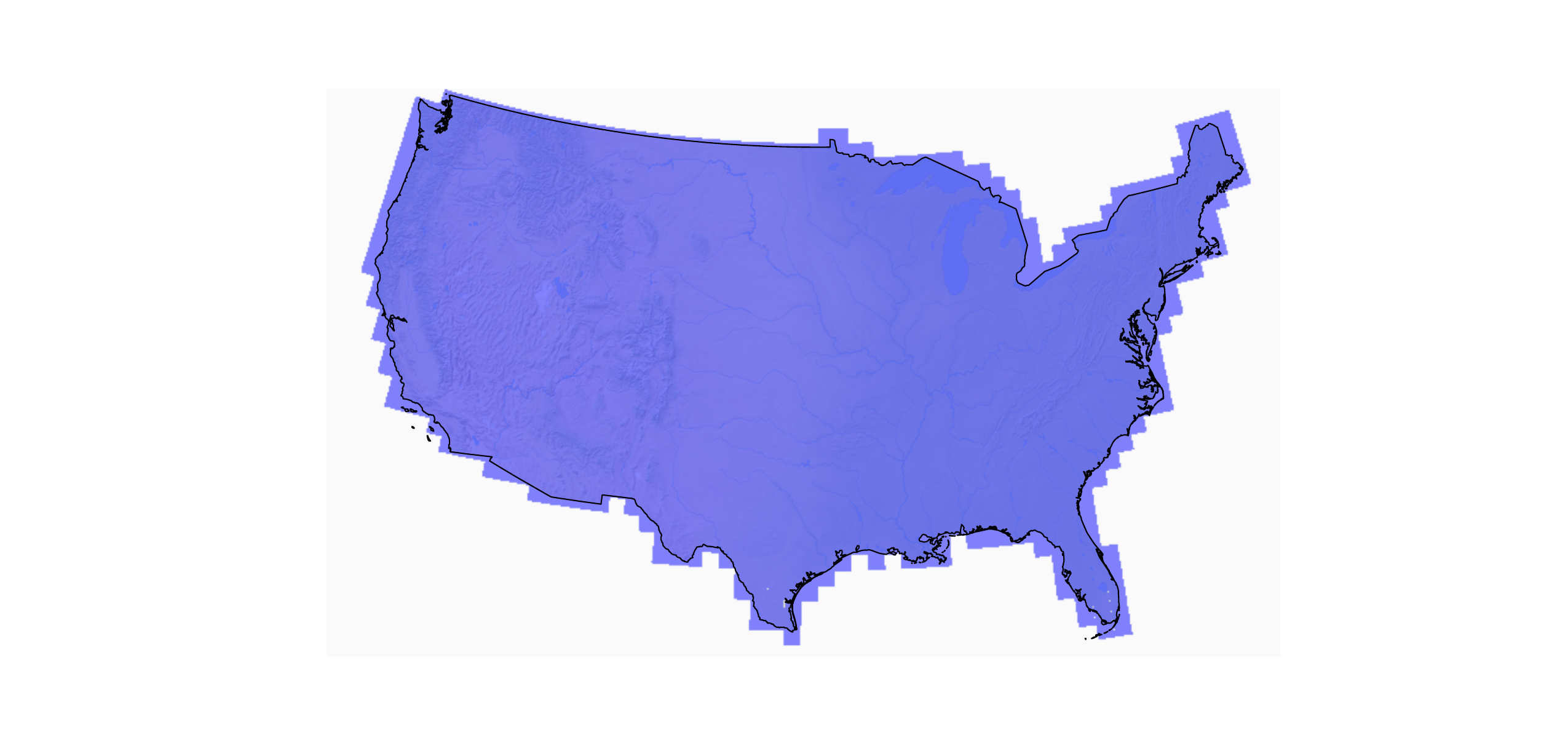}
    \end{adjustbox}
    \caption{Map of the coverage of tiles used in the visualization dataset (EPSG:5070). Blue indicates that at least one tile was included containing that region.}
    
    \begin{adjustbox}{center}
        \includegraphics[width=\linewidth, trim = 175pt 60pt 175pt 50pt, clip]{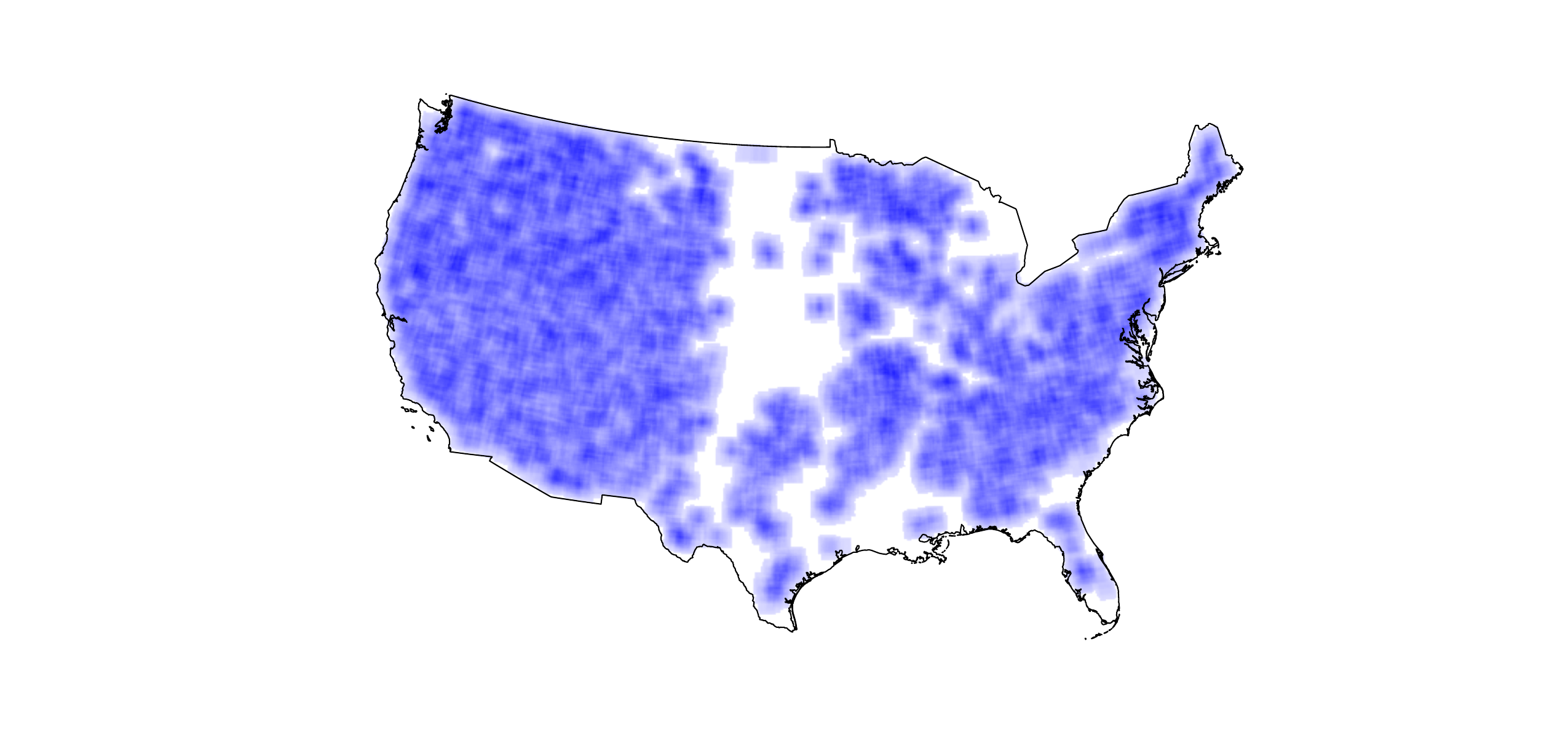}
    \end{adjustbox}
    \caption{Heatmap of the context windows in the training dataset for the model used to generate Figure~\ref{fig:usa-performance} (EPSG:5070). Darker blue regions indicate locations where multiple context windows overlapped.}
    \label{fig:map-gt-traincov}
    
\end{figure*}

\begin{figure*}[p]

    \begin{center}
        \textbf{OOD experimental configuration}
    \end{center}
    \begin{adjustbox}{center}
        \includegraphics[width=\linewidth, trim = 175pt 60pt 175pt 60pt, clip]{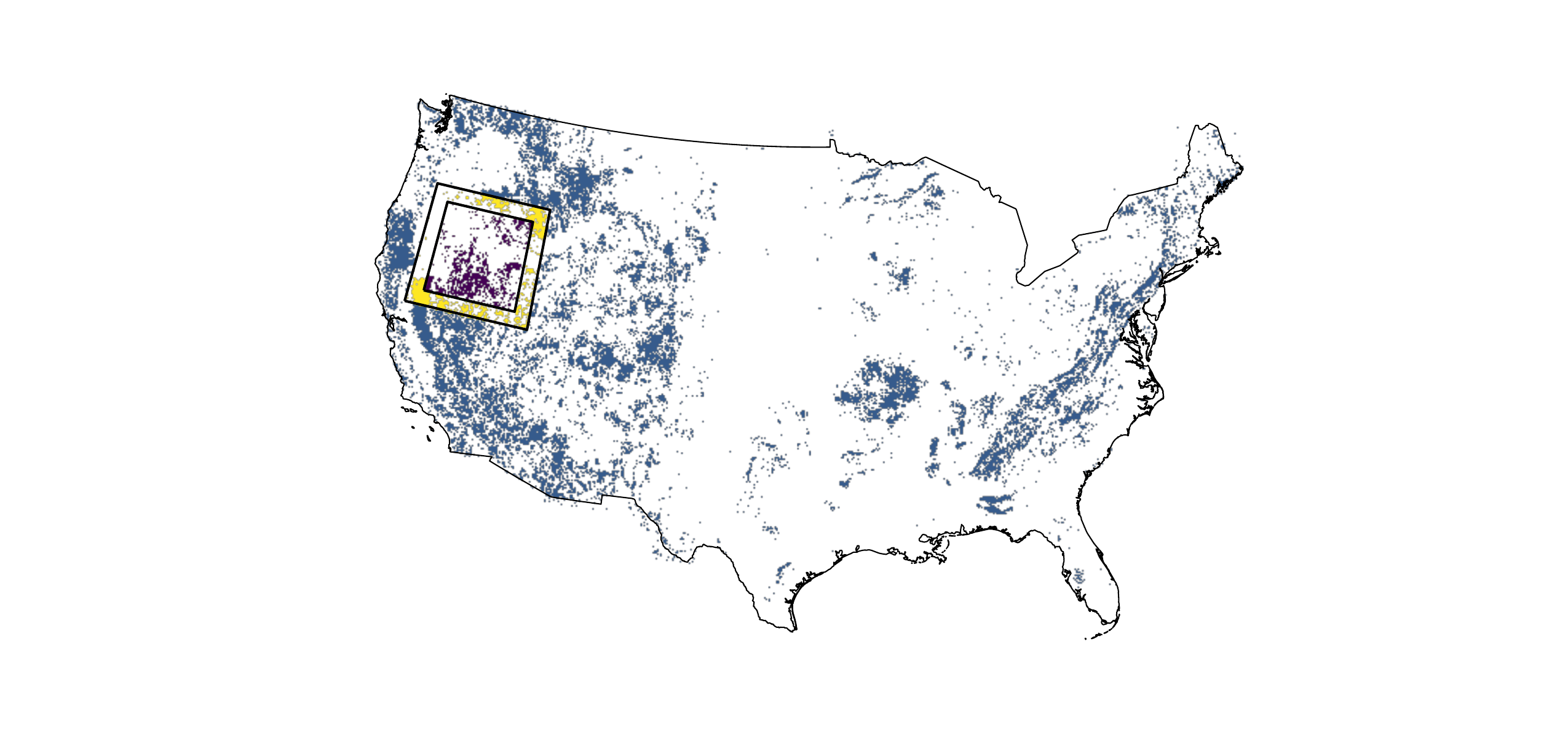}
    \end{adjustbox}
    \caption{Map of the ground-truth data used in the OOD experiments depicted in Figure~\ref{fig:prior-work}, with spatial delineations of training data (blue-green), validation data (yellow), and test data (purple).}
    \vspace{0.5cm}
    \begin{center}
        \textbf{Clustered predictions where no records exist in training data}
    \end{center}
    \begin{adjustbox}{center}
        \includegraphics[width=\linewidth, trim = 0pt 0pt 0pt 0pt, clip]{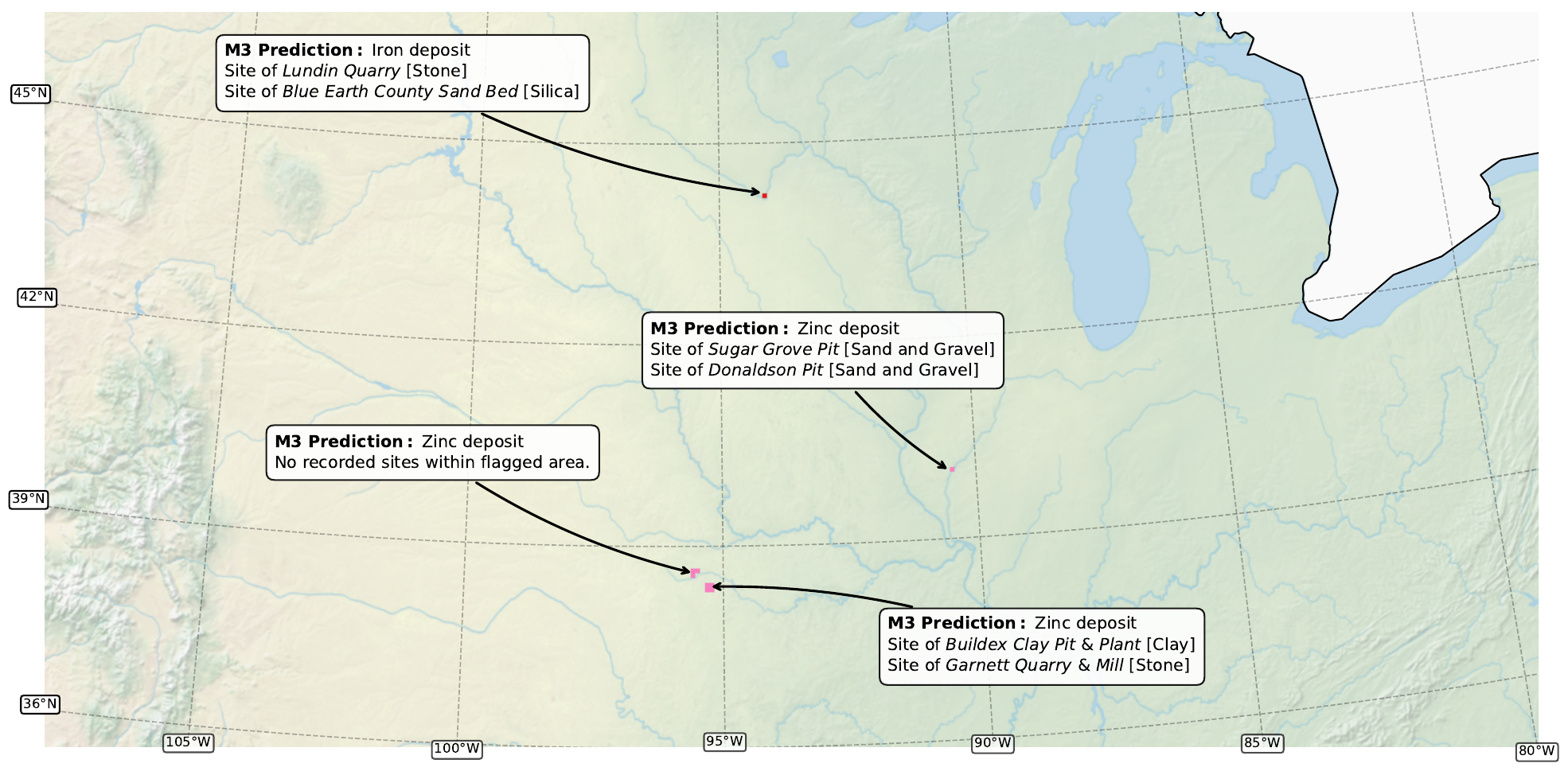}
    \end{adjustbox}
    \caption{A subset of the map in Figure~\ref{fig:map-pred-countcov} showing predictions of zinc and iron deposits on the US mainland where no records were available within the training data (i.e. they are spatially OOD with respect to the 10 minerals considered in the analysis). Clusters are labeled with matches to existing mining operations within the region wherever possible. One deposit appears to have zero matching records in the complete MRDS.}
    \label{fig:map-gt-no-records-prod}
    
\end{figure*}

\begin{figure*}[p]

    \begin{center}
                \textbf{Gold-specific ground-truth and predictions}
    \end{center}
    \begin{adjustbox}{center}
        \includegraphics[width=\linewidth, trim = 175pt 60pt 175pt 0pt, clip]{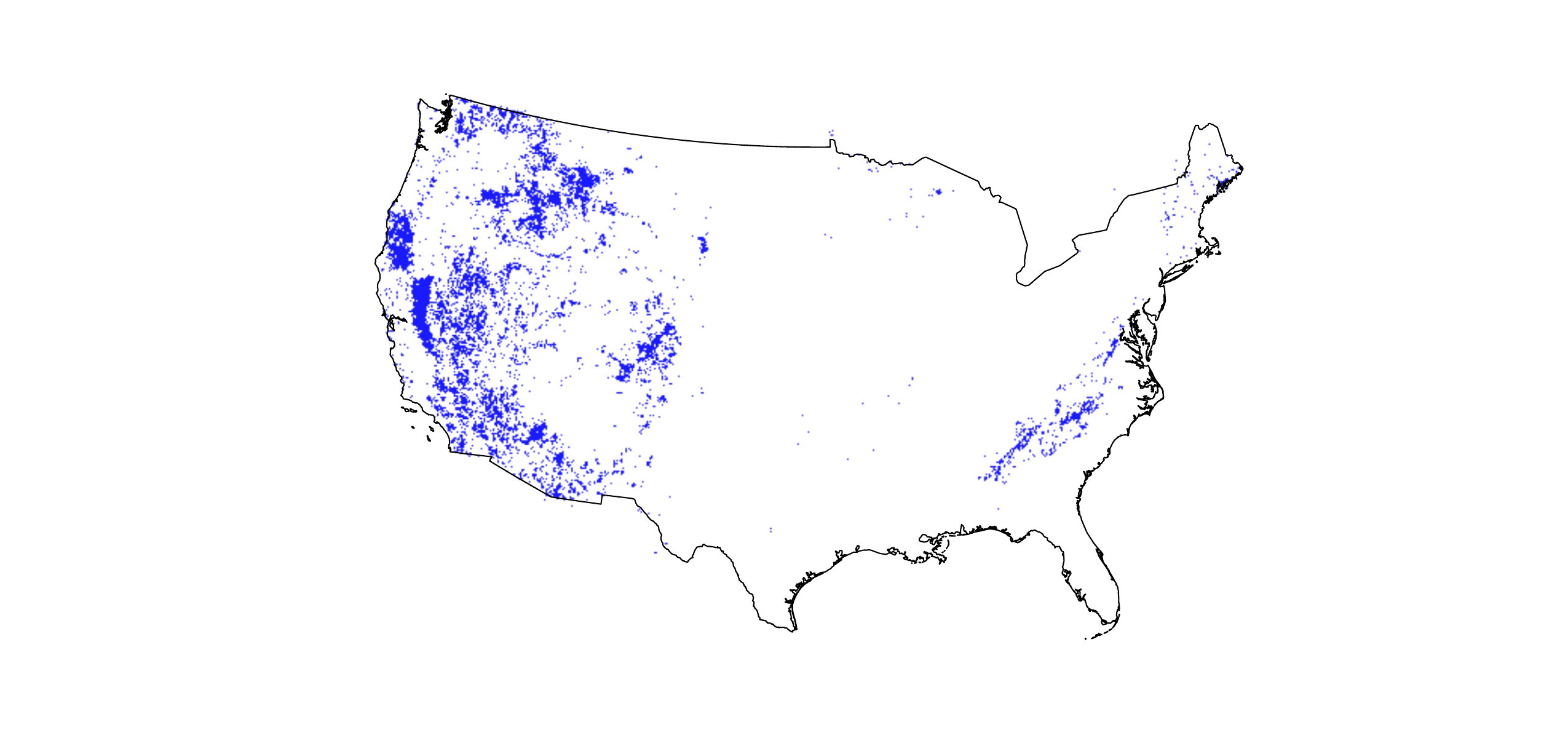}
    \end{adjustbox}
    \caption{Map of the ground-truth data used in the visualization dataset for gold (EPSG:5070). A blue point is a binary flag indicating the presence of at least one resource within a 5mi$\times$5mi bin.}
    
    \begin{adjustbox}{center}
        \includegraphics[width=\linewidth, trim = 175pt 60pt 175pt 50pt, clip]{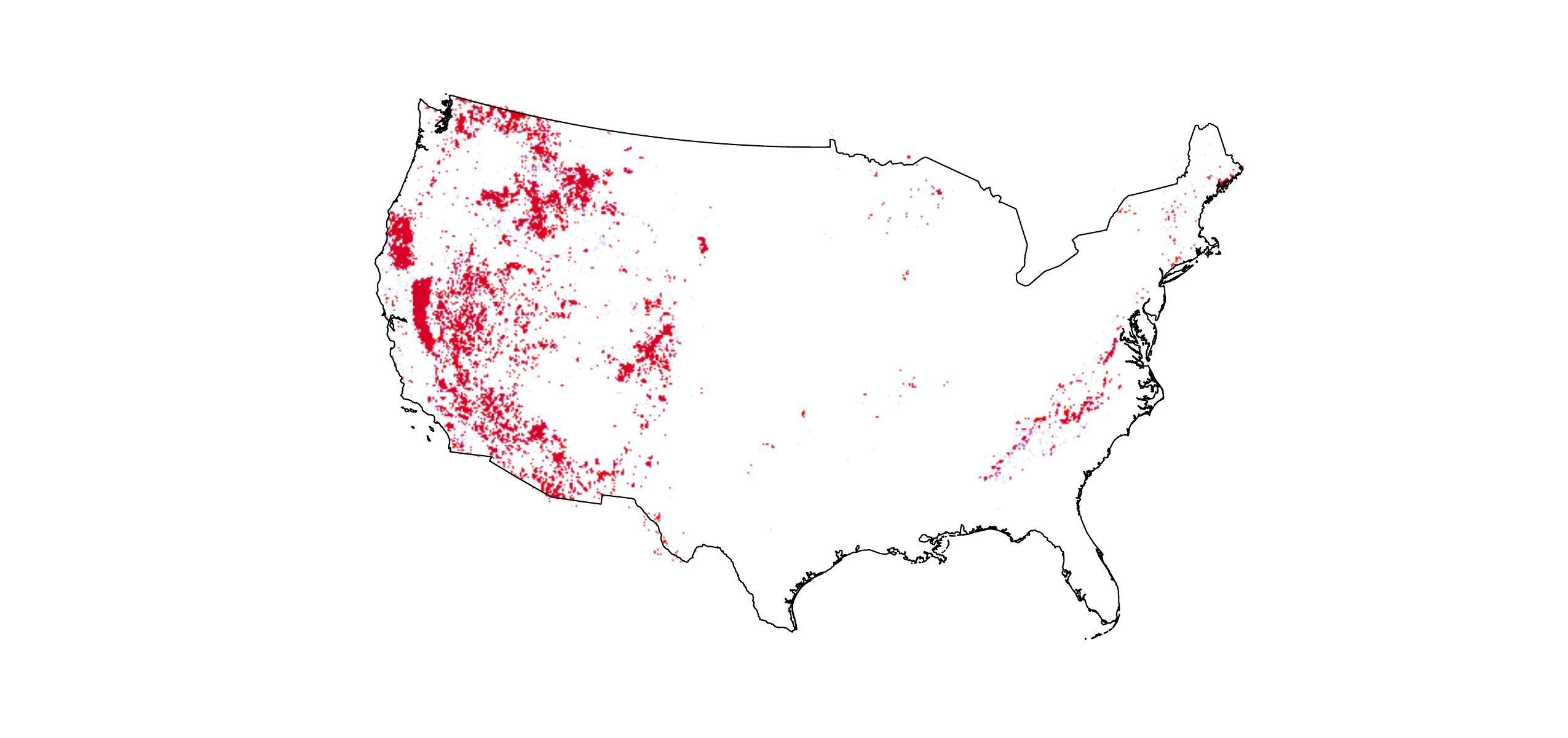}
    \end{adjustbox}
    \caption{Map of the predictions of the model for gold. A red point is a binary flag indicating the prediction of at least one resource within a 5mi$\times$5mi bin.}
    \label{fig:map-gt-pred-gold}
    
\end{figure*}

\begin{figure*}[p]

    \begin{center}
                \textbf{Silver-specific ground-truth and predictions}
    \end{center}
    \begin{adjustbox}{center}
        \includegraphics[width=\linewidth, trim = 175pt 60pt 175pt 0pt, clip]{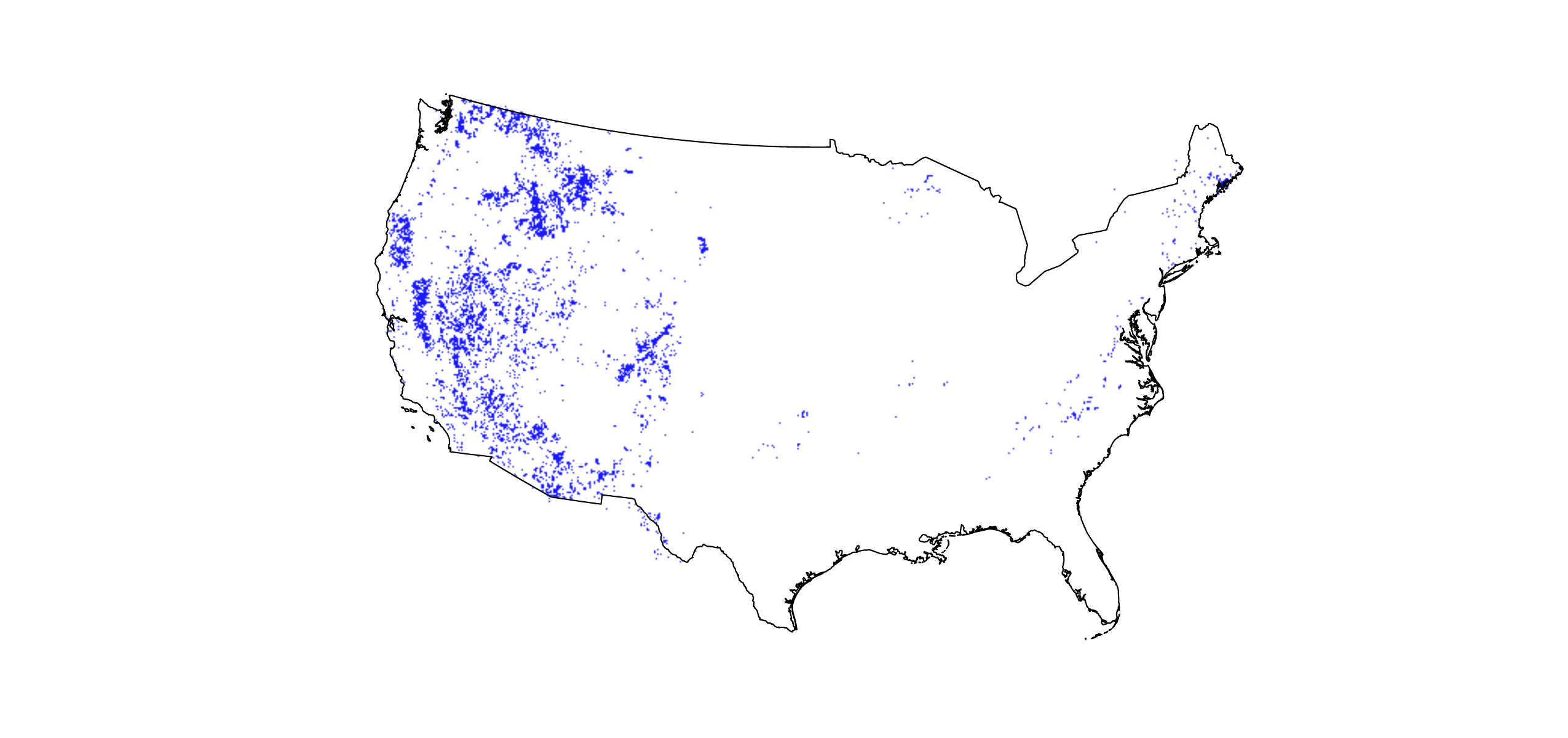}
    \end{adjustbox}
    \caption{Map of the ground-truth data used in the visualization dataset for silver (EPSG:5070). A blue point is a binary flag indicating the presence of at least one resource within a 5mi$\times$5mi bin.}
    
    \begin{adjustbox}{center}
        \includegraphics[width=\linewidth, trim = 175pt 60pt 175pt 50pt, clip]{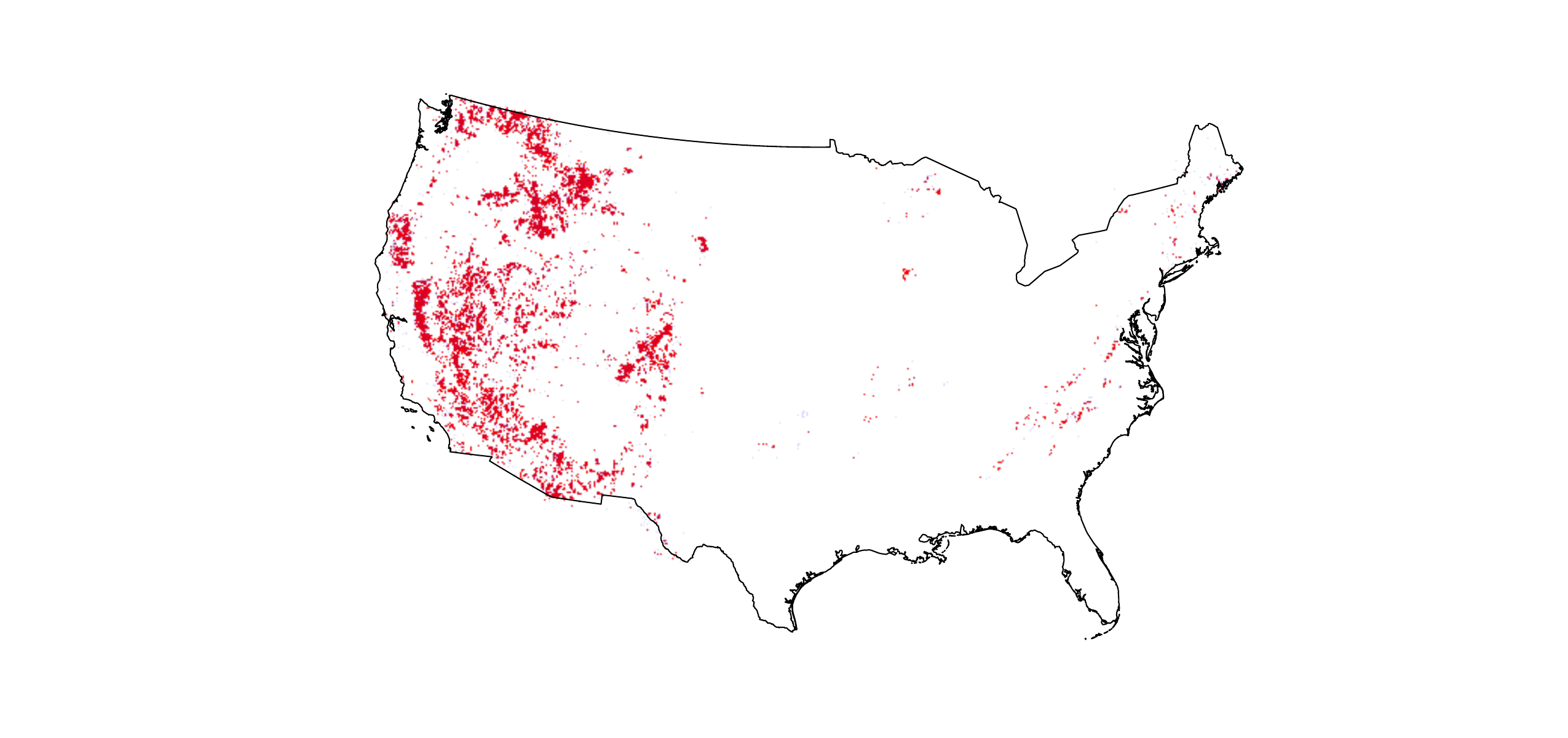}
    \end{adjustbox}
    \caption{Map of the predictions of the model for silver. A red point is a binary flag indicating the prediction of at least one resource within a 5mi$\times$5mi bin.}
    \label{fig:map-pred-silver}
    
\end{figure*}

\begin{figure*}[p]

    \begin{center}
                \textbf{Zinc-specific ground-truth and predictions}
    \end{center}
    \begin{adjustbox}{center}
        \includegraphics[width=\linewidth, trim = 175pt 60pt 175pt 0pt, clip]{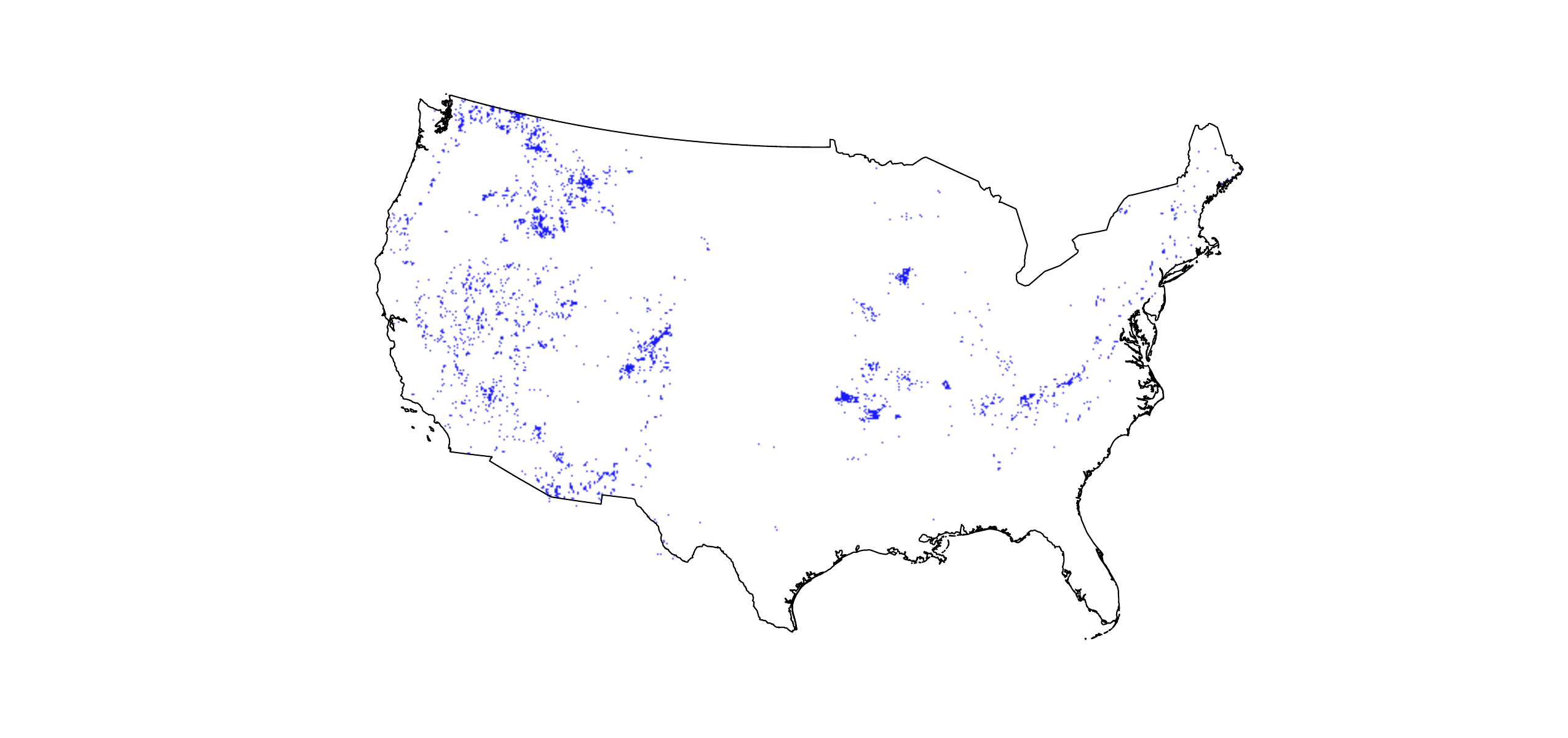}
    \end{adjustbox}
    \caption{Map of the ground-truth data used in the visualization dataset for zinc (EPSG:5070). A blue point is a binary flag indicating the presence of at least one resource within a 5mi$\times$5mi bin.}
    
    \begin{adjustbox}{center}
        \includegraphics[width=\linewidth, trim = 175pt 60pt 175pt 50pt, clip]{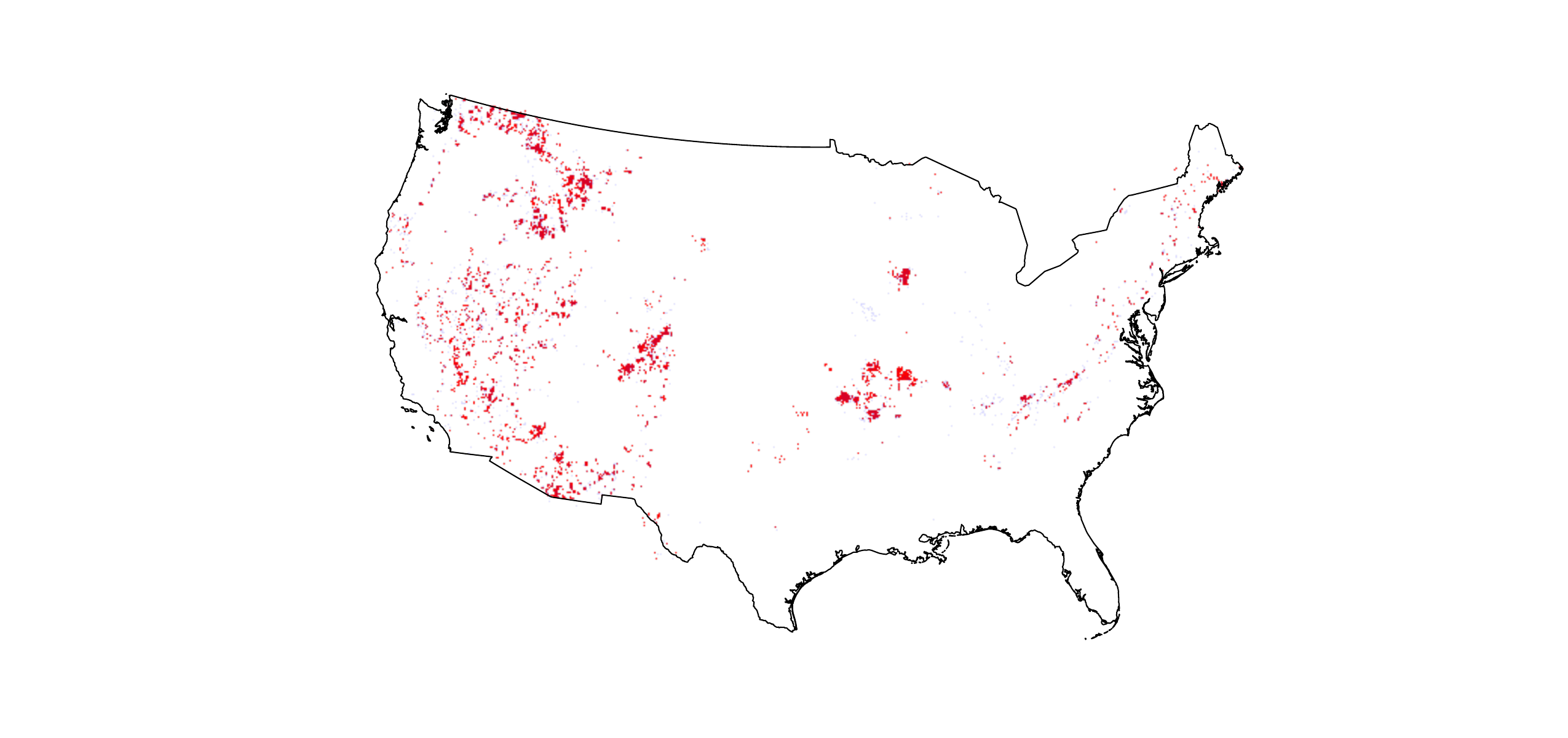}
    \end{adjustbox}
    \caption{Map of the predictions of the model for zinc. A red point is a binary flag indicating the prediction of at least one resource within a 5mi$\times$5mi bin.}
    \label{fig:map-pred-zinc}
    
\end{figure*}

\begin{figure*}[p]

    \begin{center}
                \textbf{Lead-specific ground-truth and predictions}
    \end{center}
    \begin{adjustbox}{center}
        \includegraphics[width=\linewidth, trim = 175pt 60pt 175pt 0pt, clip]{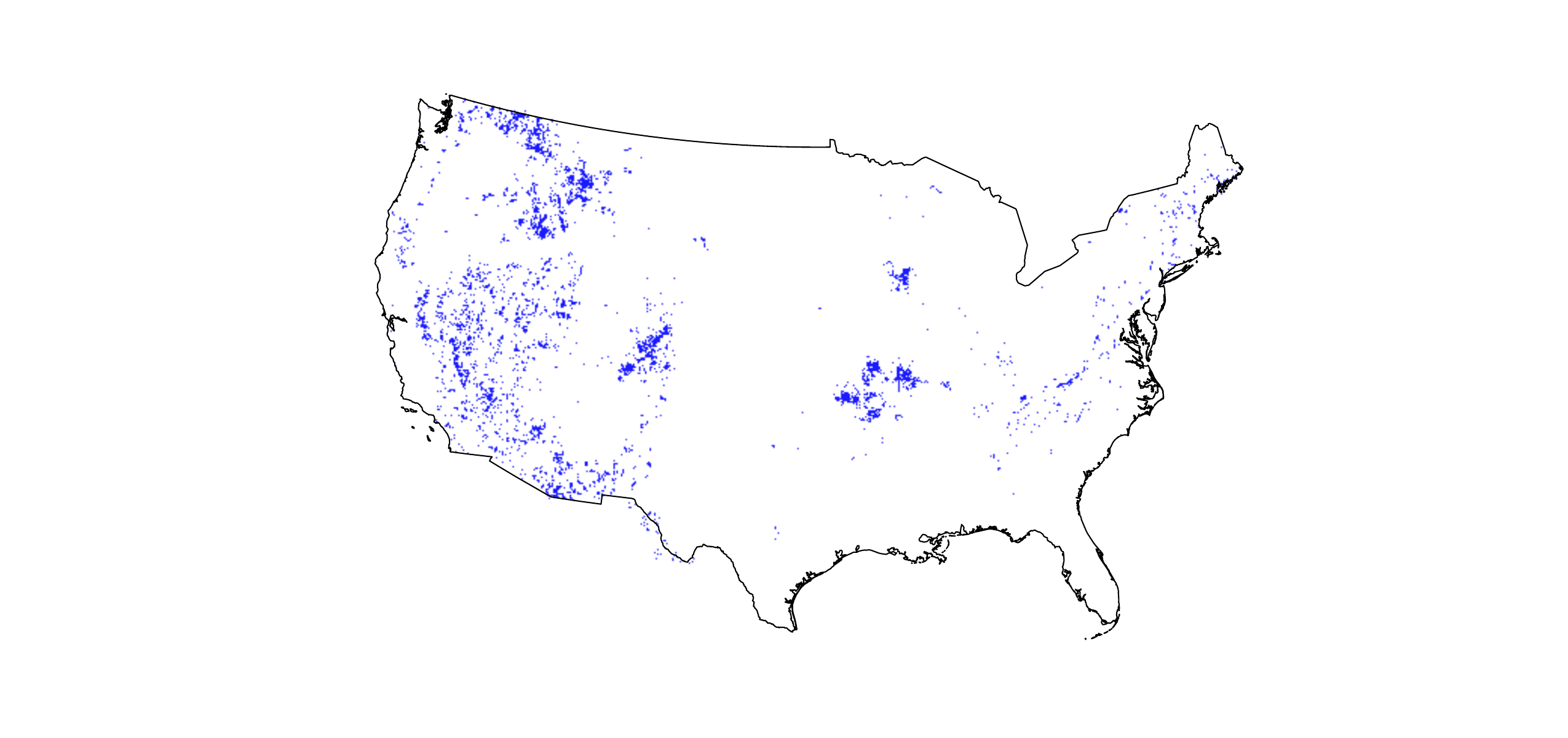}
    \end{adjustbox}
    \caption{Map of the ground-truth data used in the visualization dataset for lead (EPSG:5070). A blue point is a binary flag indicating the presence of at least one resource within a 5mi$\times$5mi bin.}
    
    \begin{adjustbox}{center}
        \includegraphics[width=\linewidth, trim = 175pt 60pt 175pt 50pt, clip]{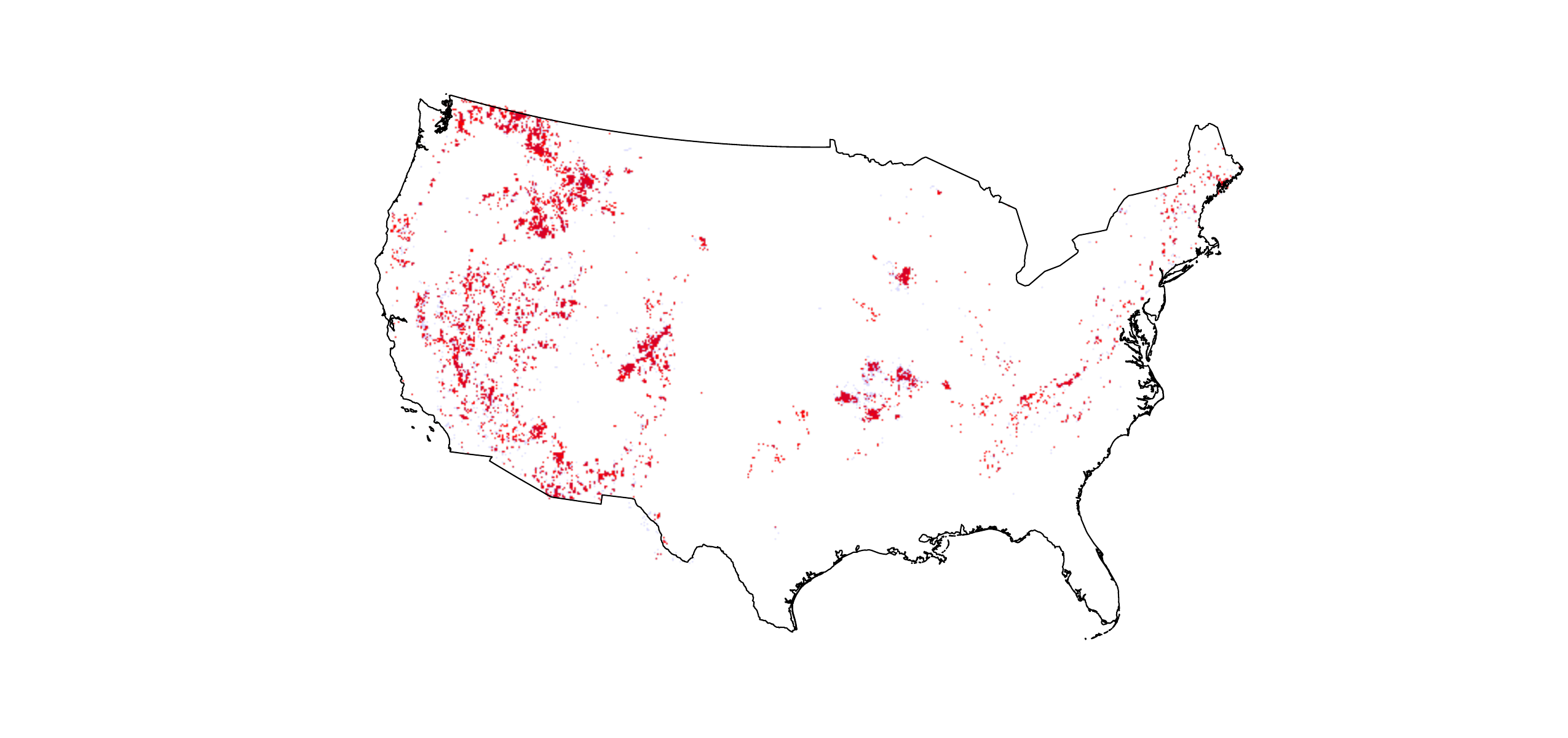}
    \end{adjustbox}
    \caption{Map of the predictions of the model for lead. A red point is a binary flag indicating the prediction of at least one resource within a 5mi$\times$5mi bin.}
    \label{fig:map-pred-lead}
    
\end{figure*}

\begin{figure*}[p]

    \begin{center}
                \textbf{Copper-specific ground-truth and predictions}
    \end{center}
    \begin{adjustbox}{center}
        \includegraphics[width=\linewidth, trim = 175pt 60pt 175pt 0pt, clip]{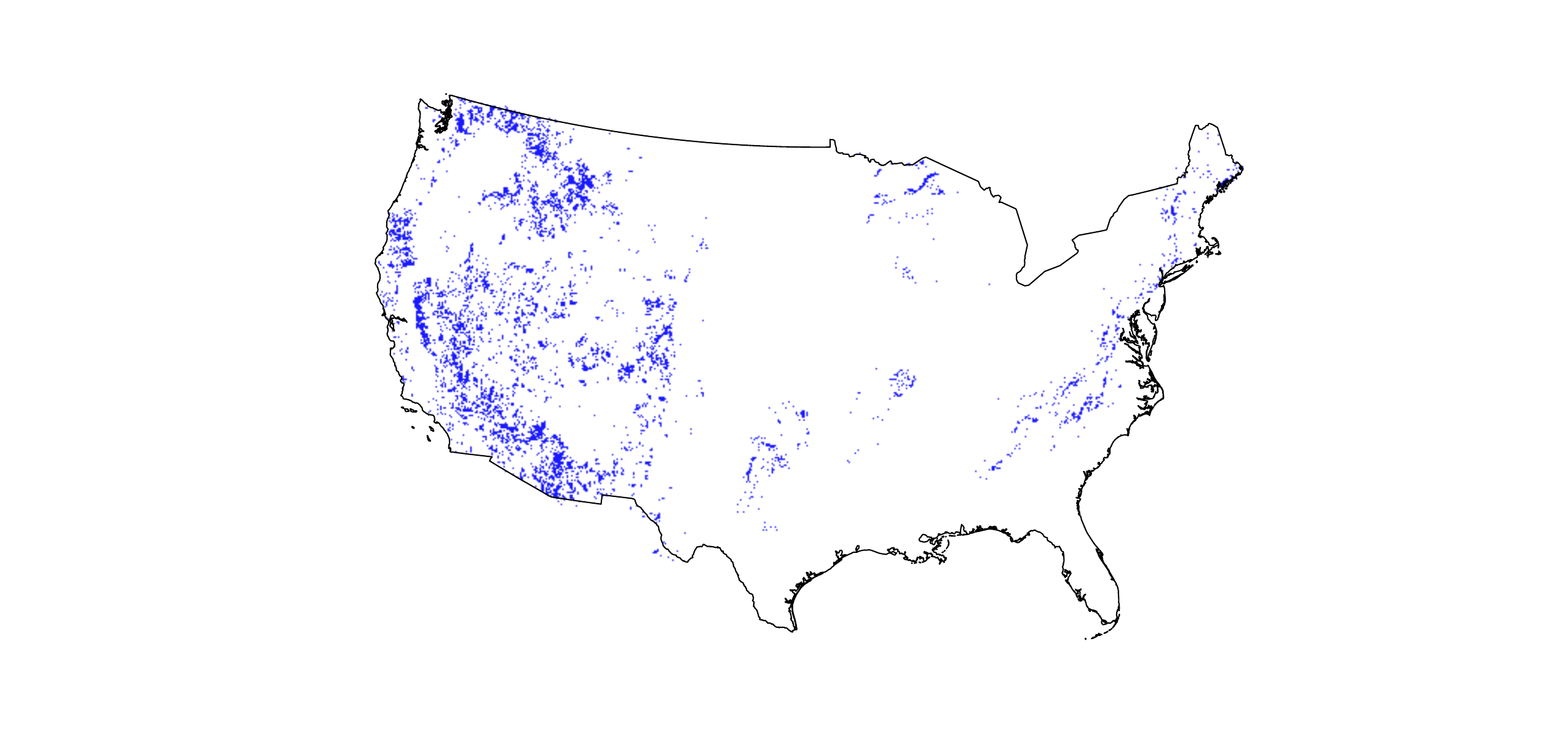}
    \end{adjustbox}
    \caption{Map of the ground-truth data used in the visualization dataset for copper (EPSG:5070). A blue point is a binary flag indicating the presence of at least one resource within a 5mi$\times$5mi bin.}
    
    \begin{adjustbox}{center}
        \includegraphics[width=\linewidth, trim = 175pt 60pt 175pt 50pt, clip]{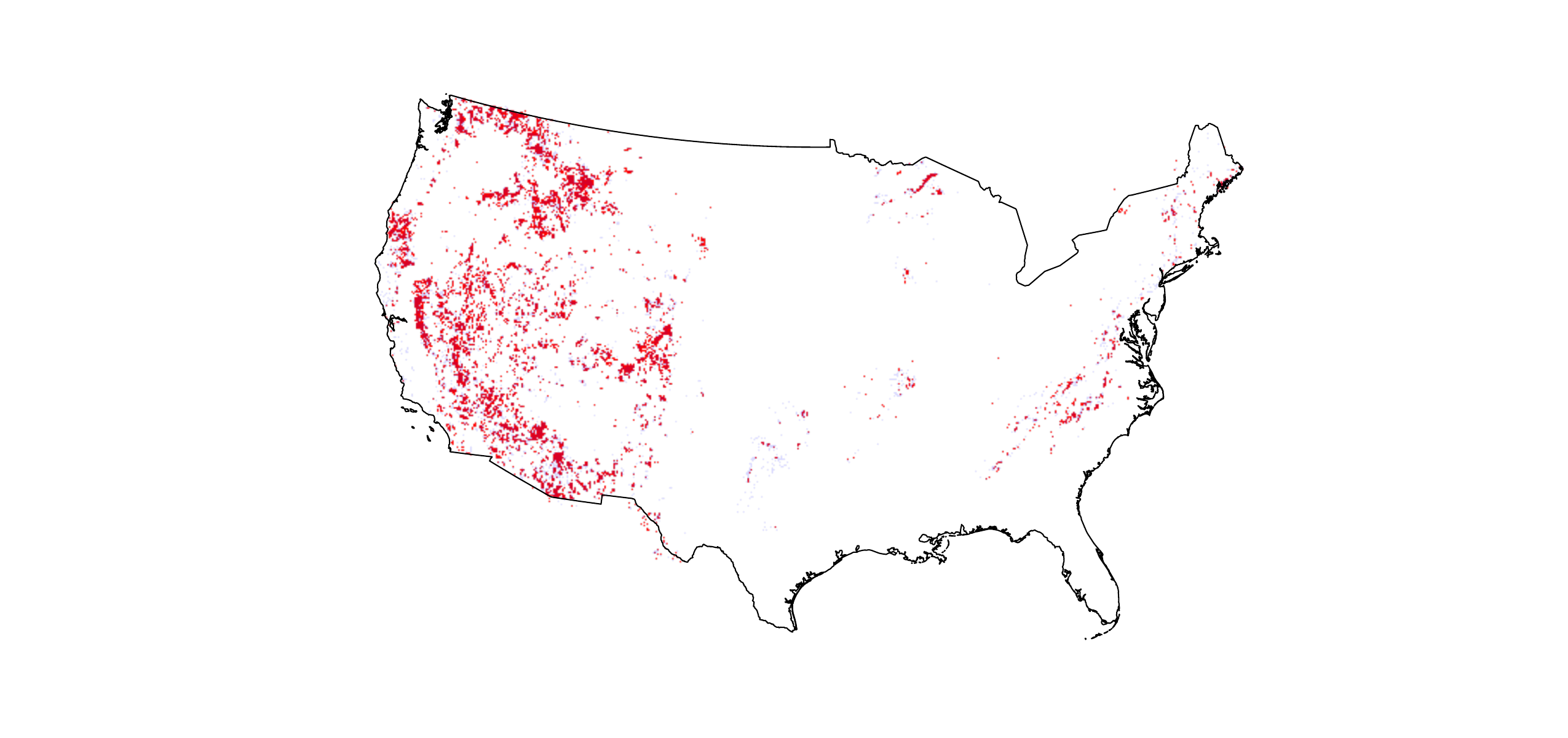}
    \end{adjustbox}
    \caption{Map of the predictions of the model for copper. A red point is a binary flag indicating the prediction of at least one resource within a 5mi$\times$5mi bin.}
    \label{fig:map-pred-copper}
    
\end{figure*}

\begin{figure*}[p]

    \begin{center}
                \textbf{Nickel-specific ground-truth and predictions}
    \end{center}
    \begin{adjustbox}{center}
        \includegraphics[width=\linewidth, trim = 175pt 60pt 175pt 0pt, clip]{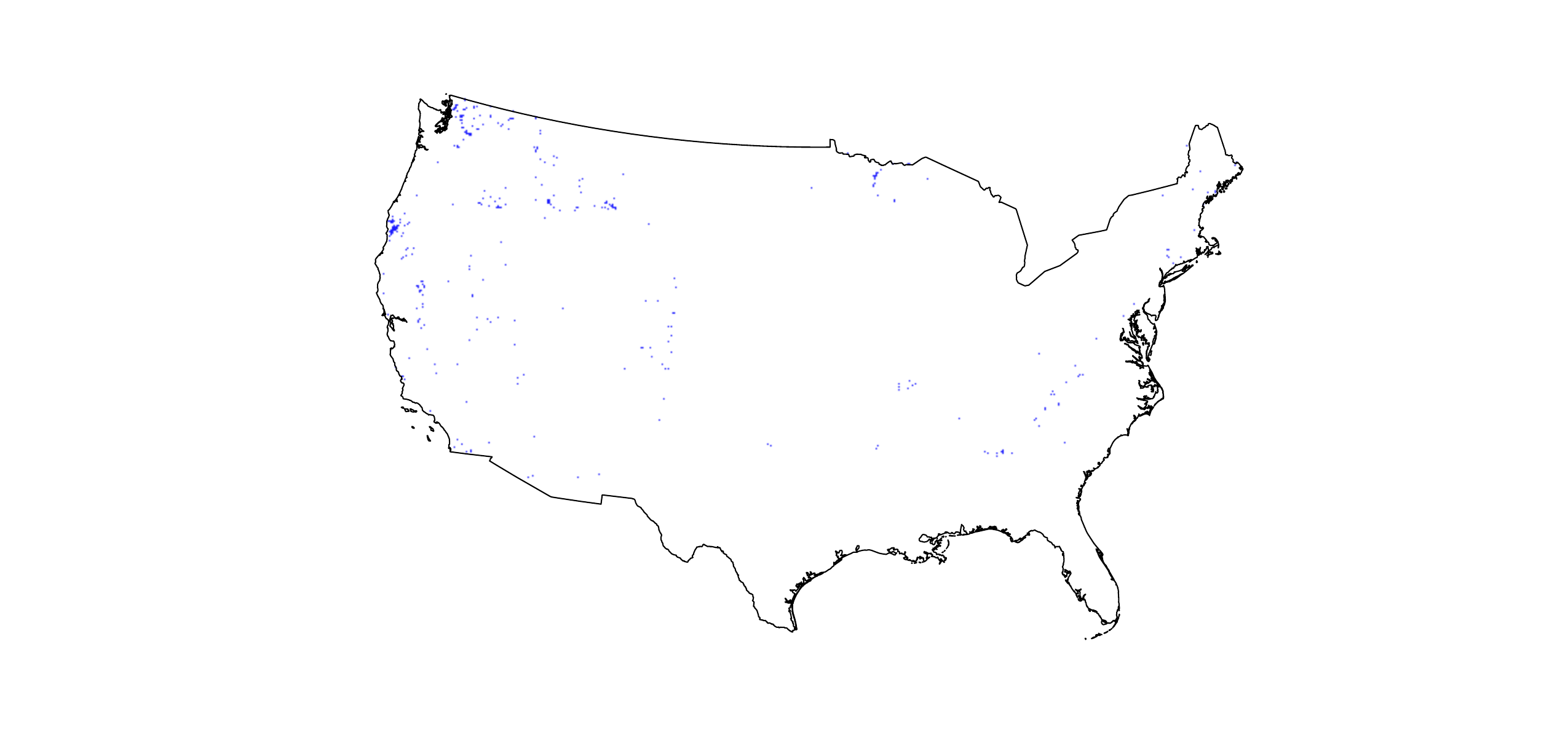}
    \end{adjustbox}
    \caption{Map of the ground-truth data used in the visualization dataset for nickel (EPSG:5070). A blue point is a binary flag indicating the presence of at least one resource within a 5mi$\times$5mi bin.}
    
    \begin{adjustbox}{center}
        \includegraphics[width=\linewidth, trim = 175pt 60pt 175pt 50pt, clip]{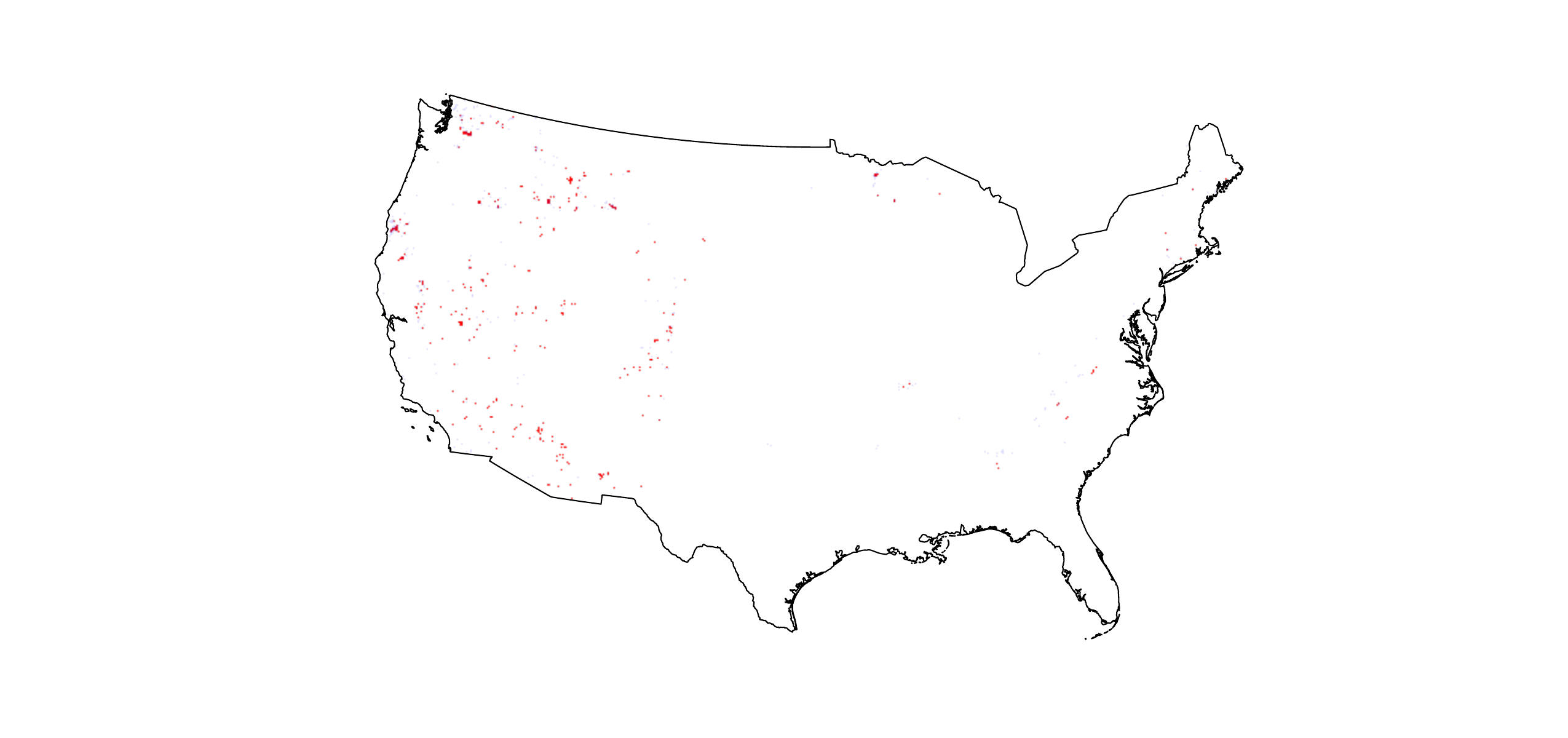}
    \end{adjustbox}
    \caption{Map of the predictions of the model for nickel. A red point is a binary flag indicating the prediction of at least one resource within a 5mi$\times$5mi bin.}
    \label{fig:map-pred-nickel}
    
\end{figure*}

\begin{figure*}[p]

    \begin{center}
                \textbf{Iron-specific ground-truth and predictions}
    \end{center}
    \begin{adjustbox}{center}
        \includegraphics[width=\linewidth, trim = 175pt 60pt 175pt 0pt, clip]{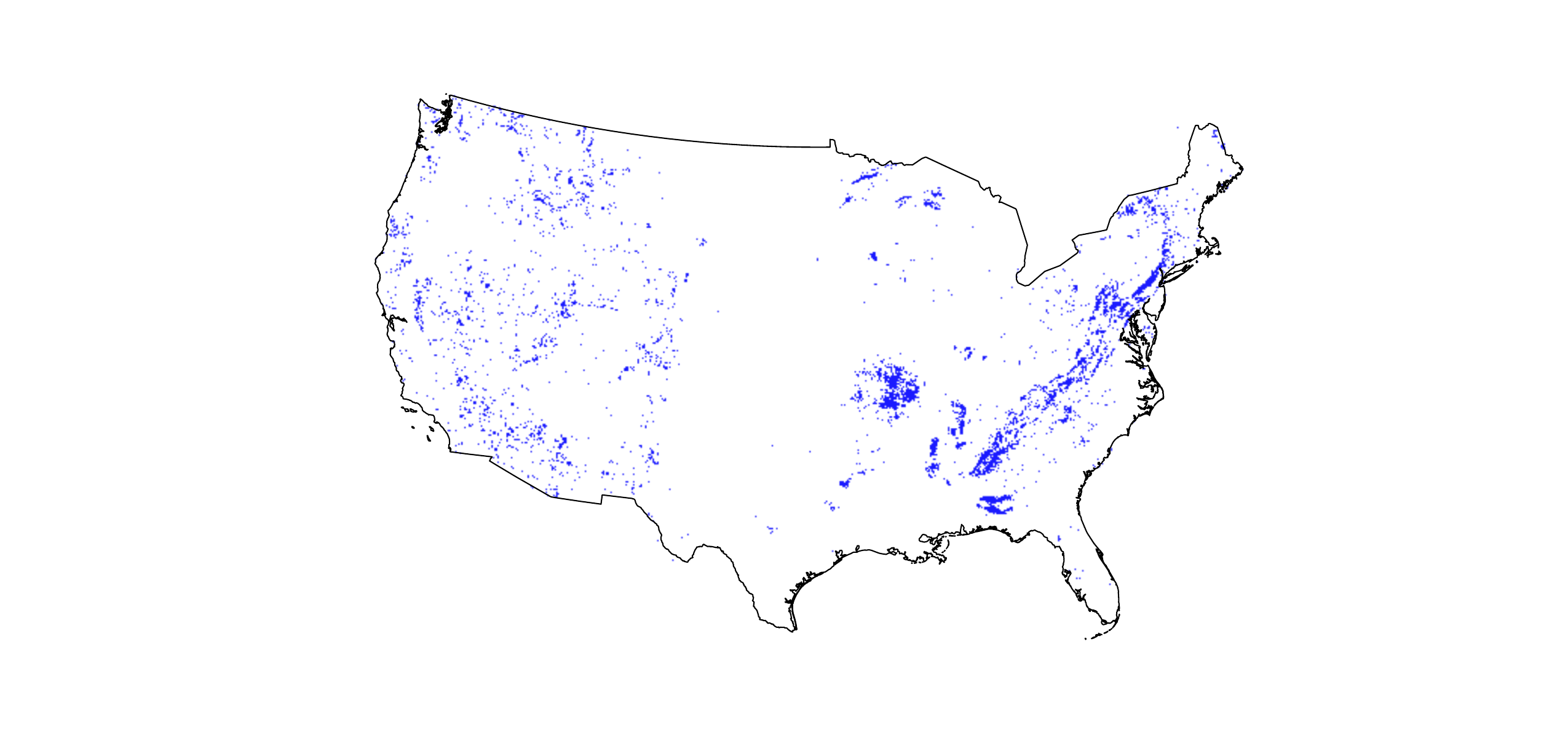}
    \end{adjustbox}
    \caption{Map of the ground-truth data used in the visualization dataset for iron (EPSG:5070). A blue point is a binary flag indicating the presence of at least one resource within a 5mi$\times$5mi bin.}
    
    \begin{adjustbox}{center}
        \includegraphics[width=\linewidth, trim = 175pt 60pt 175pt 50pt, clip]{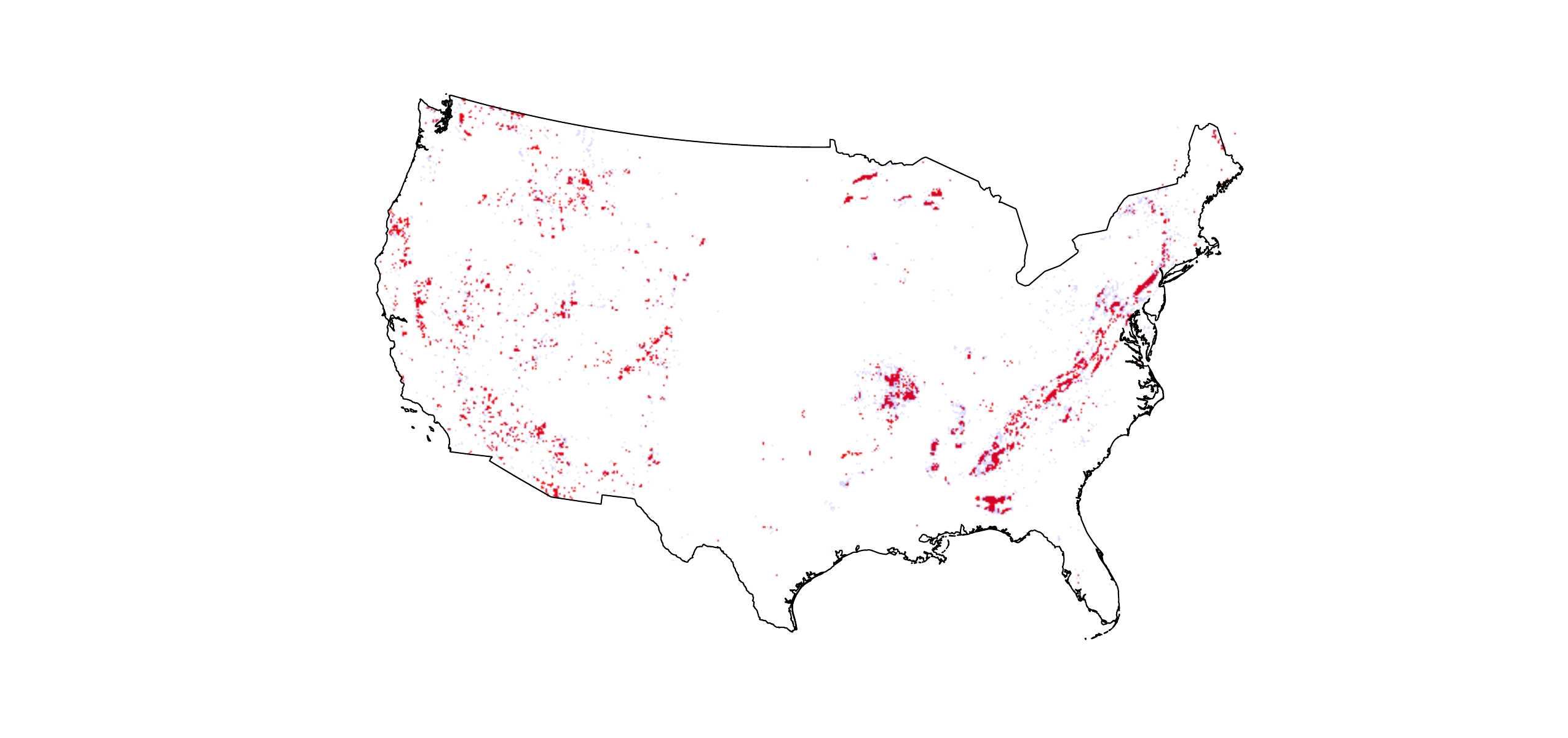}
    \end{adjustbox}
    \caption{Map of the predictions of the model for iron. A red point is a binary flag indicating the prediction of at least one resource within a 5mi$\times$5mi bin.}
    \label{fig:map-pred-iron}
    
\end{figure*}

\begin{figure*}[p]

    \begin{center}
                \textbf{Uranium-specific ground-truth and predictions}
    \end{center}
    \begin{adjustbox}{center}
        \includegraphics[width=\linewidth, trim = 175pt 60pt 175pt 0pt, clip]{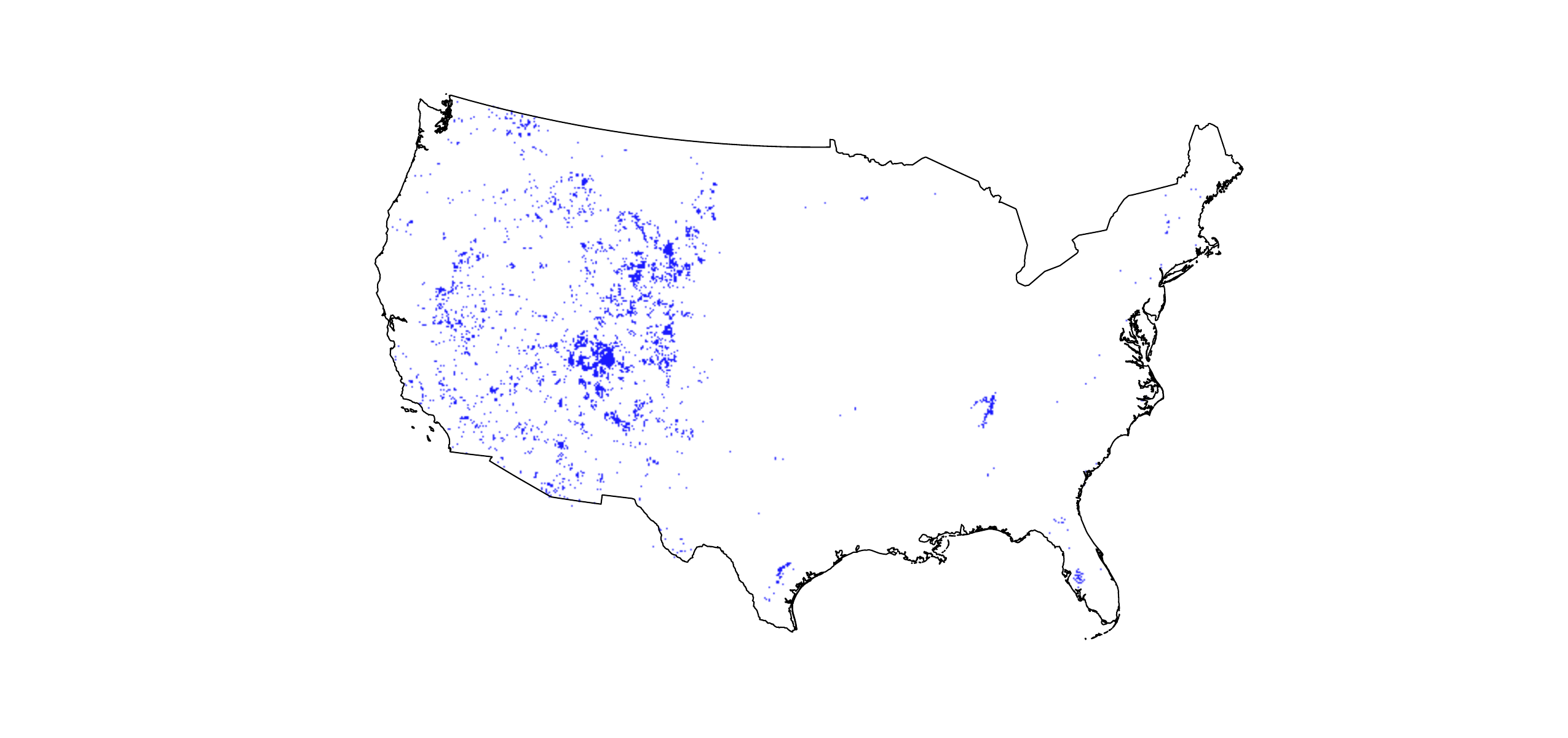}
    \end{adjustbox}
    \caption{Map of the ground-truth data used in the visualization dataset for uranium (EPSG:5070). A blue point is a binary flag indicating the presence of at least one resource within a 5mi$\times$5mi bin.}
    
    \begin{adjustbox}{center}
        \includegraphics[width=\linewidth, trim = 175pt 60pt 175pt 50pt, clip]{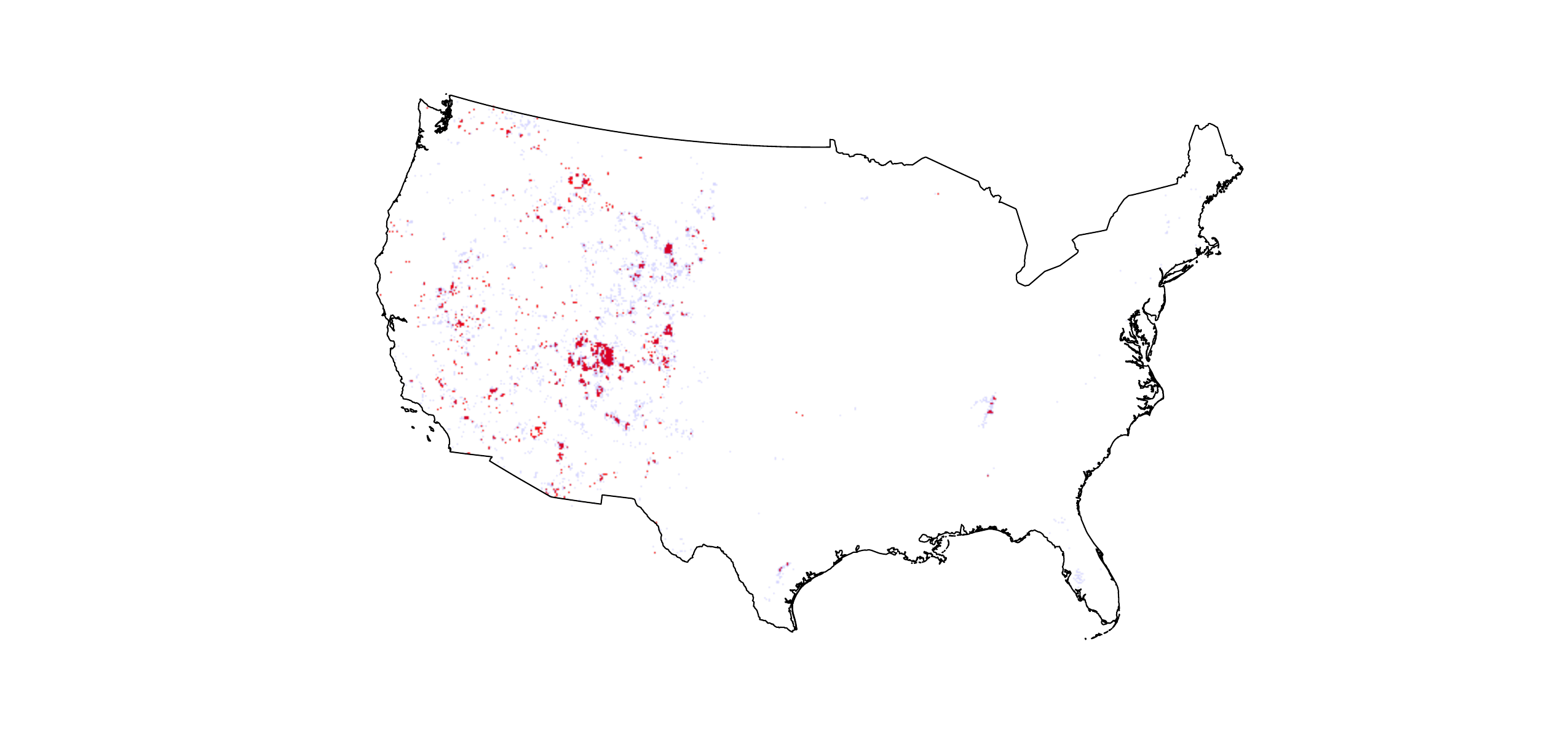}
    \end{adjustbox}
    \caption{Map of the predictions of the model for uranium. A red point is a binary flag indicating the prediction of at least one resource within a 5mi$\times$5mi bin.}
    \label{fig:map-pred-uranium}
    
\end{figure*}

\begin{figure*}[p]

    \begin{center}
                \textbf{Tungsten-specific ground-truth and predictions}
    \end{center}
    \begin{adjustbox}{center}
        \includegraphics[width=\linewidth, trim = 175pt 60pt 175pt 0pt, clip]{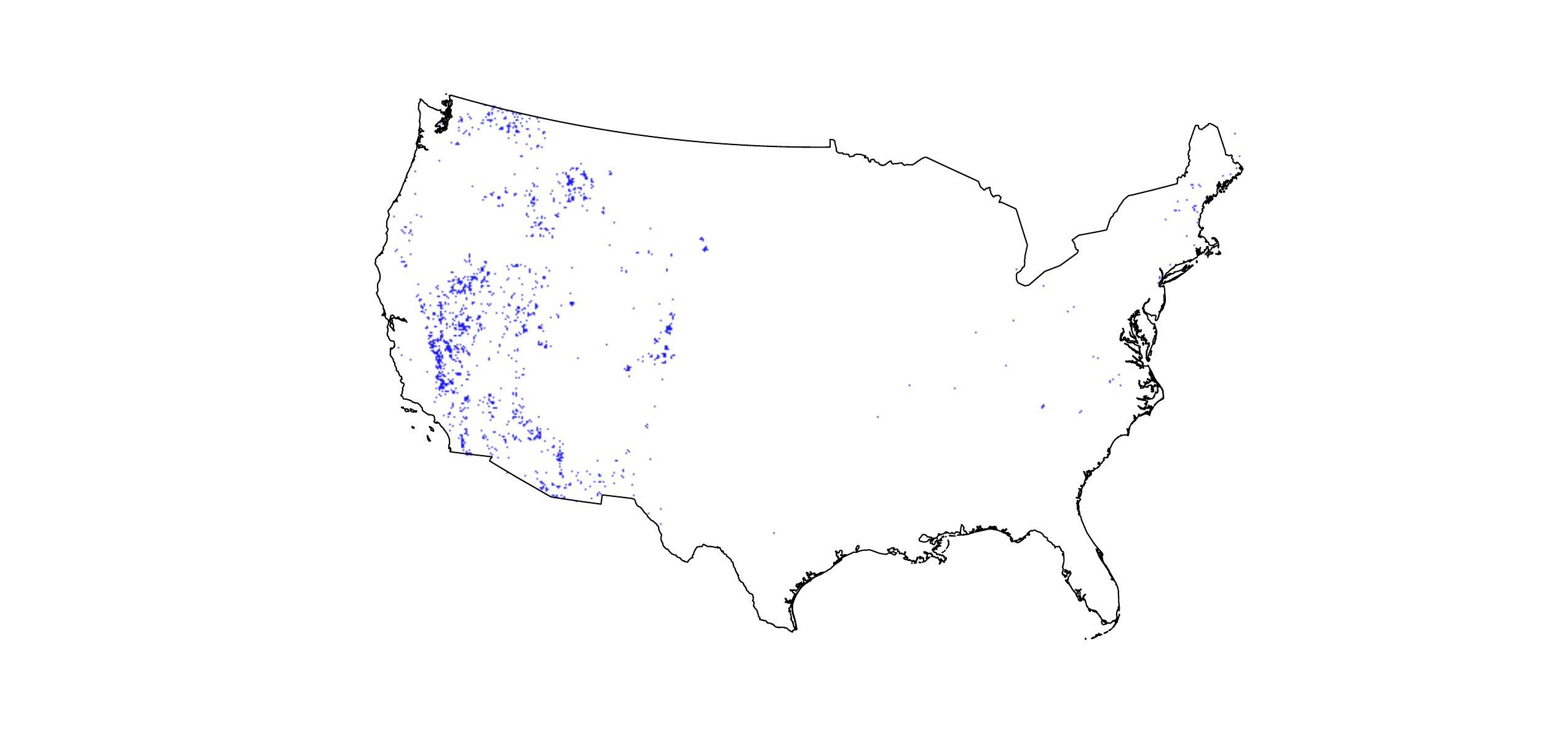}
    \end{adjustbox}
    \caption{Map of the ground-truth data used in the visualization dataset for tungsten (EPSG:5070). A blue point is a binary flag indicating the presence of at least one resource within a 5mi$\times$5mi bin.}
    
    \begin{adjustbox}{center}
        \includegraphics[width=\linewidth, trim = 175pt 60pt 175pt 50pt, clip]{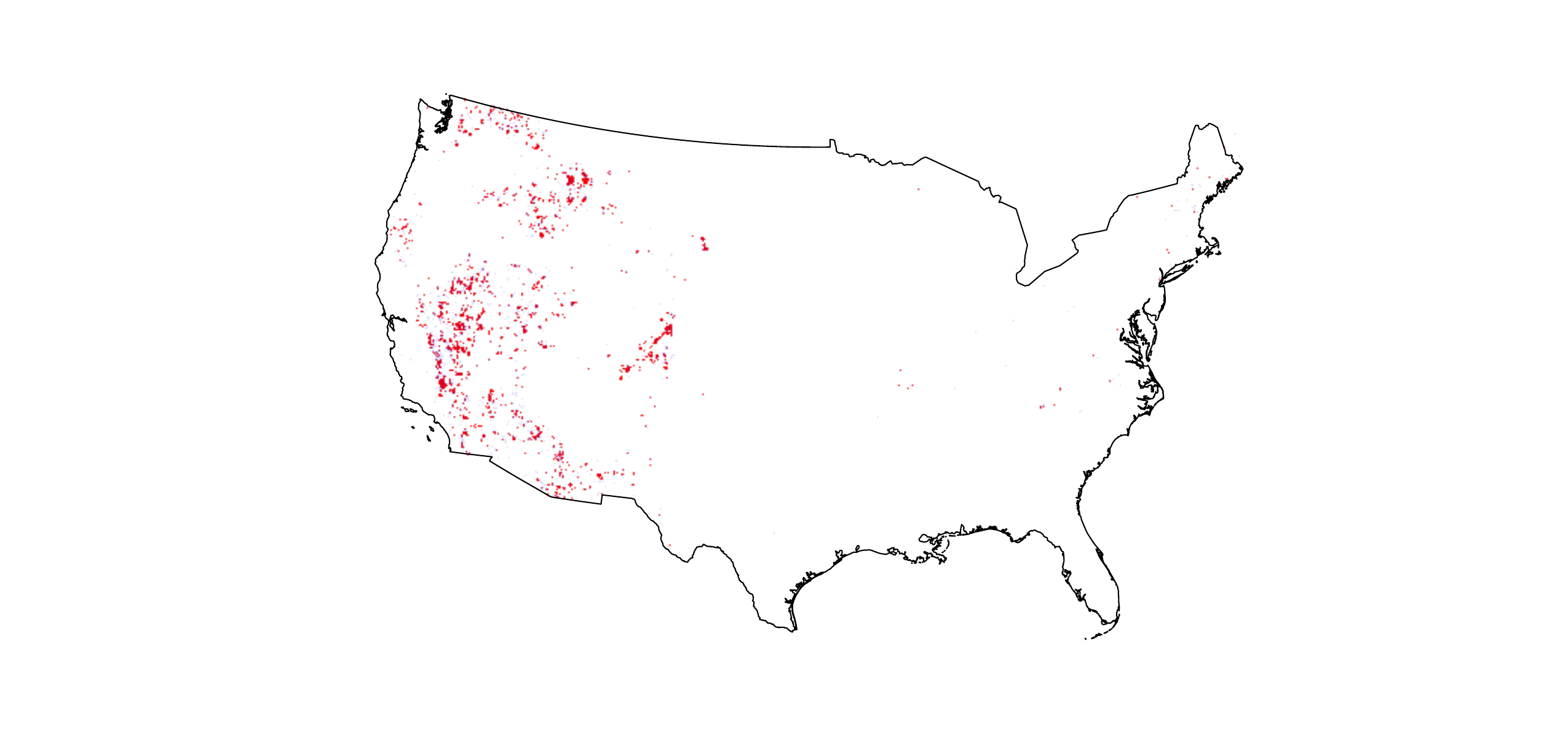}
    \end{adjustbox}
    \caption{Map of the predictions of the model for tungsten. A red point is a binary flag indicating the prediction of at least one resource within a 5mi$\times$5mi bin.}
    \label{fig:map-pred-tungsten}
    
\end{figure*}

\begin{figure*}[p]

    \begin{center}
                \textbf{Manganese-specific ground-truth and predictions}
    \end{center}
    \begin{adjustbox}{center}
        \includegraphics[width=\linewidth, trim = 175pt 60pt 175pt 0pt, clip]{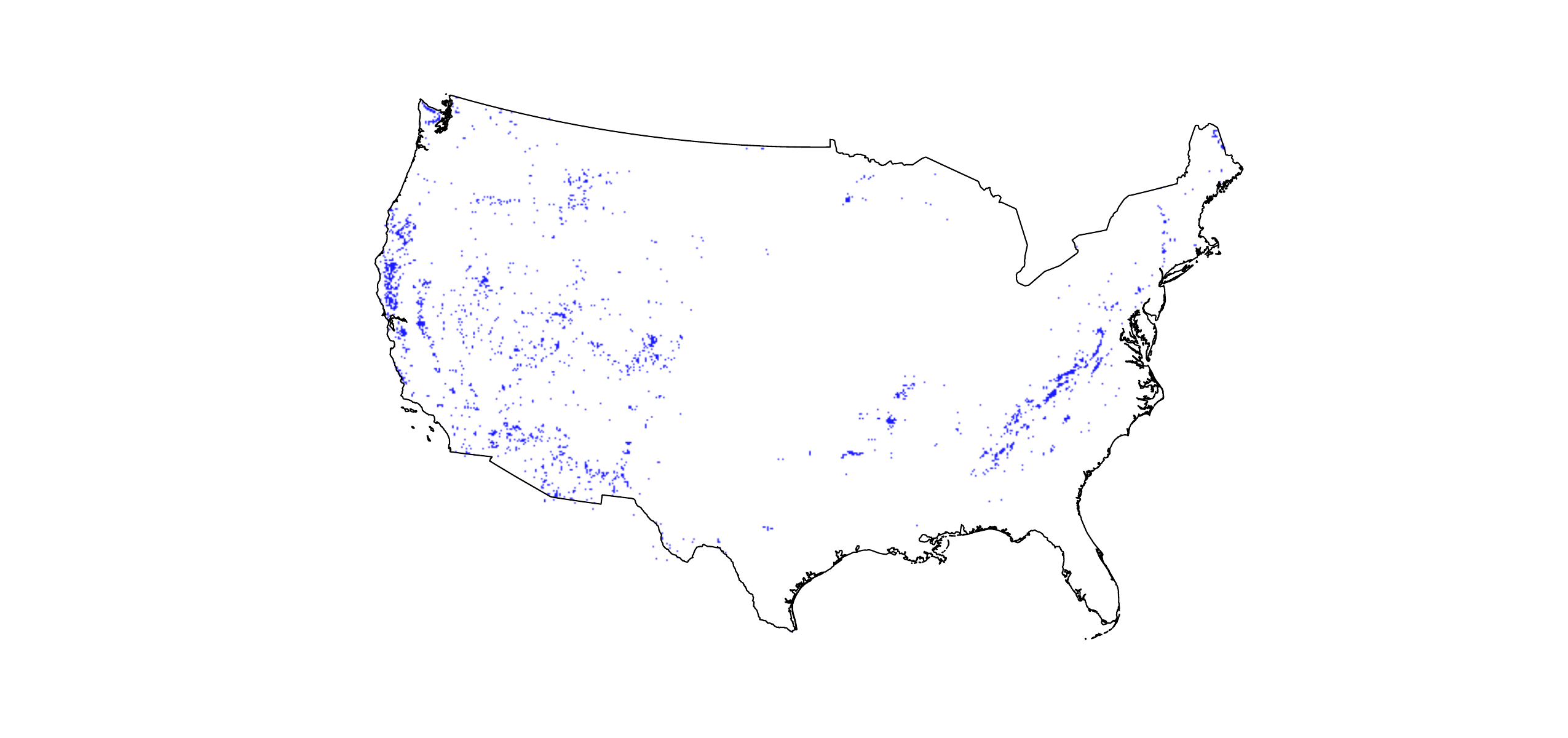}
    \end{adjustbox}
    \caption{Map of the ground-truth data used in the visualization dataset for manganese (EPSG:5070). A blue point is a binary flag indicating the presence of at least one resource within a 5mi$\times$5mi bin.}
    
    \begin{adjustbox}{center}
        \includegraphics[width=\linewidth, trim = 175pt 60pt 175pt 50pt, clip]{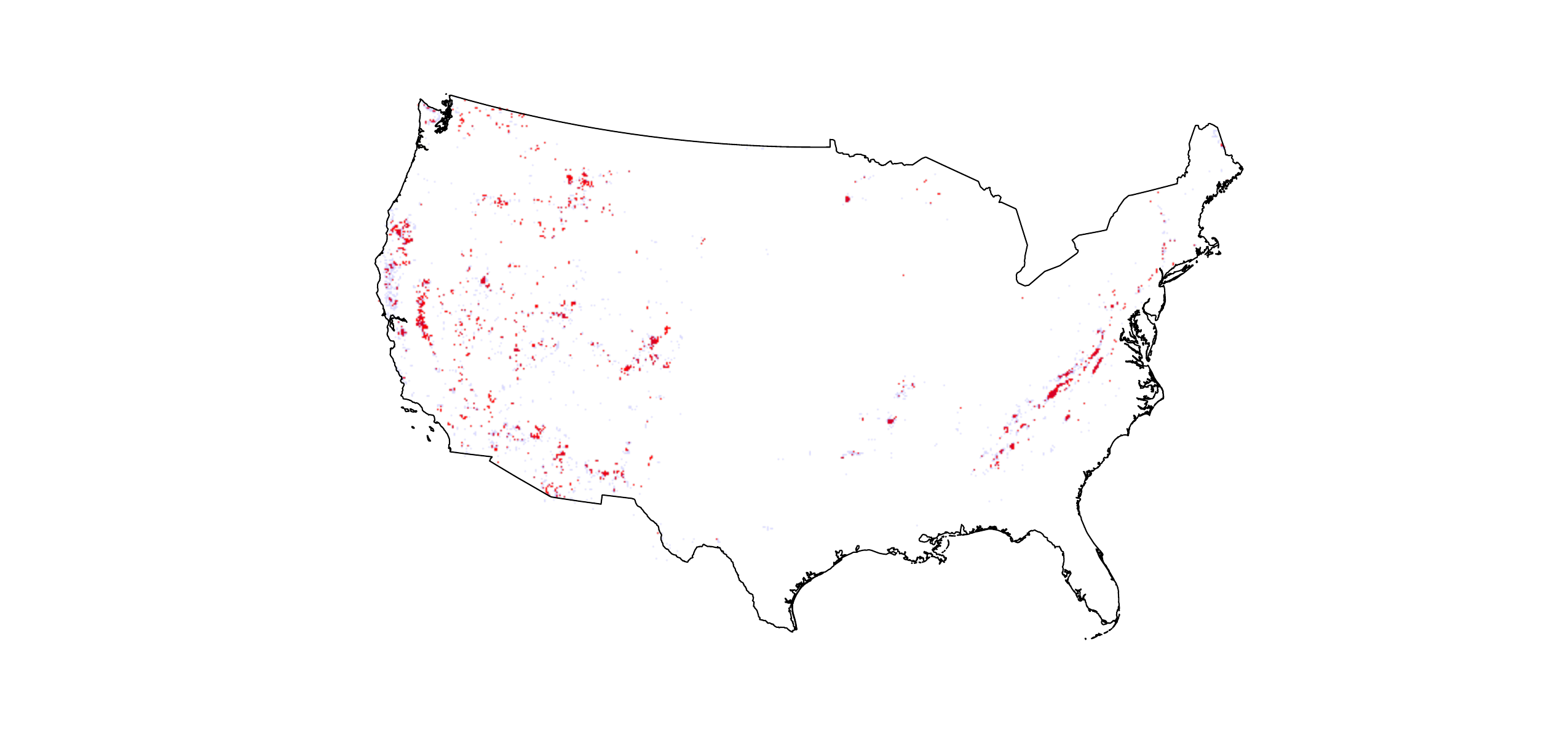}
    \end{adjustbox}
    \caption{Map of the predictions of the model for manganese. A red point is a binary flag indicating the prediction of at least one resource within a 5mi$\times$5mi bin.}
    \label{fig:map-pred-manganese}
    
\end{figure*}

In Figures~\ref{fig:map-gt-countcov}-\ref{fig:map-pred-manganese}, we include the basic components of the map in Figure~\ref{fig:usa-performance}, as well as the geographic coverage of the datasets used for visualization and model training. All prediction maps below were generated by evaluating models trained on all inputs using $N=10$K, $A=0.8$, $H=152$, $K=64$, for 30K gradient steps. Each of 3 random seeds was evaluated on 50 masks per tile in the visualization dataset, i.e. this collection of maps was produced by performing 150 model evaluations per tile.

\section{Computing resources}
\label{app:compute}

Two independent computing systems were utilized to perform the analyses in this study. 

The first was a cluster node with 2 NVIDIA V100s bearing a VRAM capacity of 64 GB (2$\times$32 GB) and 128 GB RAM. Multiple jobs were run per cluster node. We used approximately 2600 GPU hours on this system to perform the main analysis, and a total of approximately 3000 GPU hours including testing runs.

The second was a desktop computer running Ubuntu 22.04.5 LTS using an MSI MEG X670E ACE EATX AM5 motherboard and AMD Ryzen 9 7950X 4.5 GHz 16-Core Processor. It utilized 1 NVIDIA GeForce RTX 4090 graphics card bearing a VRAM capacity of 24 GB, and had access to 128 GB of DDR5-5200 CL40 RAM (4$\times$32 GB). We used approximately 400 CPU hours on this system to perform the main analysis, and a total of approximately 500 CPU hours including testing runs.

\section{Dataset licenses}
\label{app:licenses}

With the exception of the RaCA data used in the agronomic layers, all features considered in this work were in the public domain. The RaCA dataset contains personally-identifying information. As such, a usage agreement was instituted with the USDA so that we could access the unscrambled geolocations associated to each sampling campaign for research purposes. For questions regarding this usage agreement and to access the full RaCA dataset, contact $<$soilshotline@lin.usda.gov$>$.

\section{Ethical Statement}
\label{sec:impacts}
The potential consequences of the presented approach are multi-fold, and span the full range of positive to negative impacts. As established above, addressing climate change will require unprecedented quantities of minerals, and the supply of high-grade resources is small relative to the anticipated demand~\cite{olivetti2018toward,daehn2024key}. Inability to meet this demand is expected to exacerbate global inequality and inhibit developing regions and disadvantaged communities from accessing the materials, technology, and infrastructure required to survive, adapt to, and reverse the effects of climate change~\cite{wang2024regional}. Meanwhile, mining as an industrial practice can have adverse impacts on communities (e.g. noise and dust from detonation, groundwater contamination, energy consumption, and more). Rapid acceleration of mining generally accompanies deregulation and lax permitting, which can exacerbate these negative impacts.

A particularly pressing sequence of events within this sector is unfolding in the present day: the implementation of nation-wide bans on the export of critical minerals could require affected nations to develop redundant stockpiles of those minerals, even though they are already rare in the absence of supply shock~\cite{nassar2024quantifying}. Inability to extract sufficient value from known sources has led to some speculation surrounding the viability of mining in new biomes, e.g. long-studied deposits on or under the sea floor~\cite{DOICritMin}. Meanwhile, some evidence suggests that sizable deposits of relevant minerals have been missed by imperfect record-taking or incomplete prospecting, possibly because many of these metals have only gained economic value alongside the recent onset of climate change~\cite{benson2023hydrothermal}. In addition, newer approaches to mining such as \textit{in situ} recovery can leverage resources in locations and at depths not generally accessible via traditional methods. Such modern techniques generally reduce the environmental footprint of resource extraction by minimizing their intrusion into the Earth's surface. This collection of circumstances reflects multiple important reasons to re-evaluate existing prospecting records.

In this setting, M3 offers continent-scale transparency regarding the locations of mineral deposits. In turn, this knowledge can provide a clearer picture of the quantities of different resources which are readily available for extraction, can support the planning of a just energy transition~\cite{wang2024regional}, and can potentially reduce the need to mine in pristine or nontraditional environments to meet the challenge of the global energy transition. Our method can also in principle be applied to subsurface mapping of sparse features more generally, e.g. for the identification of potential geothermal energy wells~\cite{wang2025machine} and geological reservoir injection sites for CO$_2$ sequestration~\cite{wang2024machine}.

A major technical challenge with M3 is hallucination. Unlike LLM hallucinations, which are generally straightforward to reject on an evidence basis, mineralogical hallucinations may be more difficult to distinguish from a genuine discovery without intensive ground-based data collection efforts. Consistent prediction of a resource excess across multiple models might reasonably motivate a direct sampling campaign, however another danger is the ignition of geopolitical tensions when new mine sites are flagged in regions governed by fragile states. Any unexpected outputs to M3 should be treated with significant skepticism until ground-based data collection has occurred.

\end{document}